%% file: main.tex
\documentclass[runningheads]{llncs}

\usepackage[final,year=2026]{ECCV/eccv}

\usepackage{ECCV/eccvabbrv}

\usepackage{microtype}
\usepackage{graphicx}
\usepackage{booktabs}
\usepackage{multirow}
\usepackage{longtable}
\usepackage{tabularx}
\usepackage{makecell}
\usepackage{caption}
\usepackage{subcaption}        
\usepackage{array}
\usepackage{xspace}
\usepackage{tikz}
\usepackage{pgfplots}

\usepackage{amsmath}
\usepackage{amssymb}
\usepackage{mathtools}

\usepackage{algorithm}
\usepackage{algorithmic}

\usepackage[pagebackref,breaklinks,colorlinks,citecolor=eccvblue]{hyperref}
\usepackage{orcidlink}

\usepackage[most]{tcolorbox}
\usepackage{amsmath}
\usepackage[dvipsnames,svgnames]{xcolor}
\PassOptionsToPackage{dvipsnames,svgnames}{xcolor}

\newcommand{\legenditem}[2]{%
  {{\color{#1}\rule{0.8em}{0.4em}} \!#2}%
}

\usepackage{wrapfig}
\usepackage[skip=4pt]{caption}
\usepackage{rotating} 

\usepackage[capitalize,noabbrev]{cleveref}

\usepackage[textsize=tiny]{todonotes}

\usepackage[numbers]{natbib}
\usepackage{enumitem} 
\tcbuselibrary{theorems} 
\newcommand{\method}{{Vi-CD}\xspace}
\newcommand{\methodfull}{{Visual Circuit Discovery}\xspace}
\newcommand{\ours}{{Vi-CD}\xspace}

\newtcbtheorem[number within=section]{definitionbox}{Definition}%
{colback=gray!5,
 colframe=black,
 fonttitle=\bfseries,
 boxrule=0.8pt,
 sharp corners}{def}

\begin{document}

\title{Seeing Through Circuits: Faithful Mechanistic Interpretability for Vision Transformers}

\titlerunning{Faithful Mechanistic Interpretability for Vision Transformers}

\author{Nina Żukowska \and
        Wolfgang Stammer \orcidlink{0000-0003-3793-8046} \and
        Bernt Schiele \orcidlink{0000-0001-9683-5237} \and
        Jonas Fischer}

\authorrunning{N. Żukowska et al.}

\institute{Max Planck Institute for Informatics, Saarland Informatics Campus, Saarbrücken, Germany}

\maketitle
\begin{center}
    \captionsetup{type=figure}
    \includegraphics[width=0.85\textwidth]{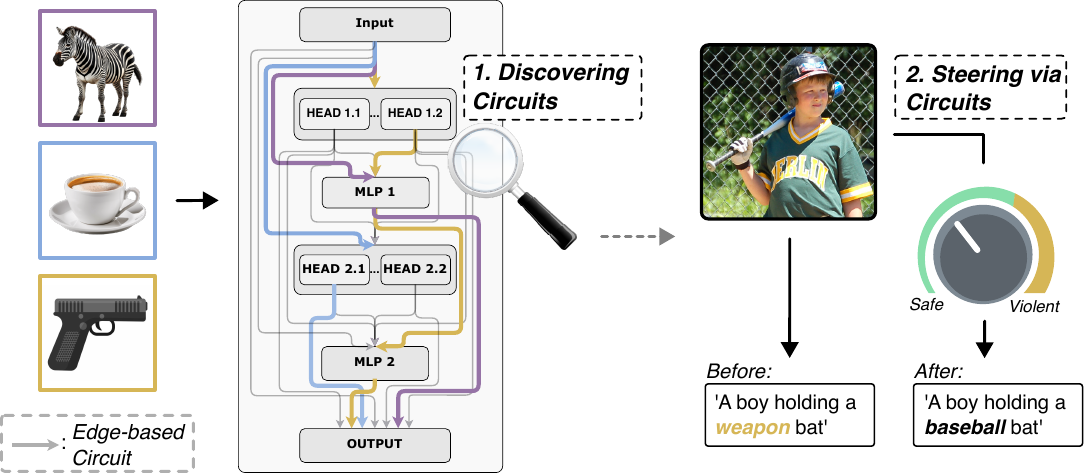}
    \captionof{figure}{\textbf{Discovering Visual Mechanistic Circuits in computation graphs. Left:} We discover sparse, edge-based circuits in Vision Transformers, where each circuit (colored paths) captures the computational subgraph responsible for recognizing a specific class. \textbf{Right:} These circuits enable\textit{ mechanistic steering}. By deactivating a discovered circuit (e.g., one associated with violent content) we can correct model behavior, such as shifting a CLIP model's interpretation of an image. 
    \label{fig:motivation}} 
\end{center}

\begin{abstract}
  \input{sections/abstract}
\end{abstract}

\input{sections/introduction}
\input{sections/related_work}
\input{sections/method}

\input{sections/experiments}
\input{sections/future_work}
\input{sections/conclusions}


\bibliographystyle{splncs04}
\bibliography{references}

\clearpage
\appendix

\input{sections/appendix_steering}
\input{sections/appendix_stability}
\input{sections/appendix_binary_classification}
\input{sections/appendix_ablations}
\input{sections/appendix_max_receiver}

\end{document}

%% file: sections/abstract.tex
{Transparency of neural networks' internal reasoning is at the heart of interpretability research, adding to trust, safety, and understanding of these models. The field of mechanistic interpretability has recently focused on studying task-specific computational graphs, defined by connections (\textit{edges}) between model components. 
Such edge-based circuits have been defined in the context of large language models, yet vision-based approaches so far only consider \textit{neuron}-based circuits. These tell \textit{which} information is encoded, but not \textit{how} it is routed through the complex wiring of a neural network.
In this work, we investigate whether useful mechanistic circuits can be identified through computational graphs in vision transformers. 
We propose an effective method for Automatic Visual Circuit Discovery (\method) that recovers class-specific circuits for classification,  identifies circuits underlying typographic attacks in CLIP, and discovers circuits that lend themselves for steering to correct harmful model behaviour. 
Overall, we find that insightful and actionable edge-based circuits can be recovered from vision transformers, adding transparency to the internal computations of these models.}

%% file: sections/introduction.tex
\section{Introduction}

Modern vision models have achieved remarkable performance across a wide range of tasks, from image classification to open-vocabulary recognition.
Yet despite their success, these models remain largely opaque: it is difficult to understand \emph{why} a model arrives at a particular prediction, and consequently difficult to control or correct its behavior when it fails.
This opacity has practical consequences beyond scientific curiosity.
Models are vulnerable to adversarial attacks such as typographic attacks on CLIP~\cite{hufe2026dyslexify, wang2025typographic, westerhoff2025scam}, susceptible to shortcut learning~\cite{SchramowskiSTBH20, geirhos2020shortcut, steinmann2024navigating}, and difficult to debug when deployed in safety-critical settings~\citep{bereska2024mechanistic}. 

Mechanistic interpretability aims to address these challenges by reverse-engineering the internal computations of neural networks.
A central concept in this field is the \emph{circuit}: a sparse subgraph of a model's computational graph that implements a specific behavior of interest~\cite{cammarata2020thread, elhage2021mathematical, wang2022interpretability, hanna2023does, lindner2023tracr}.
In transformer language models, circuit discovery has converged on an \emph{edge-centric} formulation, where models are represented as directed graphs over residual-stream connections between attention heads and MLP blocks, and the goal is to identify a minimal subgraph responsible for a given behavior~\cite{conmy2023towards, bhaskar2024finding}.
A key requirement for such circuits is \emph{faithfulness}: a circuit is faithful if restricting the model's computation to only that circuit preserves its original task performance.
Faithfulness ensures that the discovered circuit genuinely accounts for the model's behavior, rather than merely correlating with it.

In vision models, however, circuit discovery has remained largely at the neuron, channel, or feature level~\cite{cammarata2020thread, rajaram2024automatic, dreyer2024pure}.
While such node-based circuits reveal \emph{which} components are important for a given task, they do not capture \emph{how} information flows between them.
Edge-based circuits address this limitation by explicitly modeling the connections between components, offering several advantages:
they can disentangle the contributions of polysemantic neurons that participate in multiple circuits,
they enable causal interventions via edge-level activation patching,
and they support more precise notions of faithfulness since individual connections can be included or excluded without ablating entire components.
Despite these benefits, edge-based circuits have been little investigated in the vision domain. 

In this work, we investigate whether the advantages of edge-based circuit discovery, so far demonstrated primarily in large language models, transfer to large-scale vision transformers.
We introduce \methodfull{} (\method{}\footnote{Pronounced “VIK-id” (like \emph{wicked} with a \emph{v}.)}) based on sequential activation patching with corrupted, inpainted data points. \ours applies Automatic Circuit Discovery to the visual domain, enabling the extraction of sparse, faithful subgraphs from models such as ViT-B and OpenCLIP (\autoref{fig:motivation}, left).
In summary, our contributions are:
\begin{itemize}
    \item We introduce \method, an \textbf{edge-based circuit discovery for Vision Transformers}, inspired by computational graph-based mechanistic interpretability. 
    \item We \textbf{demonstrate the scalability} of \method to large-scale models such as ViT-B and OpenCLIP.
    \item We demonstrate a practical relevance of our circuits by performing \textbf{mechanistic steering to defend against typographic attacks} in CLIP, showcasing how circuit-level understanding enables targeted model corrections.
\end{itemize}

We show that our recovered circuits are both faithful and up to 10x sparser compared to prior circuit discovery approaches, capturing class-specific computational pathways with a small fraction of the model's total edges.
Importantly, we go beyond passive faithfulness evaluation by demonstrating that discovered circuits support \emph{mechanistic steering}: targeted activation or deactivation of circuit edges to deliberately alter model behavior (\autoref{fig:motivation} (right)).
We validate this in two attack settings, showing that deactivating specific circuits can defend against typographic attacks in CLIP and correct misclassifications between confusable class pairs.
This provides evidence that \ours captures genuine computational mechanisms, not mere statistical artifacts,
offering an insightful and actionable view on the internal computations of modern vision transformers.


%% file: sections/related_work.tex
\section{Related Work}

\noindent \textbf{Mechanistic Circuits in Large Language Models.}
In transformer language models, circuit discovery has converged to an \emph{edge-centric} formulation. Hereby, models are represented as directed computation graphs over residual-stream connections between attention heads and MLP blocks, and the goal is to identify a minimal subgraph responsible for a specific behavior~\cite{conmy2023towards, bhaskar2024finding}.
A central approach of Conmy \etal \cite{conmy2023towards} iteratively prunes edges via activation patching to recover mechanistic circuits.
However, the reliance on expensive per-edge interventions, motivated a family of more efficient approximations.
EAP~\cite{syed2024attribution} estimates edge importance using first-order gradient approximations, trading faithfulness for scalability.
To address this, integrated-gradient variants~\cite{hanna2024have} stabilize attribution estimates and improve faithfulness in deep nonlinear settings.
Along a complementary direction, Relevance Patching~\cite{jafari2025relp} replaces gradient-based coefficients with Layer-wise Relevance Propagation signals, while CD-T~\cite{hsu2024efficient} enables circuit extraction at multiple levels of granularity.
Crucially, all of these methods have been developed and evaluated exclusively in the context of large language models and a transfer to vision models is, due to the vastly different input, non-trivial.
Whether edge-based faithful circuits can be meaningfully recovered in vision architectures remains an open question, that we address in this work.

\noindent \textbf{Mechanistic Circuits in Vision Models.}
In contrast to language models, circuit discovery in vision models has historically focused on neuron-, channel-, or feature-level representations.
Early work~\cite{cammarata2020thread} studied circuits as compositions of interpretable features across layers in convolutional networks, establishing a neuron-centric view of visual circuits.
Subsequent work has extended this perspective in several directions:
\cite{hamblin2022pruning} identify feature-preserving subnetworks via pruning,
~\cite{dreyer2024pure} disentangle polysemantic neurons into concept-specific circuits,
and ~\cite{kowal2024visual} propose connectome-style visualizations that trace feature interactions across all layers.
In the transformer setting, ~\cite{wang2025discovering} identify \emph{influential neuron paths}, i.e., sequences of neurons whose joint activity drives predictions. 
Similarly, ~\cite{rajaram2024automatic} extract concept-specific circuits based on functional connectivity across layers.
Recently, ~\cite{anani2026certified} address the brittleness of neuron-based circuit discovery by introducing provable stability guarantees via randomized smoothing, certifying that circuit membership decisions remain invariant under bounded dataset edits.
While these approaches offer valuable insights into which components are important, they remain fundamentally \emph{node-based}: they identify relevant neurons or features but do not explicitly model connections, such as the residual computation graph, between components.
In this work, we close this gap by introducing \emph{edge-based} circuit discovery 
in vision transformers for the first time.

%% file: sections/method.tex
\section{\method: Automatic Visual Circuit Discovery}
\label{sec:method}
\begin{figure}[t]
  \centering
  \includegraphics[height=13.5\baselineskip]{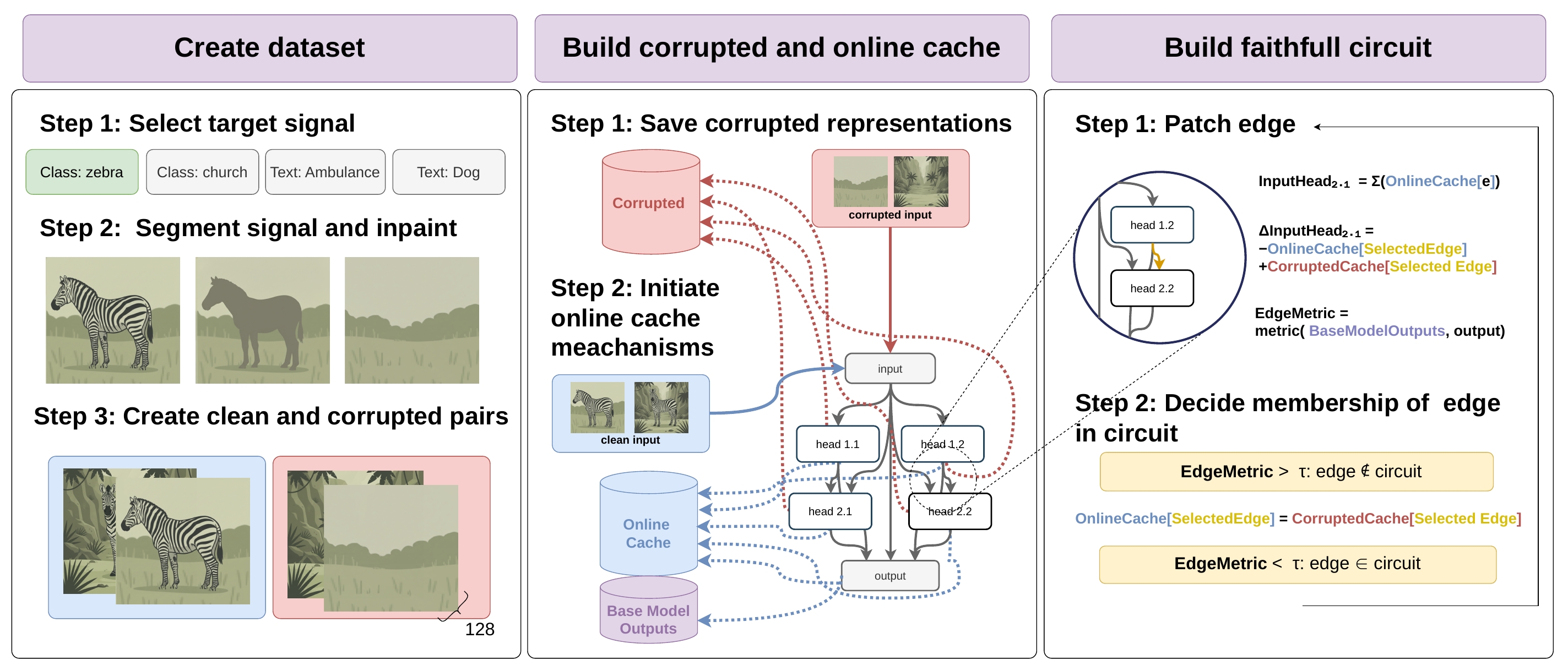}
  \caption{
\textbf{\ours: Circuit discovery in vision computation graphs.}
Left: Construction of corrupted inputs by removing class-specific information via a segmentation and inpainting pipeline.
Middle: Graph representation of the model, where nodes correspond to
residual-stream components (attention heads and MLPs) and edges represent
information flow between them.
Right: Activation patching, where contributions along edges in the
candidate circuits are replaced with clean activations while non-circuit edges
retain corrupted activations, allowing identification of faithful circuits.
}
  \label{fig:circuits_overview}
\end{figure}
We aim to identify a set of residual-stream connections in a vision transformer that is sufficient to capture the model’s behavior for a given task (e.g., classification). To this end, we formalize the model as a directed graph in which nodes correspond to model components and edges represent residual-stream connections between these components. Circuit discovery is then cast as edge selection over this graph.

To facilitate the discovery of meaningful circuits, we employ activation patching together with a decision-aligned pruning criterion. To make circuit discovery tractable in modern vision architectures, we introduce a graph-reduction strategy and construct clean–corrupted input pairs that remove object-level evidence while avoiding distributional artifacts. The resulting class circuit framework unifies the dataset, evaluation metric, and edge set, and enables downstream validation through targeted steering. An overview of the framework is provided in \autoref{fig:circuits_overview}.

We first describe the underlying assumptions of our approach. We then formalize the simplification of the computation graph and outline the key mechanism used to construct datasets for mining visual circuits. Lastly, we describe how discovered circuits can be practically used as basis for model steering.

\subsection{\textbf{Background.}\label{sec:method:background}} 
Following prior work by Conmy \textit{et al.}~\cite{conmy2023towards}, we formalize circuit extraction
in graph-theoretic terms.

\paragraph{Model and Residual Stream.}
Consider a transformer model with residual stream
$r^{(\ell)} \in \mathbb{R}^d$ at layer $\ell$.
We assume the common \emph{additive} residual stream, i.e.,
\begin{equation}
r^{(\ell+1)}
= r^{(\ell)}
+ \sum_{h \in \mathcal{H}_\ell} a_h^{(\ell)}
+ \sum_{m \in \mathcal{M}_\ell} m^{(\ell)},
\end{equation}
where $\mathcal{H}_\ell$ and $\mathcal{M}_\ell$ denote the sets of attention
heads and MLP blocks at layer $\ell$, respectively, and
$a_h^{(\ell)}, m^{(\ell)} \in \mathbb{R}^d$ are their contributions
to the residual stream.

\paragraph{Faithful Circuits.} Let us now provide a formulation of circuits in terms of the (computation) graph of the network and its residual stream connections: 
\begin{definitionbox}{Circuit}{circuit}
Let $G=(V,E)$ be a directed graph representing a model, where nodes
$v\in V$ are model components and edges $(u,v)\in E$ correspond to
additive contributions transmitted via the residual stream. A \emph{circuit} is a subset of edges $E_C\subseteq E$.
Equivalently, it is specified by indicators $i_e\in\{0,1\}$ for
each $e\in E$. 
\end{definitionbox}

Next, to discover meaningful circuits, we need to define what constitutes a \textit{good} circuit for a given task. We here utilize circuit faithfulness to assess circuit quality, adopting the established definition of~\citep{hanna2024have}:

\begin{definitionbox}{Circuit Faithfulness; \cite{hanna2024have}}{faithfulness}
Let $T$ denote a task (e.g., image classification).
A \textbf{faithful circuit} for task $T$ is defined as a subgraph $E_C \subseteq E$ such that restricting computation
to $E_C$ preserves task performance up to an $\epsilon$, i.e.,
\begin{equation}
\mathcal{M}_T(E) - \mathcal{M}_T(E_C) \le \epsilon,
\end{equation}
where $\mathcal{M}_T$ denotes the task fidelity metric, which should differ from the pruning criterion.
\end{definitionbox}

We require a mechanism that only employs candidate circuits $E_C$ within a single forward pass and enables evaluation of single-edge behavior. Because residual-stream contributions are additive, edge-level interventions can be implemented by selectively substituting sender outputs while leaving all other prior computations unchanged. This yields a well-defined procedure for creating and testing candidate circuits, which we implement via activation patching.

\paragraph{Clean and Corrupted Inputs.}
Following standard practice, we consider pairs of
\emph{clean} and \emph{corrupted} inputs. A clean input $x$ contains the
task-relevant information required for correct model behavior (e.g.,
an image with the object of class "zebra"), whereas a corrupted input
$\tilde{x}$ is constructed by removing this information while keeping
other statistics of the input largely unchanged (e.g., via inpainting, see \autoref{fig:circuits_overview}), which is non-trivial for Vision Models with non-discrete tokens. Applying the model to
$x$ and $\tilde{x}$ yields the \emph{clean run} and \emph{corrupted run},
respectively, with the \textit{corrupted run} providing the activations for circuit
interventions.

\paragraph{Activation Patching.}
For each node $u \in V$, let $r_u(x) \in \mathbb{R}^d$ denote its
contribution to the residual stream under input $x$.
For a node $v$, define its set of predecessors
$\mathcal{P}(v) = \{ u \mid (u \to v) \in E \}$.
We now swap the residual connections that are not part of the circuit by the
residual contributions we get on the \textit{corrupted} input, as they do not carry the relevant signal.
Formally, given a candidate circuit $E_C$ as defined in Def. \autoref{def:circuit}, the patched input to $v$
given corrupted input $\tilde{x}$ is
\begin{equation}
\mathrm{in}_v^{C}(\tilde{x})
=
\sum_{u \in \mathcal{P}(v)}
\left(
i_e\, r_u(x)
+
(1 - i_e)\, r_u(\tilde{x})
\right),
\end{equation}
where $i_e = \mathbf{1}_{(u \to v) \in E_C}$ is an indicator
that equals $1$ if the edge $(u \to v)$ belongs to the circuit
and $0$ otherwise. We adapt the sequential pruning approach of \cite{conmy2023towards}, as showcased on the right pane of \autoref{fig:circuits_overview} using target logit difference as the selection criterion for individual edges. If the loss after removing an edge is small enough, we remove the edge from the \textit{candidate circuit }and update the model. After circuit extraction, we evaluate faithfulness in terms of classification accuracy.

\subsection{Discovering Faithful Visual Circuits at Scale.}\label{sec:method:graph_reduction}

\begin{wrapfigure}[24]{r}{0.5\textwidth}
  \vspace{-2em}
  \centering
  \includegraphics[width=\linewidth]{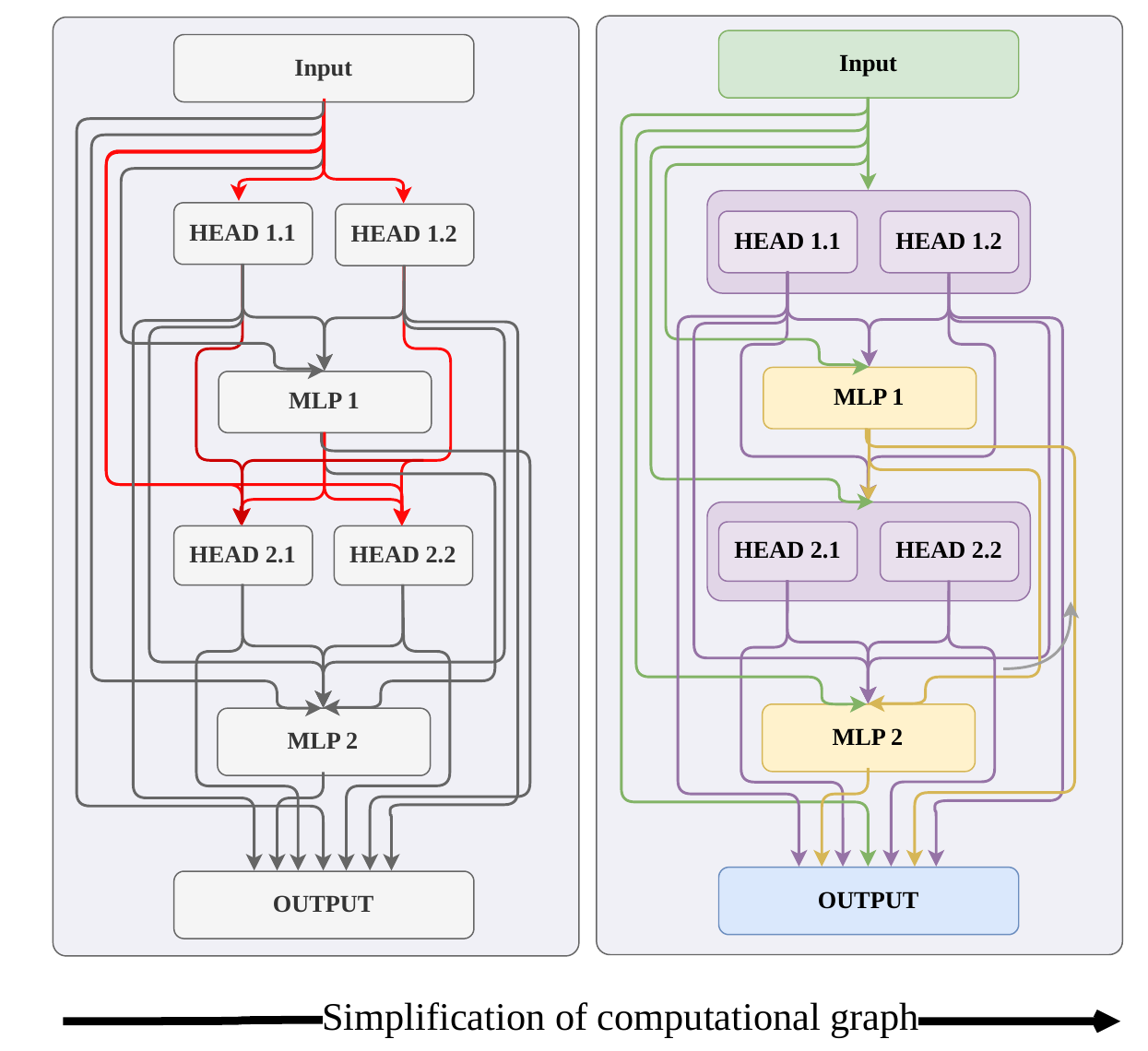}
  \caption{\textbf{Transformer circuitry in a 2-layer toy transformer.}
\textbf{Left:} \textbf{\textcolor{Red}{Red}} edges are simplified in \ours for scalability: multiple attention-head receiver nodes are collapsed into a single attention-input node.
\textbf{Right:} \textbf{\textcolor{ForestGreen}{Green}} edges correspond to \emph{input} the sender node,
\textbf{\textcolor{Dandelion}{Yellow}} edges correspond to \emph{MLPs} sender nodes, and
\textbf{\textcolor{Orchid}{Purple}} edges correspond to \emph{attention heads} sender nodes.}
  \label{fig:toy_computational_graph}
    \vspace{-2em}
\end{wrapfigure}

A full residual-stream dependency graph for a $L$-layer,
$H$-head transformer contains $O(L^2H^2)$ edges,
making exhaustive circuit search computationally infeasible.
To control complexity while preserving functional dependencies,
we introduce an \emph{attention-input node} for each attention block.
This node aggregates the residual input to the block and acts as a
shared receiver for incoming edges, while attention heads remain
independent senders.
This reduces the number of candidate edges by approximately a factor
of $H$ without modifying any of the functionality. We define the following edge types:
\textcolor{ForestGreen}{input $\to$ attn\_in, input $\to$ mlp, input $\to$ logits},
\textcolor{Orchid}{attn $\to$ attn\_in, attn $\to$ mlp, attn $\to$ logits},
\textcolor{Dandelion}{mlp $\to$ attn\_in, mlp $\to$ mlp, mlp $\to$ logits} (\autoref{fig:toy_computational_graph} depicts a visualization for a toy 2-layer transformer).

\subsection{Dataset and Task Definition.}\label{section:method:dataset}
Because vision transformers operate over spatially distributed patch
embeddings rather than discrete tokens, corruption cannot be localized
to a single representation unit. We therefore define corruption via
segmentation of the foreground object associated with the target task ( e.g. classification of a class)
followed by inpainting of the masked region. The segmentation mask
removes task-specific evidence in the foreground, and the inpainting model fills the
region to produce a visually plausible image while preserving global
background statistics and low-level structure. 

We use the ForAug dataset~\cite{nauen2025foraug}, which provides
ImageNet images processed using this segmentation-and-inpainting
pipeline for class-specific objects.
This procedure removes object evidence while
minimizing out-of-distribution artifacts, ensuring that performance
degradation reflects the absence of class-specific information rather
than distributional shift.
However, this approach is not limited to classification tasks and can, for example, easily be applied to tasks such as discovery of concept-level circuits using inpainting methods such as presented in SUB~\cite{bader2025sub}.
Below, we also investigate a different setup, where corrupted inputs are given by a harmful attack and we want to steer the model to be more safe.

For ease of evaluation, we here consider classification tasks in which the objective is to distinguish images belonging to class $A$ from those of competing classes, as defined in Def. ~\ref{def:class_circuit}.
In particular, the goal is to ensure that instances of the target class are always classified correctly.
For models such as ViT-B~\cite{dosovitskiy2020image}, logits are obtained directly from the
classification head. For contrastive models such as OpenCLIP~\cite{cherti2023reproducible}, logits are computed
as the dot product between the image embedding and the
corresponding text embedding.

We use target logit difference as the pruning metric, as it provides a scalar objective directly aligned with the model’s decision boundary.
Unlike language models, most vision classifiers do not operate over a fixed token dictionary, making distribution-level comparisons less directly aligned with a classification objective. Ablations comparing target logit-difference and Kullback–Leibler divergence-based pruning metrics
are provided in Appendix \autoref{app:kl_logit_diff_ablations}. A task-specific class circuit is now defined as follows.

\begin{definitionbox}{Class Circuit}{class_circuit}
For a given target class $A$, a \emph{class circuit} is a tuple
\[
\mathcal{C}_A = (\mathcal{D}_A, M, E_{\mathcal{C}})
\]
where $\mathcal{D}_A = \{x_i\}_{i=1}^N$ is a dataset of diverse inputs correctly classified as $A$,
$M$ is a scalar performance metric (\emph{for example \textbf{accuracy}}),
and $E_{\mathcal{C}}$ is a \textbf{faithful circuit}. The circuit is evaluated via activation patching, and task performance under $M$ is largely \textbf{preserved} under the intervention.
\end{definitionbox}

\subsubsection{Making models safe through circuits steering.}
\label{sec:methods:steering}
A prominent application of mechanistic interventions in LLMs is safety, where steering on relevant model components can prevent undesirable behaviors such as
harmful generations \cite{hufe2026dyslexify}.
Our approach naturally lends itself to such
steering applications in settings where attacks introduce corrupted
inputs. 
In typographic attacks~\cite{wang2025typographic}, text is added to an image to make the model predict a class that does not match the image’s visual content.
On such data, we discover circuits that correspond to the attack path, and we define those attack circuits analogously to class circuits Def. \ref{def:class_circuit}.

Let $E_{\mathcal{C}}$ denote a faithful circuit associated
with the attack. From paired corrupted and clean inputs, we estimate
for each circuit component $j$ a corruption-aligned direction $v_j$
as the mean normalized activation difference across examples
(Alg. ~\ref{alg:steering_vector_computation}).

\begin{wrapfigure}{r}{0.5\textwidth}
\vspace{-2em}
\begin{minipage}{0.48\textwidth}
\captionof{algorithm}{Steering Computation}
\begin{algorithmic}[1]
\label{alg:steering_vector_computation}
\FOR{each component $j$}
    \FOR{each datapoint $i \in \mathcal{D}$}
        \STATE $\hat{x}_{A}^{(i,j)} \leftarrow 
        \frac{x_{A}^{(i,j)}}{\|x_{A}^{(i,j)}\|}$
        \STATE $\hat{x}_{\neg A}^{(i,j)} \leftarrow 
        \frac{x_{\neg A}^{(i,j)}}{\|x_{\neg A}^{(i,j)}\|}$
        \STATE $\Delta^{(i,j)} \leftarrow 
        \hat{x}_{A}^{(i,j)} - \hat{x}_{\neg A}^{(i,j)}$
    \ENDFOR
    \STATE $v_j \leftarrow 
    \frac{1}{|\mathcal{D}|}
    \sum_{i \in \mathcal{D}} \Delta^{(i,j)}$
\ENDFOR
\end{algorithmic}
\end{minipage}
\vspace{-1em}
\end{wrapfigure}

During inference, we intervene on sender outputs
$h \in \mathbb{R}^{P \times d}$ at hook locations corresponding to
edges $e \in E_{\mathcal{C}}$, where $P$ denotes the number of
patches and $d$ the hidden dimension. Each component $j$ is associated
with a patchwise direction $v_j \in \mathbb{R}^{P \times d}$.
For each patch $p$, we compute the projection coefficient
\[
c_p \;=\; \frac{\langle h_p, v_{j,p} \rangle}{\|v_{j,p}\|^2 + \varepsilon}
\]
and apply directional ablation
\[
h^{\text{steered}}_p
=
h_p - \alpha \,\mathrm{ReLU}(c_p)\, v_{j,p},
\]
where $\alpha \ge 0$ controls the steering strength.
This intervention corresponds to a \emph{directional ablation}
(projection removal), widely studied in activation steering
\citep{arditi2024refusal}. Geometrically, the operation modifies the
activation within the two-dimensional subspace spanned by the
activation $h$ and the feature direction $v$ \citep{vu2025angular}.
ReLU ensures that only positively aligned projections are removed,
preventing amplification of anti-correlated features.
Interventions are applied only to circuits edges in $E_{\mathcal{C}}$. As
the circuit is faithful, suppressing the corruption-aligned direction
at upstream edges prevents its propagation to downstream components
while leaving unrelated computations largely unaffected.


%% file: sections/experiments.tex
\section{Experimental Evaluations}

In this section, we evaluate the circuits discovered by \ours, analyze their properties, and assess their usefulness in downstream safety-related tasks. 
We first benchmark \ours against existing edge-based circuit discovery methods, measuring how well extracted circuits recover the original model performance. 
We then examine the discovered circuits with respect to class specificity, compositional structure, and robustness across experimental runs. Finally, we investigate downstream applications, evaluating whether circuits can be used for targeted steering to improve robustness against typographic and label-corruption attacks.
Throughout our evaluations we aim to answer following research questions:
\begin{enumerate}[leftmargin=*,label=\textbf{RQ\arabic*.}]
    \item Does \method reliably recover faithful circuits in vision models? (\autoref{sec:exp:benchmark})
    \item Does \method produce computation graphs close in circuit space for similar classes (\autoref{sec:exp:analysis})?
    \item Can discovered circuits be used for targeted steering and improve robustness against typographic and label-corruption attacks? (\autoref{sec:exp:typoattacks},~\autoref{sec:exp:rococores})
\end{enumerate}

\subsection{Experimental Setup}\label{sec:setup}

We evaluate \method{} in the context of \textbf{RQ1} on a supervised ViT-B trained on ImageNet~\cite{dosovitskiy2020image} and OpenCLIP ViT-B/32~\cite{radford2021clip, cherti2023reproducible}.
Each model is represented as a directed computation graph over attention heads and MLP blocks connected via residual-stream edges (\autoref{sec:method}).
We compare against EAP~\cite{syed2024attribution} and EAP-IG~\cite{hanna2024have}, adapted to Vision Transformers using the ViTPrisma library~\cite{joseph2025prisma}, with the same simplified computation graph enforced, as well as random edge pruning as a lower bound.
Circuit quality is measured via \emph{faithfulness} (classification accuracy when computation is restricted to the circuit) and \emph{sparsity} (fraction of edges retained).
For cross-class analysis (\textbf{RQ2}), each circuit is represented as a set of edges in the computation graph, and similarity between circuits is measured via Jaccard similarity ($|A \cap B| / |A \cup B|$).
In the context of \textbf{RQ3}, we evaluate circuit-based steering as a defense against typographic attacks~\cite{wang2025typographic} on OpenCLIP, using three attack regimes: large text overlays, small text overlays, and bezel-style overlays (examples in~\autoref{app:steering}).
Typographic circuits are extracted for 13 effective attack words using paired images, where the typographic image serves as the clean input and its uncorrupted counterpart as the corrupted input.
Performance is measured by the trade-off between zero-shot accuracy retention and attack success rate (ASR).
We additionally evaluate steering on the RoCOCO danger benchmark~\cite{park2024rococo}, where danger-related words are inserted into captions and the goal is to reduce the Recall Score of Manipulated Samples (RSMS).
Full experimental setup details are provided in~\autoref{app:setup}.

\subsection{Benchmarking Circuit Extraction in ViTs}\label{sec:exp:benchmark}
\input{tables/method_comparison_table}

To address \textbf{RQ1}, we extract class-specific circuits for each ImageNet class on both ViT-B and OpenCLIP, and measure classification accuracy as a function of circuit sparsity.
\autoref{fig:method-comp-2x2} reports results for \method{}, EAP, EAP-IG (with step counts 3, 5, and 10), and random edge pruning.

We observe that \method{} vastly outperforms all baselines, \textbf{discovering faithful circuits that are approximately $10\times$ sparser} than comparably accurate circuits found by the strongest baseline, EAP-IG.
On ViT-B, \method{} recovers near-perfect classification accuracy while retaining fewer than 10\% of edges, whereas EAP-IG requires substantially more edges to reach similar performance and EAP remains close to random pruning throughout.
This pattern holds on OpenCLIP, where \method{} again achieves high faithfulness at extreme sparsity levels while baseline methods plateau at low accuracy until a much larger fraction of edges is included.
Among the baselines, EAP-IG consistently outperforms EAP, with only marginal differences across step counts.

These results confirm that \method{} reliably recovers faithful edge-based circuits in vision transformers, producing dramatically more compact circuits than existing methods adapted from the language domain. Let us now move on to investigating circuit-class similarities.



\subsection{Cross-class Similarity}\label{sec:exp:analysis}
\input{tables/ridge_plots}

To address \textbf{RQ2}, we measure pairwise Jaccard similarity between circuits extracted for different object and animal classes to analyze whether circuits encode class-specific or shared computations.
We visualize the resulting similarity distributions using ridge plots (\autoref{fig:ridge_plots}).

We observe several patterns.
First, circuits for semantically related classes exhibit higher overlap than those for unrelated classes.
For instance, circuits extracted for dog breeds such as \emph{Pembroke Corgi} and \emph{Cardigan Corgi} share substantially more edges with each other than with circuits for object categories like \emph{Revolver} or \emph{Tench}.
Conversely, the \emph{Revolver} circuit shows highest similarity with other weapon classes (\emph{Assault Rifle}, \emph{Rifle}) and low overlap with animal classes.
This suggests that the model reuses computational pathways for visually or semantically related categories, consistent with prior findings~\cite{dorszewski2025colors}.

Second, circuits extracted for the same class across different runs are not identical.
While they share a core subset of edges, the recovered circuits differ in additional connections, indicating that the underlying computation is implemented by partially overlapping subgraphs rather than a single fixed circuit.
We explore this phenomenon further in~\autoref{app:stability}, where we investigate whether multiple circuits can encode similar behaviors.

\subsection{Steering Against Typographic Attacks}\label{sec:exp:typoattacks}
\input{tables/small_typographic_corruption}

To address \textbf{RQ3}, we evaluate whether discovered typographic circuits can be steered to defend against typographic attacks.
\autoref{tab:summary_table_circuit_steering_typographic} reports average results across attacks for all three attack regimes.
Circuit steering consistently reduces the attack success rate (ASR) across all settings, with Top-1 ASR dropping from 39.1\% to 2.8\% for large text and from 39.4\% to 1.6\% for small text overlays.
This defense comes at a modest cost to clean accuracy, with the bezel attack showing the largest drop, likely due to its more intrusive visual corruption.

\begin{figure}[t]
    \centering
    \includegraphics[width=1\linewidth]{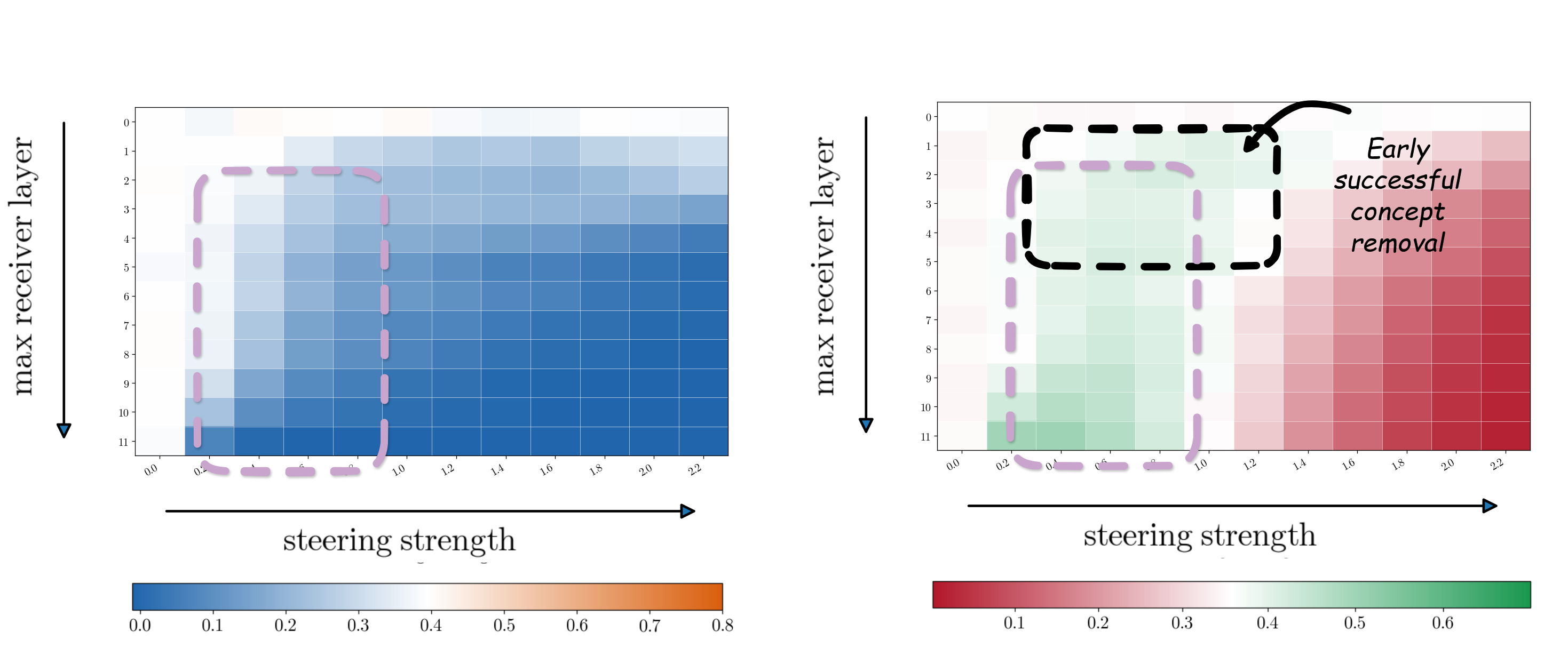}
    \caption{\textbf{Circuit-based steering prevents typographic attacks without harming performance.} \textbf{Left: Attack success rate ($\downarrow$ lower is better). Right: performance retention on accuracy ($\uparrow$ higher is better)}. We show metrics against steering strength and the maximal receiver layer in the circuit. \textcolor{Orchid}{Purple} region indicates regions where steering is successfully applied. See results for random circuits in \autoref{app:max_receiver_full}.
    }
    \label{fig:edge_distribution}
\end{figure}

\autoref{fig:edge_distribution} shows ASR and accuracy retention as a function of steering strength and the maximal receiver layer included in the circuit.
Notably, even early-layer interventions successfully lower the attack success rate while preserving performance, suggesting that typographic information is routed through identifiable pathways already in the earlier layers of the model.
This pattern is not observed for random circuits of the same size (\autoref{app:max_receiver_full}), confirming that the effect is specific to the discovered circuits rather than an artifact of general ablation.
Full per-attack results are reported in~\autoref{tab:big_circuit_stering_typographic} and~\autoref{fig:facets_multiple_plots_normed_mean_steering}.


\subsection{Steering on the RoCOCO Danger Benchmark}\label{sec:exp:rococores}

\input{tables/rococo_steering}

We further investigate \textbf{RQ3} and validate circuit-based steering on the RoCOCO benchmark~\cite{park2024rococo}, where danger-related words are inserted into captions to manipulate retrieval.
Using weapon-related circuits (Assault Rifle, Revolver, Rifle) with a steering strength of 0.4 (\autoref{sec:methods:steering}), we report retrieval performance and safety metrics in~\autoref{tab:rococo_steering}. 

Steering consistently reduces the RSMS, with all three circuits lowering it from 11.68\% to around 5\%, more than halving the model's susceptibility to manipulated danger cues.
Crucially, this safety improvement comes without sacrificing retrieval quality: steered models maintain or slightly improve recall across all $k$ levels, with the Rifle circuit even boosting $R_\text{mean}$ from 62.74\% to 64.37\%.
This suggests that the discovered circuits specifically capture the vulnerability pathway exploited by the attack, and that deactivating it does not degrade the model's general retrieval capabilities.
Full specifications are provided in~\autoref{app:steering:rococo}.

Overall, we answer \textbf{RQ3} affirmatively: discovered circuits can be leveraged for targeted steering that measurably improves model robustness, highlighting that faithful edge-based circuits are not only interpretable but actionable.

%% file: tables/method_comparison_table.tex
\begin{figure}[t]
    \centering
    \renewcommand{\arraystretch}{1.0}
    \newcommand{\panelw}{0.47\textwidth}

    \begin{tabular}{cc}
        \multicolumn{1}{c}{\footnotesize\textsc{OpenCLIP}} &
        \multicolumn{1}{c}{\footnotesize\textsc{ViT-B}} \\
        \cmidrule(lr){1-1}\cmidrule(lr){2-2}

        \includegraphics[width=\panelw]{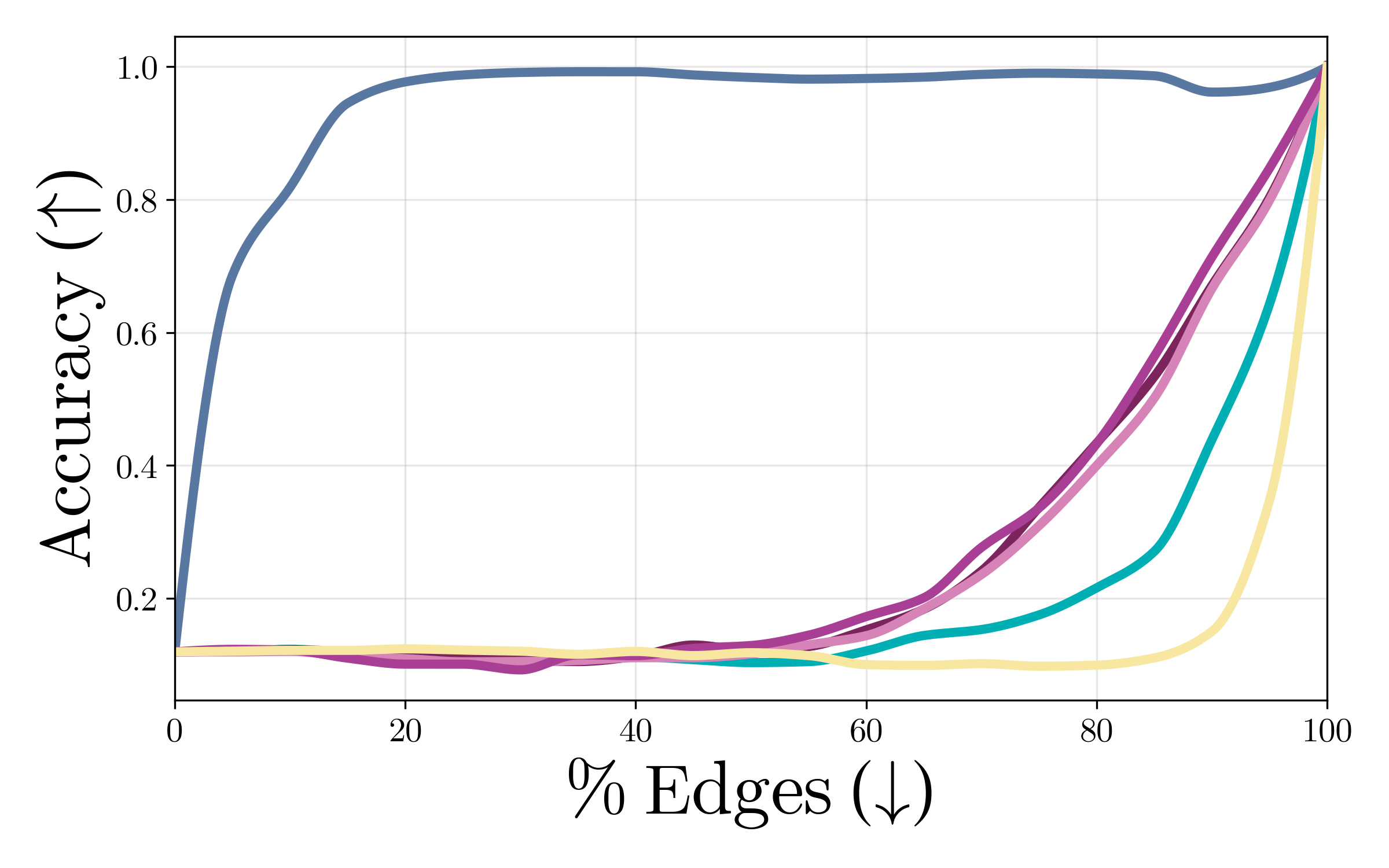}
        &
        \includegraphics[width=\panelw]{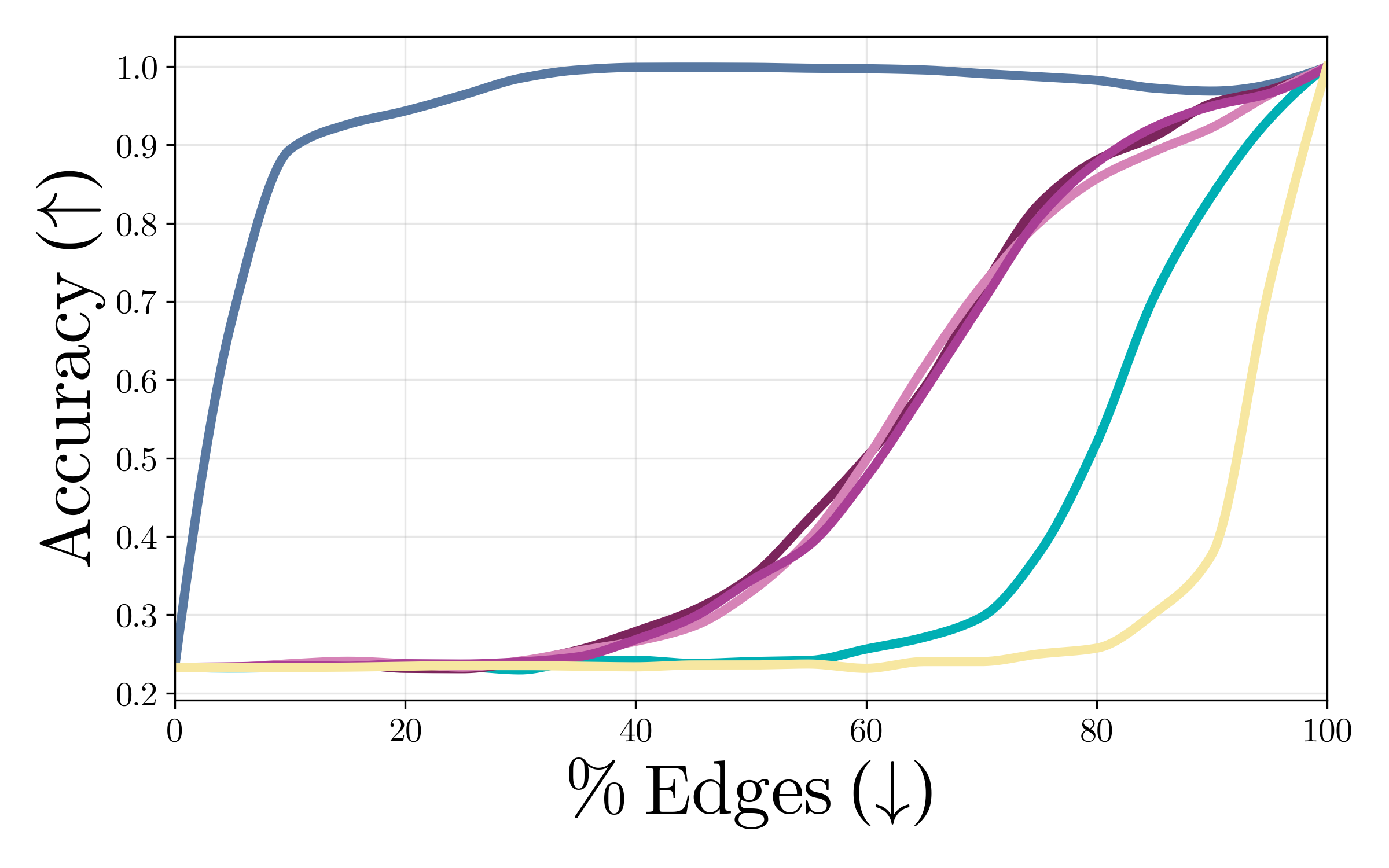}
        \\[-0.3em]

    \end{tabular}

    \vspace{0.4em}

    {\footnotesize
    \legenditem{MidnightBlue}{\method(ours)  }
    \legenditem{TealBlue}{EAP  }
    \legenditem{Sepia}{EAP-IG-10  }
    \legenditem{Mulberry}{EAP-IG-5  }
    \legenditem{Thistle}{EAP-IG-3  }
    \legenditem{LightGoldenrod}{Random}
    }

    \caption{\textbf{\ours finds 10x sparser circuits.} 
    We report accuracy of the circuit on the target class ($\uparrow$ higher is better) as faithfulness and report different sparsity levels for circuits as edges remaining ($\downarrow$ lower is better) for different circuit extraction methods indicated by colors. We compare linear probe classification performance of (ViT-B)OpenCLIP and of a supervised ViT-B on Imagenet data.}
    \label{fig:method-comp-2x2}
\end{figure}

%% file: tables/ridge_plots.tex
\begin{figure}[t!]
\centering

\begin{subfigure}[t]{0.48\textwidth}
	\centering
	\includegraphics[width=\linewidth]{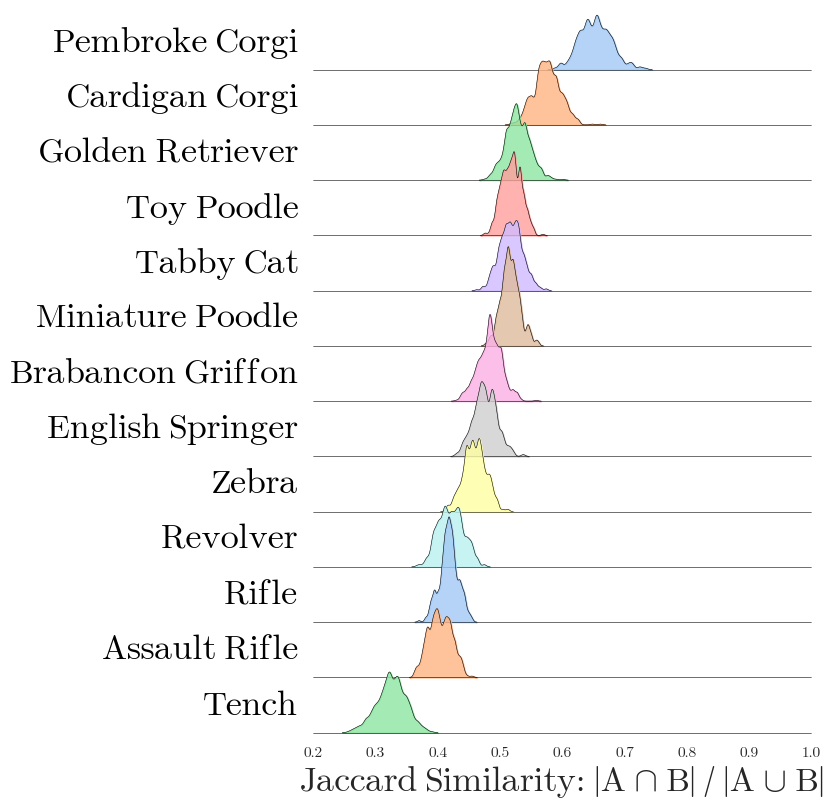}
\end{subfigure}\hfill
\begin{subfigure}[t]{0.48\textwidth}
	\centering
	\includegraphics[width=\linewidth]{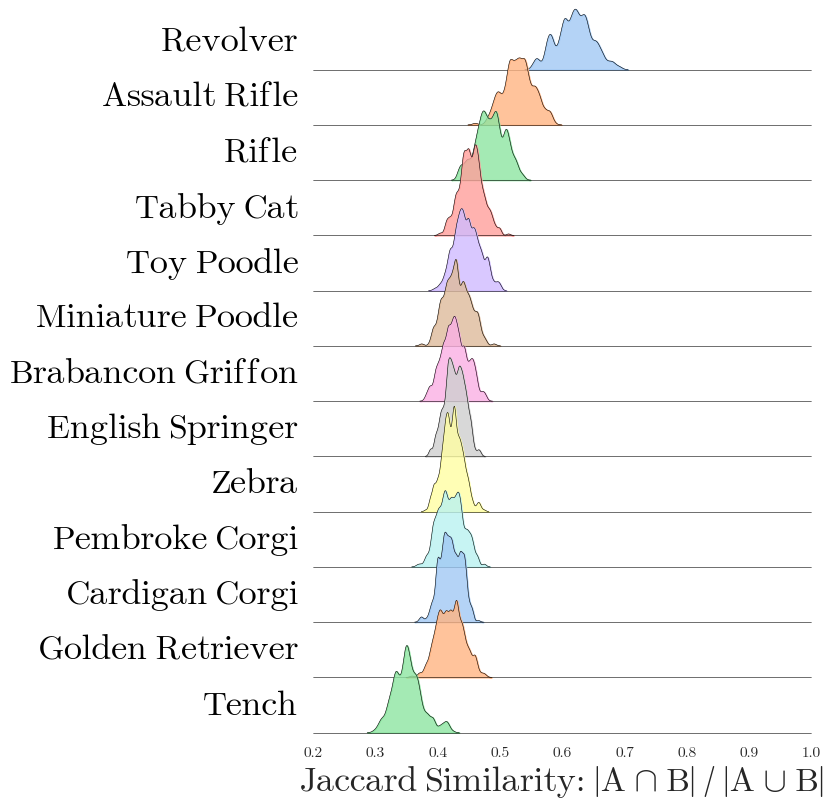}
\end{subfigure}

\caption{\textbf{\ours discovers circuits reflecting semantic similarity of classes.} Pairwise Jaccard similarity ($|A \cap B| / |A \cup B|$) of example class circuits of class \textit{Pembroke Corgi} (left) and \textit{Revolver} (right) against other classes in OpenCLIP.
}
\label{fig:ridge_plots}
\end{figure}





%% file: tables/small_typographic_corruption.tex

\begin{table}[t!]
\centering
\caption{\textbf{Typographic attacks on images.} Average performance of circuit steering against typographic attacks on ImageNet (unsteered base vs.\ steered with Attack Success Rate $\downarrow$90\%).
Circuit-steering defenses are succesful, greatly reducing attack success with only marginal loss in performance on clean images.}
\begin{minipage}{0.9\linewidth}
\label{tab:summary_table_circuit_steering_typographic}
\setlength{\tabcolsep}{5pt}
\resizebox{\linewidth}{!}{%
\begin{tabular}{@{}ll  cc cc cc@{}}
\toprule
 &  & \multicolumn{2}{c}{\textbf{Big Text}} & \multicolumn{2}{c}{\textbf{Small Text}} & \multicolumn{2}{c}{\textbf{Bezel}} \\
\cmidrule(lr){3-4} \cmidrule(lr){5-6} \cmidrule(lr){7-8}
\textbf{Setting} & \textbf{Metric} & Base & Steered & Base & Steered & Base & Steered \\
\midrule
\multirow{2}{*}{Clean ImageNet}
 & Top-1 Acc.\ (\%) & \textbf{57.0} & 55.8 & \textbf{57.1} & 55.7 & \textbf{57.1} & 49.9 \\
 & Top-5 Acc.\ (\%) & \textbf{81.9} & 80.9 & \textbf{81.9} & 80.9 & \textbf{81.9} & 75.6 \\
\midrule
\multirow{4}{*}{\shortstack[l]{Corrupted\\ImageNet}}
 & Top-1 Acc.\ (\%) & 34.7 & \textbf{50.0} & 34.7 & \textbf{49.1} & 34.4 & \textbf{45.0} \\
 & Top-5 Acc.\ (\%) & 69.9 & \textbf{75.9} & 70.1 & \textbf{75.3} & 69.8 & \textbf{71.2} \\
 & ASR Top-1 (\%) & 39.1 & \textbf{2.8} & 39.4 & \textbf{1.6} & 39.5 & \textbf{3.1} \\
 & ASR Top-5 (\%) & 68.0 & \textbf{8.1} & 68.4 & \textbf{7.6} & 68.4 & \textbf{12.7} \\
\bottomrule
\end{tabular}%
}
\end{minipage}
\end{table}

%% file: tables/rococo_steering.tex
\begin{table}[t]
\centering
\small
\caption{%
  \textbf{RoCOCO retrieval under circuit-based steering} ($\alpha=0.4$). 
  \textbf{R@$k$}: fraction of queries where the correct item ranks in the top~$k$ ($\uparrow$higher is better). 
  \textbf{R\textsubscript{mean}}: mean of R@1/5/10 ($\uparrow$). 
  \textbf{RSMS}~\cite{park2024rococo}: R@$k$ on adversarially manipulated items, measuring retrieval of danger-class images for benign queries ($\downarrow$ lower is better). 
  Mean~$\pm$~std over three seeds.
}
\label{tab:rococo_steering}
\setlength{\tabcolsep}{6pt}
\resizebox{0.9\linewidth}{!}{%
\begin{tabular}{@{}ll cccc@{}}
\toprule
 &  & \textbf{Base} & \textbf{Assault Rifle} & \textbf{Revolver} & \textbf{Rifle} \\
\midrule
\multirow{3}{*}{Recall}
 & R@1 (\%)  & 40.42 & \textbf{42.36 $\pm$ 0.42} & \textbf{42.01 $\pm$ 0.36} & \textbf{43.47 $\pm$ 0.48} \\
 & R@5 (\%)  & 69.48 & 68.59 $\pm$ 0.46 & 68.39 $\pm$ 0.24 & \textbf{70.22 $\pm$ 0.61} \\
 & R@10 (\%) & 78.32 & 78.04 $\pm$ 0.38 & 77.65 $\pm$ 0.21 & \textbf{79.41 $\pm$ 0.43} \\
\midrule
\multirow{1}{*}{Aggregate}
 & R\textsubscript{mean} (\%) & 62.74 & \textbf{63.00 $\pm$ 0.37} & 62.68 $\pm$ 0.22 & \textbf{64.37 $\pm$ 0.47} \\
\midrule
\multirow{1}{*}{Safety}
 & RSMS (\%) & 11.68 & \textbf{5.15 $\pm$ 0.24} & \textbf{5.69 $\pm$ 0.41} & \textbf{4.86 $\pm$ 0.21} \\
\bottomrule
\end{tabular}%
}
\end{table}

%% file: sections/future_work.tex
\section{Discussion}

Our results demonstrate that faithful, edge-based circuits can be recovered in large-scale vision transformers, extending a paradigm previously focused on language models.
With our suggested approach \ours we discover up to $10\times$ sparser circuits compared to existing approaches while preserving task performance, confirming that \textit{specific model behaviors can be attributed to compact subgraphs rather than distributed computation} across the entire network.
Importantly, these circuits are not merely descriptive: steering experiments reduce typographic attack success rates by more than 90\% and halve safety violations on the RoCOCO benchmark, without degrading general model performance.
This serves as causal validation of our circuits: the predictable effects of steering confirm that the \textit{recovered subgraphs capture genuine mechanisms rather than statistical artifacts}.

We further find evidence of compositional structure: unions of class-specific circuits support zero-shot binary classification without additional training, suggesting that the model organizes computation into separable mechanisms rather than entangled representations (\autoref{app:binary_classification}).
At the same time, circuits recovered for the same class from different samples of the same distribution show moderate but imperfect overlap (Jaccard similarities of 0.6--0.8), yet show distinctly higher similarity to class-circuits than non-class circuits (\autoref{fig:ridge_plots}).
Rather than a failure of the method, we interpret this as reflecting genuine redundancy in the model: multiple partially overlapping subgraphs can implement similar behaviors, consistent with OR-circuit structure observed in language models~\cite{mondorf-etal-2025-circuit}.
This suggests that circuit discovery may be better understood as producing a \emph{distribution} over plausible circuits rather than a single ground-truth subgraph (\autoref{app:stability}). 

Regarding our corruption strategy: we remove class-specific foreground content via segmentation and inpainting while preserving the background, so that edges encoding spurious background correlations carry no activation difference during patching. This encourages the discovered circuits to reflect object-level pathways rather than background artifacts.
A practical limitation of our approach is computational cost: compared to approximate methods such as EAP and EAP-IG, \method{} requires patching each edge at least once. While we already scale to medium-sized foundation models such as OpenCLIP, making efficient discovery in the largest available transformers is an important direction for future work.

Looking forward, several directions may extend this work. Combining edge-level \textit{and} neuron-level analyses, identifying which neurons within a circuit edge carry the most semantically interpretable information, would yield richer mechanistic descriptions than either approach provides alone. Extending circuit discovery beyond classification to tasks such as grounding or retrieval remains largely unexplored. Additionally, circuits identified for specific objects or concepts could be leveraged to detect outliers or anomalous inputs, providing another application of circuit-based interpretability in vision systems. 
Additionally, understanding how circuit structure evolves over the course of training could shed light on when and how models acquire specific capabilities.
Relatedly, investigating the connection between circuit structure and model robustness under distribution shift may open new avenues for diagnosing and correcting model failures.

%% file: sections/conclusions.tex
\section{Conclusion}

In this work, we introduced \method{}, for the first time bringing circuit discovery on computation graphs to vision transformers.
Our approach recovers sparse, faithful computational subgraphs that capture class-specific behavior using a fraction of the model's edges, substantially outperforming gradient-based approximations adapted from the language domain.
The discovered circuits exhibit meaningful semantic structure, with related classes sharing significantly more computational pathways than unrelated ones, revealing how vision transformers organize their internal representations.
Crucially, our circuits are not only interpretable but actionable: circuit-based steering effectively defends against typographic attacks and reduces susceptibility to caption manipulation on retrieval benchmarks, in both cases with minimal clean accuracy cost.
These results suggest that edge-based mechanistic interpretability, previously confined to language models, offers a promising and practical path toward understanding and controlling large-scale vision systems.

%% file: sections/appendix_steering.tex
\section{Mining class Circuits}
\label{app:class_circuits}

\paragraph{Dataset.}
Class circuits are mined using the ForAug dataset~\cite{nauen2025foraug}, 
which provides ImageNet images processed via a segmentation-and-inpainting 
pipeline. For each target class, foreground objects are segmented and 
inpainted to produce corrupted counterparts that remove class-specific object
evidence while preserving background statistics and low-level structure, see examples in \autoref{tab:example_class_circuits_points}. 
For each class circuit, we use $N = 128$ images drawn from the corresponding 
ImageNet class. Images are filtered to retain only those correctly 
classified by the base model prior to circuit extraction.
\input{tables/examples_datapoints_class_circuits}

\paragraph{Circuit Extraction.}
Circuits are extracted using \method{} with the sequential activation 
patching procedure described in \autoref{sec:method}. We use the 
\emph{target logit difference} as the pruning criterion, as it provides 
a scalar objective aligned with the model's decision boundary. 
The maximum number of visited nodes per run is set to $900$. 
\input{tables/class_circuits_thresholds}
We have defined multiple selection thresholds, see \autoref{tab:thr_class_circuits} for details. We optimize for performance retention and maximal sparsity, setting the thresholds to achieve over 80\% performance retention. Edges with a target logit difference below 
this threshold are included in the circuit; edges above are pruned.

\section{Experimental Setup}
\label{app:setup}
\paragraph{Models.}
We evaluate \method{} on two Vision Transformer architectures:
a supervised ViT-B trained on ImageNet~\cite{dosovitskiy2020image} and OpenCLIP ViT-B/32~\cite{radford2021clip, cherti2023reproducible} (\texttt{open-clip: laion/CLIP-ViT-B-32-DataComp.XL-s13B-b90K}).
The former represents a standard classification model, while the latter enables us to study circuits in a contrastive vision-language setting.
For OpenCLIP-based classification, we use the zero-shot ImageNet classifier matrix derived from text embeddings provided in the ViTPrisma library~\cite{joseph2025prisma}.
\paragraph{Tasks.}
We define three circuit discovery tasks of increasing complexity:
(i)~\emph{class circuits}, which capture the computation responsible for recognizing a specific ImageNet class,
(ii)~\emph{typographic attack circuits}, which identify the mechanism by which typographic perturbations alter CLIP predictions, and 
(iii)~\emph{binary decision circuits}, which isolate the computation distinguishing between a pair of classes, 
Class circuits are extracted for each Imagenette class and attack circuits are described in detail in their respective sections (\autoref{sec:exp:analysis},~\autoref{sec:exp:typoattacks}).

\paragraph{Computational graph.}
We represent each model as a directed computation graph over attention heads and MLP blocks connected via residual-stream edges, following the formulation in~\autoref{sec:method}.
For ease of application, we introduce an explicit residual input node for attention heads, as described in~\autoref{sec:method:graph_reduction}.
All circuit discovery methods operate on the same graph structure to ensure a fair comparison.

\paragraph{Baselines.}
We compare \method{} against adapted approximate methods like Edge Attribution Patching (EAP)~\cite{syed2024attribution} and EAP with Integrated Gradients (EAP-IG)~\cite{hanna2024have}, which we adapt to Vision Transformers building on the ViTPrisma library~\cite{joseph2025prisma}.
For EAP-IG, we report results across multiple step counts (3, 5, 10).
We additionally include random edge pruning as a lower bound.

\paragraph{Evaluation metrics.}
We measure circuit quality along two axes.
\emph{Faithfulness} is assessed by classification accuracy of the extracted circuit on the target task: we restrict computation to only the circuit edges and measure whether the model's predictions are preserved.
We use the target logit as the attribution metric for gradient-based methods.
\emph{Sparsity} is reported as the fraction of edges retained relative to the full computational graph, where lower is better.

\section{Typographic Attacks: Steering using Faithful Circuits}
\label{app:steering}

\subsection{Overview}

We study \emph{activation steering} as a defense against typographic
corruptions by explicitly estimating and subtracting corruption-induced
directions in representation space. Steering vectors are derived from
\emph{faithful circuits}, i.e., subsets of the model's edges between
components previously identified as causally relevant for text sensitivity and
adversarial behavior.

\subsection{Data and Typographic Corruptions}

For each typographic corruption type (e.g., Bezel~\citep{wang2025typographic},
Multiple Small Texts~\citep{wang2025typographic}, Big Text on Image), we
construct paired datasets consisting of:
\begin{itemize}
    \item \textbf{Clean images} without text overlays.
    \item \textbf{Corrupted images} generated from \textit{clean images} using controlled text-placement regimes.
\end{itemize}

To ensure fair evaluation across corruption types, classification is performed
using an ImageNet classifier head that encodes the \emph{full ImageNet label
vocabulary}, rather than a reduced or short-label subset. Importantly, in the \emph{Big
Text on Image} regime, text overlays are placed so as not to fully occlude the
main object, ensuring that failures cannot be trivially attributed to object
occlusion. Qualitative examples of all corruption regimes are shown in ~\autoref{fig:demo_corruption_altair}.

\input{tables/typographic_corruption_types}

The words used for the typographic attacks were selected from a subset of the
ImageNet classes, sourced from previous work \cite{wang2025typographic}, we measure the effectiveness of the attacks in ~\autoref{table:attack_success_rate}.
\input{tables/attack_success_rate}

\subsection{Model Instrumentation and Representation Collection}

We use a CLIP ViT-B/32 vision transformer initialized from the OpenCLIP
implementation \texttt{open-clip: laion/CLIP-ViT-B-32-DataComp.XL-s13B-b90K}
\citep{radford2021clip,cherti2023reproducible}. For each hook location $h$ (defined by sender location), we
collect representations for both clean and corrupted inputs:
\[
\{\, \mathbf{r}^{\text{text}}_{i,h},\; \mathbf{r}^{\text{no-text}}_{i,h} \,\}_{}^N
\]
across 10 batches of size 16 with clean and corrupted pairs, with $N=50$ (number of patches) for
the CLIP model.

\subsection{Steering Vector Construction and Aggregation}

For each hook location, we compute a per-sample difference vector:
\[
\Delta_i = \mathbf{r}^{\text{text}}_i - \mathbf{r}^{\text{no-text}}_i.
\]
We consider two normalization regimes:

\paragraph{Pre-Normed Steering.}
Representations are first $\ell_2$-normalized per sample:
\[
\hat{\mathbf{r}} = \frac{\mathbf{r}}{\|\mathbf{r}\|_2}, \quad
\Delta^{\text{pre}}_i = \hat{\mathbf{r}}^{\text{text}}_i - \hat{\mathbf{r}}^{\text{no-text}}_i.
\]

\paragraph{Post-Normed Steering.}
The difference is taken first and normalized afterward:
\[
\Delta^{\text{post}}_i =
\frac{\mathbf{r}^{\text{text}}_i - \mathbf{r}^{\text{no-text}}_i}
{\|\mathbf{r}^{\text{text}}_i - \mathbf{r}^{\text{no-text}}_i\|_2}.
\]

\paragraph{Aggregation across samples.}
To summarize corruption effects, we aggregate per-sample steering vectors
using \textbf{mean} and \textbf{medoid} estimators. For multi-head activations
(e.g., attention outputs), medoids are computed \emph{per head and per
position} to preserve head-specific structure.

\begin{table}[h!]
\centering
\caption{Steering Vector Aggregation: Pre- vs Post-Normalization and Mean vs Medoid.}
\label{tab:steering_summary}
\begin{tabular}{c|c|c}
\toprule
 & \textbf{Mean} & \textbf{Medoid} \\
\midrule
\textbf{Pre-Normed}  & Average of $\Delta_i^{\text{pre}}$  & Most representative $\Delta_i^{\text{pre}}$  \\
\textbf{Post-Normed} & Average of $\Delta_i^{\text{post}}$ & Most representative $\Delta_i^{\text{post}}$ \\
\bottomrule
\end{tabular}
\end{table}

Formally, given a set of sample-level difference vectors $\{\Delta_i\}_{i=1}^N$,
the medoid $\Delta^{\text{medoid}}$ is defined as the vector that minimizes the
sum of pairwise cosine distances to all other vectors:
\[
\Delta^{\text{medoid}} = \arg\max_{\Delta_i} \sum_{j=1}^N \cos(\Delta_i, \Delta_j),
\]
where $\cos(\Delta_i, \Delta_j)$ denotes the cosine similarity between vectors
$\Delta_i$ and $\Delta_j$. Steering can successfully be performed in every
listed regime, with the best results observed under pre-normalization. We
conduct all subsequent experiments using the pre-normed mean regime (see
\autoref{tab:steering_summary} for definitions; results in
\autoref{fig:orange_steering}).

\input{tables/steering_types_comparison}
\subsection{Applying Steering}
\label{app:steering:applying}

Given a steering vector $\mathbf{s}_h$ at hook $h$, we apply steering 
by intervening on sender outputs $h \in \mathbb{R}^{P \times d}$ at 
hook locations corresponding to edges $e \in E_{\mathcal{C}}$, where 
$P$ denotes the number of patches and $d$ the hidden dimension. Each 
component $j$ is associated with a patchwise direction 
$v_j \in \mathbb{R}^{P \times d}$. For each patch $p$, we compute 
the projection coefficient
\[
c_p \;=\; \frac{\langle h_p,\, v_{j,p} \rangle}{\|v_{j,p}\|^2 + \varepsilon}
\]
and apply directional ablation
\[
h^{\text{steered}}_p \;=\; h_p - \alpha\,\mathrm{ReLU}(c_p)\,v_{j,p},
\]
where $\alpha \geq 0$ is a scalar steering strength. 
We sweep $\alpha$ over a range of values to study the trade-off 
between corruption suppression and clean accuracy degradation 
(\autoref{tab:steering_summary}).

\input{tables/defence_steering_big_and_small_multiple_text}
\input{tables/facets_multiple_plots_normed_mean_steering}

\subsection{RoCOCO Experiments}
\label{app:steering:rococo}
\paragraph{Benchmark and Task.}
We evaluate circuit-based steering on the RoCOCO 
benchmark~\cite{park2024rococo}, which stress-tests 
image-text matching models on the MS-COCO dataset. 
The benchmark targets the \emph{image-to-text retrieval} 
task: given a query image, the model must retrieve the 
correct caption from a pool of candidates. To probe 
robustness, RoCOCO injects danger-related words (e.g., 
weapon names) into captions, creating adversarially 
manipulated caption sets. A vulnerable model will 
incorrectly retrieve these danger-class captions 
in response to benign image queries, because the 
image embedding inadvertently encodes residual 
danger-related concepts that inflate its similarity 
to the manipulated captions. 

We address this by applying circuit-based steering 
directly to the \emph{image representations}: by 
suppressing danger-aligned directions along the 
edges of the discovered weapon circuits, we remove 
residual danger concepts from the image embedding 
before it is matched against captions. This prevents 
the manipulated captions from being spuriously 
retrieved, without modifying the text encoder or 
the captions themselves. As a control, we apply 
the same steering procedure along random edges 
of matched size to confirm that the effect is 
specific to the mechanistically discovered circuits 
rather than a consequence of general ablation. 
Safety is measured via the Recall Score of 
Manipulated Samples (RSMS), the fraction of 
benign image queries for which a manipulated 
danger caption is incorrectly ranked in the 
top-$k$ results ($\downarrow$ lower is better), alongside standard retrieval metrics R@$k$ 
and R$_\text{mean}$ ($\uparrow$ higher is better).

\paragraph{Results.}
~\autoref{tab:rococo_steering} reports retrieval 
and safety metrics at a fixed steering strength 
of $\alpha = 0.4$ for all three weapon circuits 
and their random counterparts. Circuit-based 
steering consistently and substantially reduces 
RSMS across all three weapons, lowering it from 
a baseline of $11.68\%$ to approximately $5\%$, 
more than halving the model's susceptibility to 
danger-cue manipulation.

\input{tables/rococo_full_table}

In contrast, random baseline circuits of matched 
size also reduce RSMS to approximately $9\%$, 
suggesting that ablating any set of edges has 
some general suppressive effect on danger-concept 
encoding. However, the substantially stronger 
reduction achieved by the mechanistically 
discovered circuits indicates that the weapon 
circuits concentrate the danger-aligned computation 
more precisely than would be expected by chance, 
and that targeted circuit-based steering is a 
more effective and efficient intervention than 
random ablation of equivalent size.

\begin{figure}[H]
    \centering
    \includegraphics[width=0.9\linewidth]{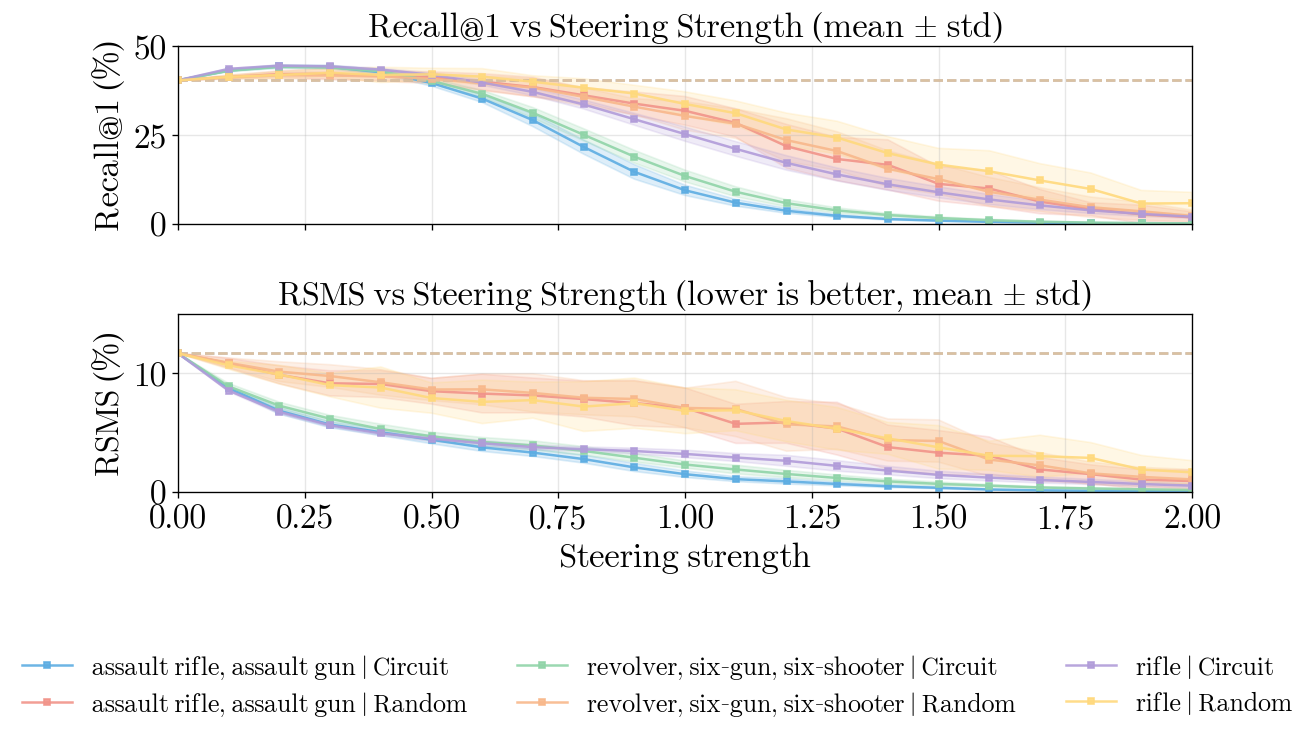}
    \caption{\textbf{RoCOCO steering as a function of steering 
    strength $\alpha$.} Top: R@1 retrieval performance 
    ($\uparrow$ higher is better). Bottom: RSMS 
    ($\downarrow$ lower is better). Solid lines show 
    steering applied along discovered circuit edges; 
    dashed lines show steering along random edges of 
    matched size. Shaded regions denote mean~$\pm$~std 
    over three seeds.}
    \label{fig:rococo_plots}
\end{figure}

\paragraph{Effect of Steering Strength.}
\autoref{fig:rococo_plots} shows how RSMS 
and R@1 evolve as a function of steering 
strength $\alpha$ for all three circuit and 
random baseline conditions. Circuit-based 
steering reduces RSMS steeply even at low 
$\alpha$, while Recall@1 remains largely stable 
across a wide range of steering strengths, 
indicating a favorable safety-utility 
trade-off. The random baseline exhibits a 
markedly weaker and noisier RSMS reduction 
across the same range of $\alpha$, further 
confirming that the circuit edges encode 
danger-related concepts more precisely than 
an arbitrary edge set of the same size.

%% file: tables/examples_datapoints_class_circuits.tex
\begin{table}[htbp]
\centering
\setlength{\tabcolsep}{4pt} 
\begin{tabular}{m{0.08\linewidth} 
                >{\centering\arraybackslash}m{0.2\linewidth} 
                >{\centering\arraybackslash}m{0.2\linewidth} 
                >{\centering\arraybackslash}m{0.2\linewidth} 
                >{\centering\arraybackslash}m{0.2\linewidth}}
\toprule
 & \textbf{Church} & \textbf{Cassette Player} & \textbf{English Springer} & \textbf{Gas Pump}\\
\midrule
\textbf{Clean} &
\includegraphics[width=\linewidth]{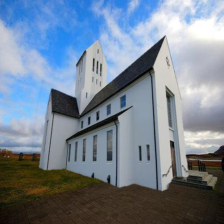} &
\includegraphics[width=\linewidth]{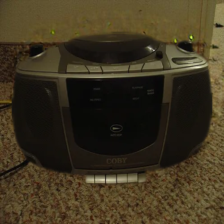} &
\includegraphics[width=\linewidth]{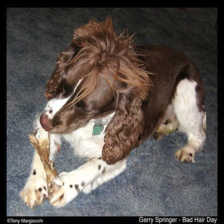} &
\includegraphics[width=\linewidth]{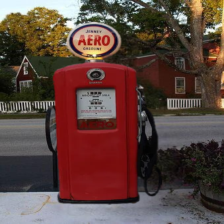} \\[4pt]
\textbf{Corrupted} &
\includegraphics[width=\linewidth]{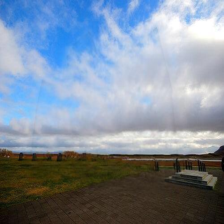} &
\includegraphics[width=\linewidth]{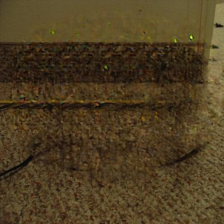} &
\includegraphics[width=\linewidth]{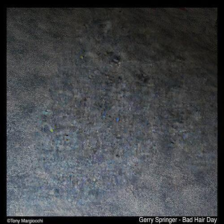} &
\includegraphics[width=\linewidth]{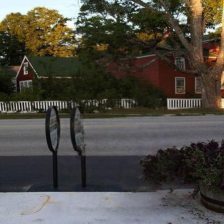} \\
\bottomrule
\end{tabular}
\caption{\textbf{Example data used for class circuits.} Rows show clean and corrupted inputs, while columns correspond to different ImageNet classes. Corrupted inputs remove object-level signal.}
\label{tab:example_class_circuits_points}
\end{table}

%% file: tables/class_circuits_thresholds.tex
\begin{table}[t]
\centering
\caption{Per-class pruning thresholds $\delta$ used for circuit extraction in OpenCLIP and ViT-B.}
\label{tab:thresholds}
\scriptsize
\setlength{\tabcolsep}{4pt}
\renewcommand{\arraystretch}{0.9}
\begin{tabular}{lrr@{\hspace{12pt}}lrr}
\toprule
\textbf{Class} & \textbf{OpenCLIP} $\delta$ & \textbf{ViT-B} $\delta$ &
\textbf{Class} & \textbf{OpenCLIP} $\delta$ & \textbf{ViT-B} $\delta$ \\
\midrule
tench             & 1.00e-3 & 5.11e-2 & zebra             & 4.57e-4 & 5.51e-2 \\
English springer  & 7.29e-4 & 5.72e-2 & tabby cat         & 4.57e-4 & 2.05e-2 \\
cassette player   & 7.29e-4 & 2.86e-2 & golden retriever  & 4.57e-4 & 3.48e-2 \\
chain saw         & 6.96e-4 & 2.66e-2 & Brabancon griffon & 4.57e-4 & 5.11e-2 \\
church            & 6.61e-4 & 4.50e-2 & Pembroke corgi    & 4.57e-4 & 6.13e-2 \\
French horn       & 7.96e-4 & 3.48e-2 & Cardigan corgi    & 4.57e-4 & 3.68e-2 \\
garbage truck     & 6.61e-4 & 7.15e-2 & toy poodle        & 4.57e-4 & 3.88e-2 \\
gas pump          & 7.96e-4 & 6.33e-2 & miniature poodle  & 4.57e-4 & 1.64e-2 \\
golf ball         & 4.57e-4 & 5.31e-2 & standard poodle   & 4.57e-4 & 6.33e-2 \\
parachute         & 4.57e-4 & 4.50e-2 & Mexican hairless  & 4.57e-4 & 3.88e-2 \\
\bottomrule
\label{tab:thr_class_circuits}
\end{tabular}
\end{table}

%% file: tables/typographic_corruption_types.tex
\begin{figure*}[t]
\centering

\begin{minipage}[t]{0.32\linewidth}
\centering
\textbf{Bezel Corruption}\\[-0.2em]
{\footnotesize Text placed along the image border, simulating UI or frame-like overlays.}\\[0.4em]
\fbox{\includegraphics[width=\linewidth]{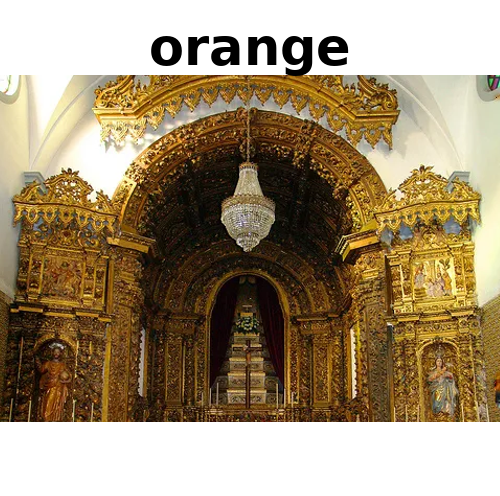}}
\end{minipage}
\hfill
\begin{minipage}[t]{0.32\linewidth}
\centering
\textbf{Multiple Small Texts}\\[-0.2em]
{\footnotesize Many small text elements distributed across the image area.}\\[0.4em]
\fbox{\includegraphics[width=\linewidth]{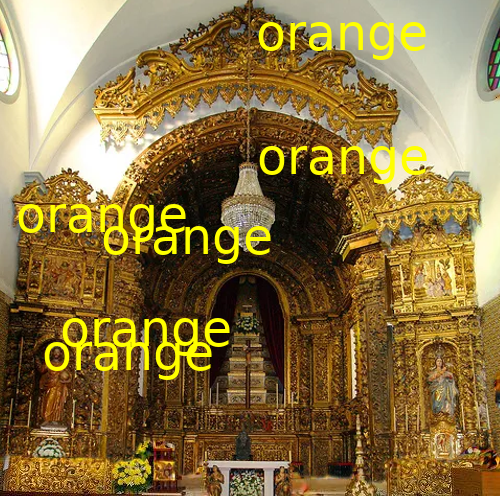}}
\end{minipage}
\hfill
\begin{minipage}[t]{0.32\linewidth}
\centering
\textbf{Big Text on Image}\\[-0.2em]
{\footnotesize A single large text overlay, however not occluding a substantial image region.}\\[0.4em]
\fbox{\includegraphics[width=\linewidth]{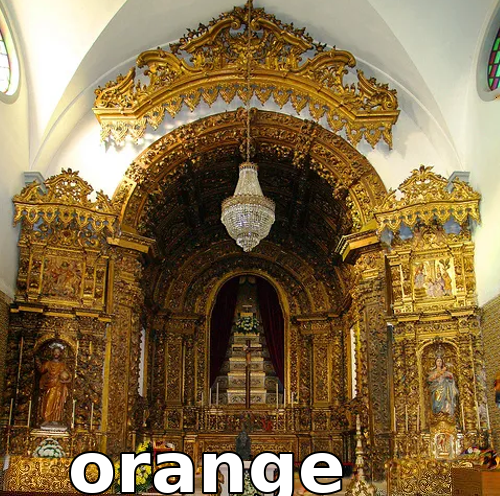}}
\end{minipage} \\
\vspace{0.5em}
\caption{
\textbf{Types of typographic corruptions.}
Left to right: Bezel, Multiple Small Texts, and Big Text on Image typographic corruptions.}
\label{fig:demo_corruption_altair}
\end{figure*}

%% file: tables/attack_success_rate.tex
\begin{table}[ht!]
\centering
\small
\caption{Different typographic attack success rate typographic corruptions. Values report Top-1 and Top-5 success rates, meaning the corruption attack target was listed in the Top-1 or Top-5. \label{table:attack_success_rate}}
\setlength{\tabcolsep}{6pt}
\resizebox{\textwidth}{!}{%
\begin{tabular}{l l l c c c c c c}
\toprule
Index & Corruption Text & Classifier Embedding & \multicolumn{2}{c}{Bezel} & \multicolumn{2}{c}{Big Text} & \multicolumn{2}{c}{Multiple Small Text} \\
\cmidrule(lr){4-5} \cmidrule(lr){6-7} \cmidrule(lr){8-9}
& & & Top-1 & Top-5 & Top-1 & Top-5 & Top-1 & Top-5 \\
\midrule
1 & \texttt{whistle} & whistle & \textbf{66.0\%} & 94.4\% & \textbf{42.6\%} & 74.3\% & \textbf{64.1\%} & 85.9\% \\
2 & \texttt{espresso} & espresso & \textbf{79.7\%} & 98.2\% & \textbf{47.1\%} & 75.2\% & \textbf{76.6\%} & 90.3\% \\
3 & \texttt{wing} & wing & 2.5\% & 10.3\% & 1.7\% & 7.4\% & 1.6\% & 8.3\% \\
4 & \texttt{overskirt} & overskirt & \textbf{67.5\%} & 94.7\% & \textbf{34.6\%} & 67.9\% & \textbf{74.7\%} & 91.0\% \\
5 & \texttt{convertible} & convertible & 10.9\% & 41.4\% & 9.2\% & 32.0\% & 19.0\% & 43.7\% \\
6 & \texttt{toyshop} & toyshop & \textbf{76.1\%} & 96.0\% & \textbf{45.8\%} & 72.6\% & \textbf{70.4\%} & 85.6\% \\
7 & \texttt{bluetick} & bluetick & 3.9\% & 11.4\% & 2.0\% & 6.9\% & 1.9\% & 6.6\% \\
8 & \texttt{consomme} & consomme & \textbf{41.3\%} & 77.1\% & \textbf{19.6\%} & 44.9\% & \textbf{56.2\%} & 77.3\% \\
9 & \texttt{mask} & mask & \textbf{61.7\%} & 92.5\% & \textbf{32.0\%} & 67.1\% & \textbf{58.5\%} & 85.7\% \\
10 & \texttt{orange} & orange & \textbf{83.2\%} & 99.1\% & \textbf{56.5\%} & 86.9\% & \textbf{84.3\%} & 94.7\% \\
11 & \texttt{cricket} & cricket & 37.0\% & 82.2\% & 16.8\% & 44.6\% & 43.4\% & 75.7\% \\
12 & \texttt{ipod} & iPod & \textbf{54.4\%} & 87.0\% & \textbf{38.6\%} & 70.3\% & \textbf{36.0\%} & 60.2\% \\
13 & \texttt{forklift} & forklift & 31.5\% & 58.3\% & 18.2\% & 36.8\% & 24.3\% & 44.4\% \\
14 & \texttt{necklace} & necklace & \textbf{60.0\%} & 92.9\% & \textbf{25.0\%} & 56.0\% & \textbf{50.6\%} & 75.0\% \\
15 & \texttt{cat} & tabby, tabby cat & 19.7\% & 61.8\% & 6.1\% & 26.5\% & 7.9\% & 40.0\% \\
16 & \texttt{zebra} & zebra & \textbf{48.4\%} & 83.8\% & \textbf{22.7\%} & 52.9\% & \textbf{36.2\%} & 64.0\% \\
17 & \texttt{poodle} & standard poodle & 15.4\% & 54.6\% & 9.4\% & 35.8\% & 10.2\% & 38.4\% \\
18 & \texttt{bear} & brown bear, bruin, Ursus arctos & 37.7\% & 80.0\% & 15.2\% & 49.7\% & 27.3\% & 66.1\% \\
19 & \texttt{boar} & wild boar, boar, Sus scrofa & 3.1\% & 22.5\% & 2.0\% & 10.2\% & 3.1\% & 18.2\% \\
20 & \texttt{ambulance} & ambulance & \textbf{71.4\%} & 93.4\% & \textbf{40.1\%} & 65.2\% & \textbf{62.0\%} & 80.0\% \\
21 & \texttt{pick} & pick, plectrum, plectron & 1.7\% & 12.9\% & 1.2\% & 7.8\% & 0.4\% & 3.4\% \\
22 & \texttt{bannister} & bannister, banister, balustrade, balusters, handrail & 0.4\% & 2.7\% & 0.5\% & 2.7\% & 0.5\% & 4.4\% \\
23 & \texttt{ballplayer} & ballplayer, baseball player & 2.7\% & 13.9\% & 2.3\% & 9.8\% & 1.9\% & 9.2\% \\
24 & \texttt{projectile} & projectile, missile & 0.0\% & 0.8\% & 0.0\% & 1.3\% & 0.0\% & 1.2\% \\
25 & \texttt{brassiere} & brassiere, bra, bandeau & 0.3\% & 5.9\% & 0.7\% & 5.2\% & 0.4\% & 16.4\% \\
26 & \texttt{mouse} & mouse, computer mouse & 29.2\% & 66.7\% & 14.2\% & 36.9\% & 44.1\% & 73.9\% \\
27 & \texttt{dishwasher} & dishwasher, dish washer, dishwashing machine & \textbf{73.6\%} & 95.2\% & \textbf{41.1\%} & 67.0\% & \textbf{65.1\%} & 80.8\% \\
28 & \texttt{church} & church, church building & \textbf{75.4\%} & 96.0\% & \textbf{48.5\%} & 75.0\% & \textbf{74.9\%} & 90.3\% \\
29 & \texttt{cloak} & cloak & \textbf{77.7\%} & 97.6\% & \textbf{35.7\%} & 67.7\% & \textbf{41.9\%} & 69.1\% \\
\bottomrule
\end{tabular}}
\end{table}

%% file: tables/steering_types_comparison.tex
\begin{figure}[htbp]
\centering
\setlength{\tabcolsep}{2pt}
\renewcommand{\arraystretch}{1.0}
\begin{tabular}{>{\centering\arraybackslash}m{0.5cm} >{\centering\arraybackslash}m{0.5cm} >{\centering\arraybackslash}m{0.78\textwidth}}
\toprule
& & \textbf{Results} \\
\midrule
\multirow{2}{*}{\begin{sideways}\small\textbf{Pre-Normed}\end{sideways}} &
\begin{sideways}\small\textbf{Medoid}\end{sideways} &
\includegraphics[width=0.7\linewidth]{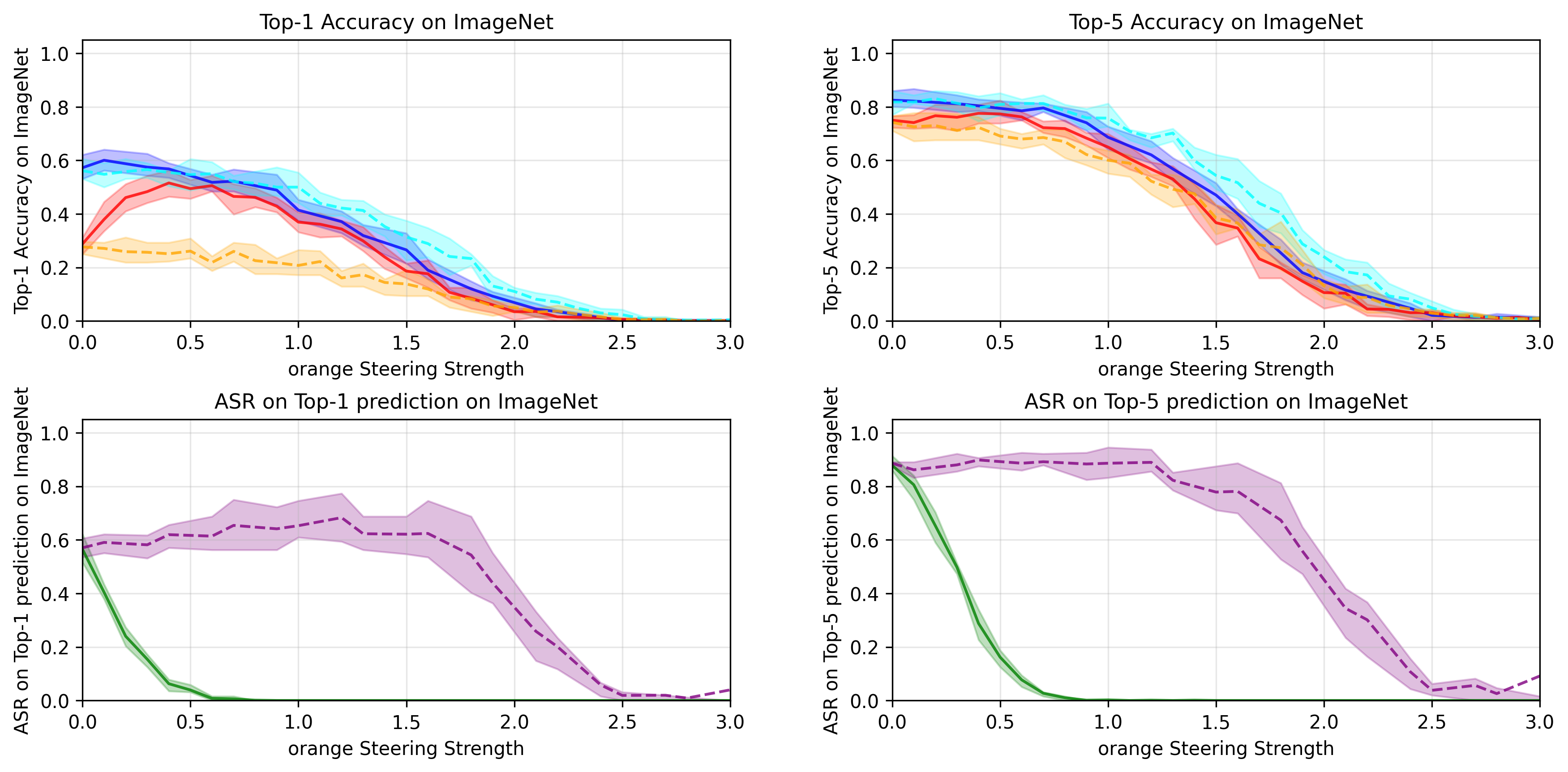} \\
\cmidrule{2-3}
& \begin{sideways}\small\textbf{Mean}\end{sideways} &
\includegraphics[width=0.7\linewidth]{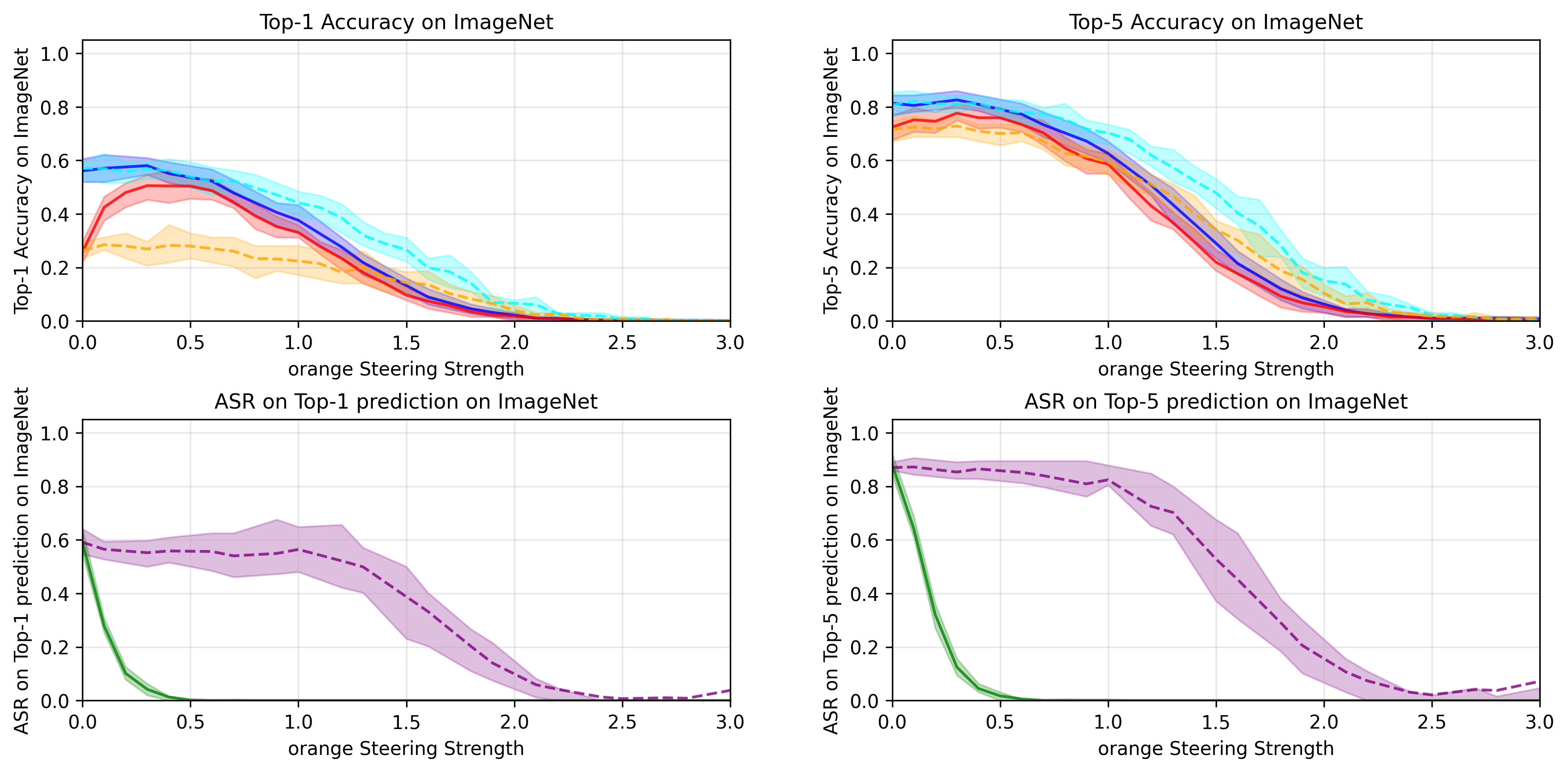} \\
\midrule
\multirow{2}{*}{\begin{sideways}\small\textbf{Post-Normed}\end{sideways}} &
\begin{sideways}\small\textbf{Medoid}\end{sideways} &
\includegraphics[width=0.7\linewidth]{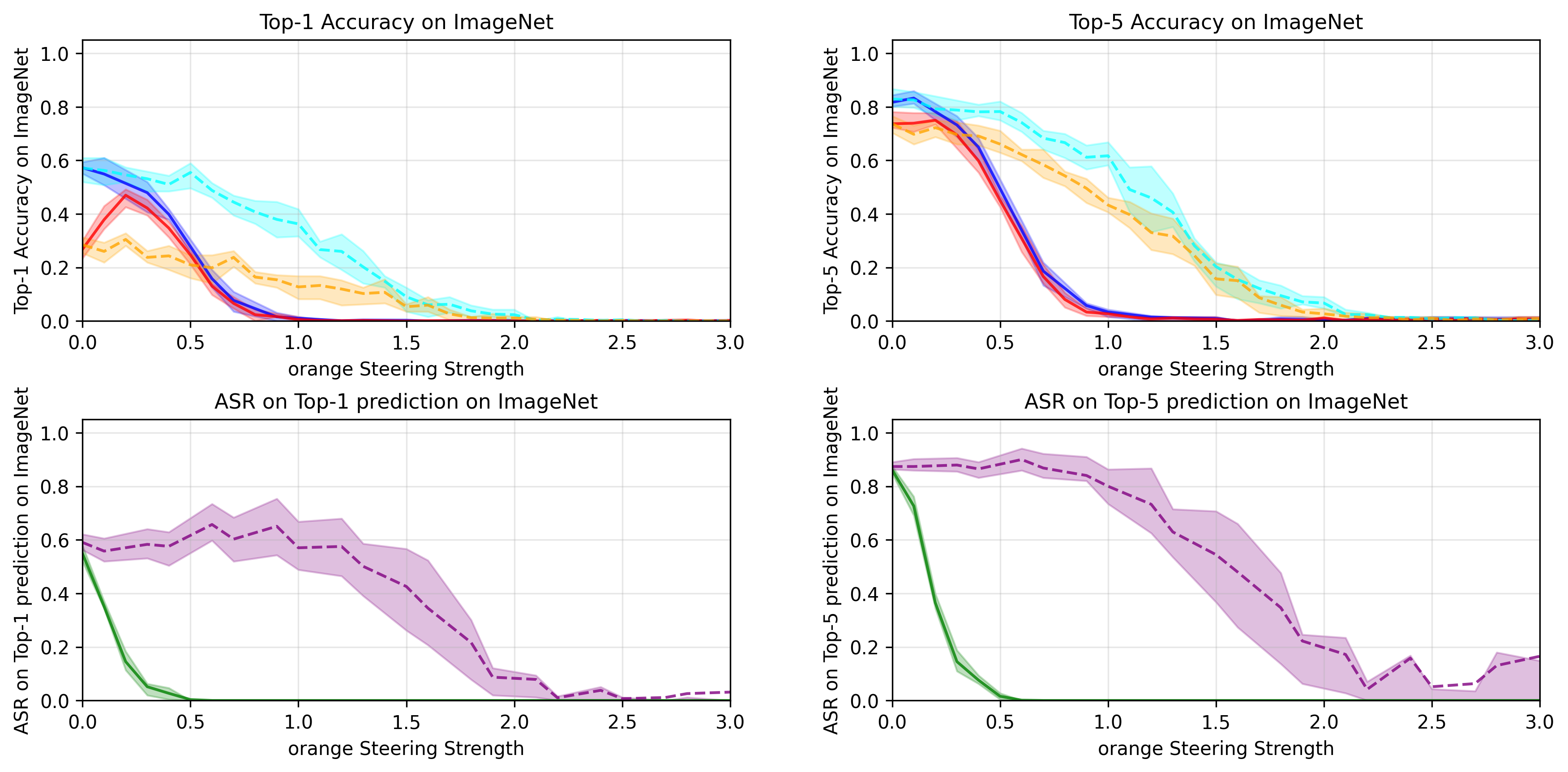} \\
\cmidrule{2-3}
& \begin{sideways}\small\textbf{Mean}\end{sideways} &
\includegraphics[width=0.7\linewidth]{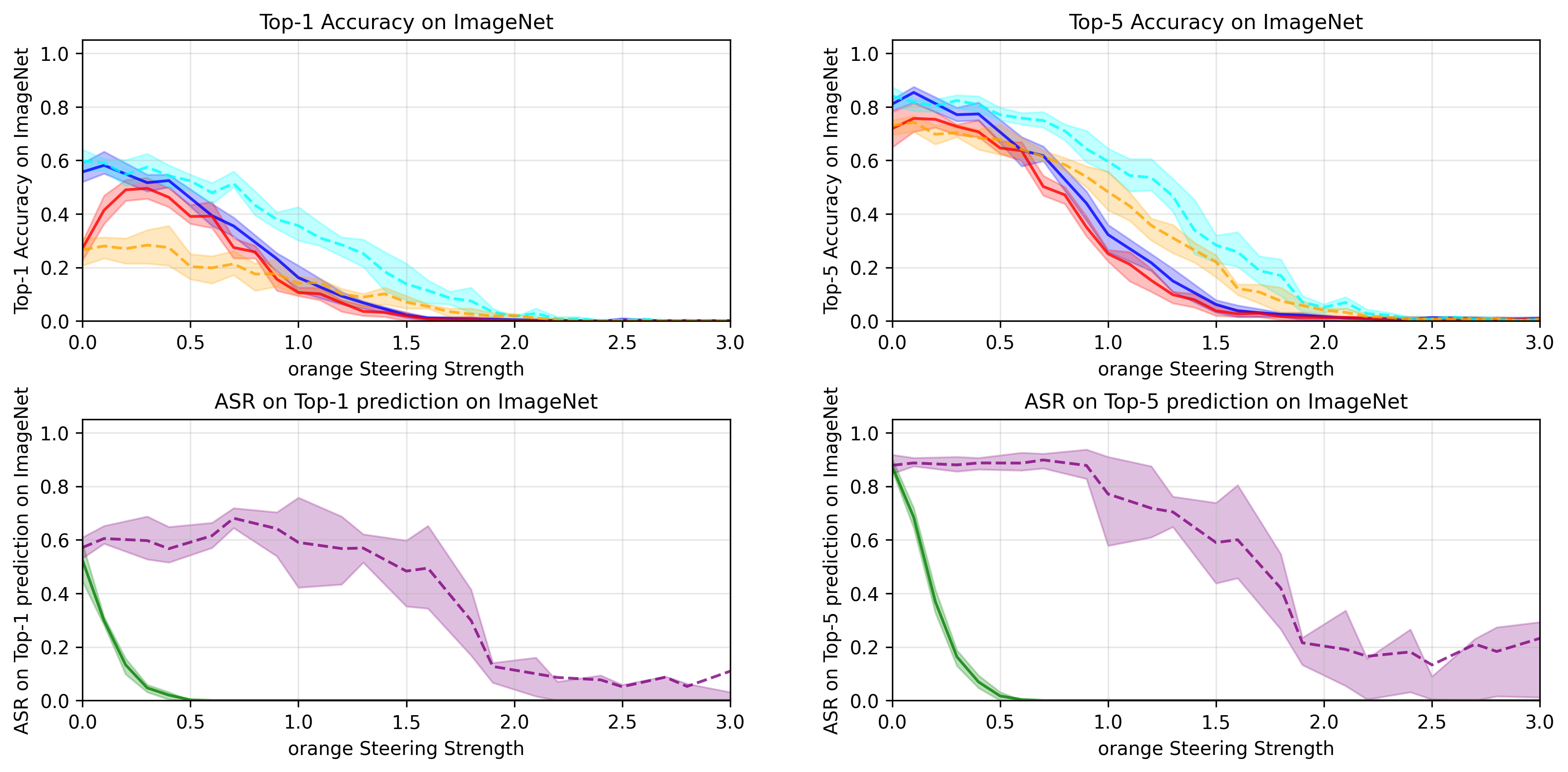} \\
\bottomrule
\end{tabular}
\caption{\textbf{Faceted results for the typographic object ``Orange'' with Big Text overlay.} 
Each row shows one combination of normalization regime (Pre-/Post-Normed) and 
aggregation method (Medoid/Mean). The \textit{x}-axis denotes steering strength 
$\alpha$; the \textit{y}-axis reports either ImageNet accuracy or attack success 
rate (ASR). \textbf{Accuracy:} \textcolor{blue}{Blue} --- clean input, circuit 
edges; \textcolor{red}{Red} --- corrupted input, circuit edges; 
\textcolor{cyan}{Cyan} (dashed) --- clean input, same-size non-circuit edges; 
\textcolor{Dandelion}{Orange} (dashed) --- corrupted input, same-size non-circuit 
edges. \textbf{ASR:} \textcolor{ForestGreen}{Green} --- circuit edges; 
\textcolor{Orchid}{Purple} (dashed) --- same-size non-circuit edges. Results are 
averaged across 20 circuits, each evaluated on a randomly sampled batch; shaded 
regions denote the interquartile range (Q25--Q75).}
\label{fig:orange_steering}
\end{figure}

%% file: tables/defence_steering_big_and_small_multiple_text.tex
\begin{center}
\tiny
\begin{longtable}{lll c c c c c c}
\toprule
& & &
\multicolumn{2}{c}{\textbf{Big Text on Image}} &
\multicolumn{2}{c}{\textbf{Multiple Small Text}} &
\multicolumn{2}{c}{\textbf{Bezel}} \\
\cmidrule(lr){4-5} \cmidrule(lr){6-7} \cmidrule(lr){8-9}
\textbf{Attack Word} & \textbf{Setting} & \textbf{Metric}
& \textbf{Base} & \textbf{ASR$\downarrow$90\%} & \textbf{Base} & \textbf{ASR$\downarrow$90\%} & \textbf{Base} & \textbf{ASR$\downarrow$90\%} \\
\midrule
\endfirsthead
\toprule
& & &
\multicolumn{2}{c}{\textbf{Big Text on Image}} &
\multicolumn{2}{c}{\textbf{Multiple Small Text}} &
\multicolumn{2}{c}{\textbf{Bezel}} \\
\cmidrule(lr){4-5} \cmidrule(lr){6-7} \cmidrule(lr){8-9}
\textbf{Attack Word} & \textbf{Setting} & \textbf{Metric}
& \textbf{Base} & \textbf{ASR$\downarrow$90\%} & \textbf{Base} & \textbf{ASR$\downarrow$90\%} & \textbf{Base} & \textbf{ASR$\downarrow$90\%} \\
\midrule
\endhead
\multirow{6}{*}{\textbf{ambulance}}
& Clean IN & ACC Top-1 (\%) & 56.9$_{\pm 5.7}$ & \textbf{58.0}$_{\pm 5.9}$ & 57.9$_{\pm 6.2}$ & 54.9$_{\pm 6.0}$ & 58.3$_{\pm 5.8}$ & 49.8$_{\pm 5.6}$ \\
&  & ACC Top-5 (\%) & 81.1$_{\pm 5.0}$ & \textbf{82.3}$_{\pm 4.9}$ & 81.9$_{\pm 4.7}$ & 80.8$_{\pm 4.6}$ & 82.8$_{\pm 4.6}$ & 75.2$_{\pm 5.4}$ \\
\cmidrule(lr){2-9}
& Corr. IN & ACC Top-1 (\%) & 33.6$_{\pm 6.0}$ & \textbf{48.7}$_{\pm 6.0}$ & 33.3$_{\pm 6.1}$ & \textbf{46.2}$_{\pm 5.8}$ & 32.3$_{\pm 6.0}$ & \textbf{44.9}$_{\pm 6.6}$ \\
&  & ACC Top-5 (\%) & 69.3$_{\pm 5.7}$ & \textbf{75.4}$_{\pm 4.8}$ & 69.5$_{\pm 5.7}$ & \textbf{72.9}$_{\pm 5.3}$ & 68.7$_{\pm 5.8}$ & \textbf{71.0}$_{\pm 6.2}$ \\
&  & ASR Top-1 (\%) & 41.4$_{\pm 6.2}$ & 2.9$_{\pm 2.1}$ & 40.6$_{\pm 6.4}$ & 0.7$_{\pm 1.2}$ & 41.0$_{\pm 6.3}$ & 2.4$_{\pm 1.9}$ \\
&  & ASR Top-5 (\%) & 64.3$_{\pm 5.6}$ & 7.3$_{\pm 3.4}$ & 65.3$_{\pm 5.8}$ & 4.0$_{\pm 2.5}$ & 64.9$_{\pm 5.8}$ & 6.9$_{\pm 3.5}$ \\
\midrule
\multirow{6}{*}{\textbf{church}}
& Clean IN & ACC Top-1 (\%) & 57.6$_{\pm 6.3}$ & 55.5$_{\pm 6.3}$ & 57.3$_{\pm 6.1}$ & 56.6$_{\pm 5.7}$ & 57.5$_{\pm 6.0}$ & 52.6$_{\pm 6.7}$ \\
&  & ACC Top-5 (\%) & 82.4$_{\pm 4.5}$ & 80.0$_{\pm 5.8}$ & 81.9$_{\pm 5.1}$ & \textbf{82.0}$_{\pm 4.7}$ & 82.0$_{\pm 4.2}$ & 77.7$_{\pm 5.2}$ \\
\cmidrule(lr){2-9}
& Corr. IN & ACC Top-1 (\%) & 28.9$_{\pm 5.9}$ & \textbf{49.4}$_{\pm 6.3}$ & 30.6$_{\pm 5.5}$ & \textbf{50.6}$_{\pm 7.1}$ & 30.2$_{\pm 5.5}$ & \textbf{46.7}$_{\pm 6.2}$ \\
&  & ACC Top-5 (\%) & 67.5$_{\pm 5.8}$ & \textbf{75.3}$_{\pm 4.9}$ & 68.2$_{\pm 5.2}$ & \textbf{76.7}$_{\pm 5.2}$ & 68.2$_{\pm 5.9}$ & \textbf{72.7}$_{\pm 5.6}$ \\
&  & ASR Top-1 (\%) & 46.5$_{\pm 6.3}$ & 2.0$_{\pm 1.7}$ & 46.9$_{\pm 5.7}$ & 2.5$_{\pm 1.9}$ & 48.3$_{\pm 5.8}$ & 4.7$_{\pm 2.7}$ \\
&  & ASR Top-5 (\%) & 75.3$_{\pm 4.7}$ & 6.9$_{\pm 3.0}$ & 76.7$_{\pm 5.3}$ & 10.9$_{\pm 3.6}$ & 76.9$_{\pm 5.2}$ & 21.6$_{\pm 5.5}$ \\
\midrule
\multirow{6}{*}{\textbf{cloak}}
& Clean IN & ACC Top-1 (\%) & 57.1$_{\pm 5.8}$ & 55.5$_{\pm 5.8}$ & 56.6$_{\pm 6.0}$ & 54.9$_{\pm 6.6}$ & 56.2$_{\pm 6.1}$ & 47.2$_{\pm 6.2}$ \\
&  & ACC Top-5 (\%) & 81.9$_{\pm 4.4}$ & 81.6$_{\pm 5.1}$ & 81.7$_{\pm 4.7}$ & 79.9$_{\pm 5.0}$ & 81.3$_{\pm 4.5}$ & 73.1$_{\pm 5.6}$ \\
\cmidrule(lr){2-9}
& Corr. IN & ACC Top-1 (\%) & 37.8$_{\pm 5.9}$ & \textbf{51.9}$_{\pm 5.7}$ & 37.5$_{\pm 6.3}$ & \textbf{50.8}$_{\pm 5.8}$ & 38.0$_{\pm 6.1}$ & \textbf{43.0}$_{\pm 6.2}$ \\
&  & ACC Top-5 (\%) & 71.7$_{\pm 5.6}$ & \textbf{76.5}$_{\pm 5.3}$ & 71.5$_{\pm 5.0}$ & \textbf{76.1}$_{\pm 4.9}$ & 71.3$_{\pm 5.5}$ & 68.7$_{\pm 5.2}$ \\
&  & ASR Top-1 (\%) & 33.2$_{\pm 5.5}$ & 2.9$_{\pm 2.0}$ & 36.4$_{\pm 5.9}$ & 1.9$_{\pm 1.8}$ & 36.4$_{\pm 5.8}$ & 2.2$_{\pm 1.8}$ \\
&  & ASR Top-5 (\%) & 66.5$_{\pm 6.2}$ & 9.5$_{\pm 3.6}$ & 68.7$_{\pm 5.9}$ & 8.0$_{\pm 3.8}$ & 67.8$_{\pm 5.4}$ & 7.6$_{\pm 3.5}$ \\
\midrule
\multirow{6}{*}{\textbf{consomme}}
& Clean IN & ACC Top-1 (\%) & 57.1$_{\pm 5.3}$ & 56.9$_{\pm 5.7}$ & 58.1$_{\pm 7.0}$ & 57.0$_{\pm 6.3}$ & 58.1$_{\pm 5.8}$ & 49.0$_{\pm 6.4}$ \\
&  & ACC Top-5 (\%) & 82.7$_{\pm 4.1}$ & 82.1$_{\pm 4.4}$ & 82.5$_{\pm 5.1}$ & 81.4$_{\pm 4.7}$ & 82.5$_{\pm 4.4}$ & 75.3$_{\pm 5.7}$ \\
\cmidrule(lr){2-9}
& Corr. IN & ACC Top-1 (\%) & 43.4$_{\pm 6.5}$ & \textbf{49.7}$_{\pm 6.0}$ & 43.5$_{\pm 6.1}$ & \textbf{47.8}$_{\pm 6.4}$ & 42.0$_{\pm 6.2}$ & \textbf{42.9}$_{\pm 6.1}$ \\
&  & ACC Top-5 (\%) & 72.2$_{\pm 5.6}$ & \textbf{76.0}$_{\pm 4.8}$ & 72.5$_{\pm 5.9}$ & \textbf{74.2}$_{\pm 5.0}$ & 71.6$_{\pm 5.4}$ & 69.2$_{\pm 5.7}$ \\
&  & ASR Top-1 (\%) & 19.1$_{\pm 4.8}$ & 1.5$_{\pm 1.5}$ & 18.7$_{\pm 4.9}$ & 1.1$_{\pm 1.3}$ & 19.2$_{\pm 4.6}$ & 1.9$_{\pm 1.8}$ \\
&  & ASR Top-5 (\%) & 44.6$_{\pm 7.1}$ & 4.9$_{\pm 2.7}$ & 45.1$_{\pm 5.3}$ & 5.5$_{\pm 2.8}$ & 45.6$_{\pm 5.8}$ & 8.7$_{\pm 4.0}$ \\
\midrule
\multirow{6}{*}{\textbf{dishwasher}}
& Clean IN & ACC Top-1 (\%) & 57.0$_{\pm 6.6}$ & 56.1$_{\pm 6.2}$ & 57.4$_{\pm 6.3}$ & 54.2$_{\pm 6.8}$ & 58.1$_{\pm 5.7}$ & 50.3$_{\pm 5.8}$ \\
&  & ACC Top-5 (\%) & 82.4$_{\pm 4.6}$ & 81.9$_{\pm 4.6}$ & 82.4$_{\pm 4.8}$ & 79.9$_{\pm 5.1}$ & 82.8$_{\pm 4.9}$ & 76.2$_{\pm 5.5}$ \\
\cmidrule(lr){2-9}
& Corr. IN & ACC Top-1 (\%) & 32.2$_{\pm 5.5}$ & \textbf{47.9}$_{\pm 6.0}$ & 31.0$_{\pm 5.5}$ & \textbf{44.6}$_{\pm 6.0}$ & 31.5$_{\pm 6.0}$ & \textbf{42.0}$_{\pm 6.2}$ \\
&  & ACC Top-5 (\%) & 65.6$_{\pm 5.8}$ & \textbf{74.3}$_{\pm 4.8}$ & 65.7$_{\pm 5.7}$ & \textbf{71.6}$_{\pm 5.5}$ & 65.2$_{\pm 5.9}$ & \textbf{68.2}$_{\pm 5.7}$ \\
&  & ASR Top-1 (\%) & 40.7$_{\pm 6.3}$ & 2.8$_{\pm 2.3}$ & 40.3$_{\pm 6.2}$ & 1.1$_{\pm 1.4}$ & 40.8$_{\pm 5.9}$ & 2.1$_{\pm 1.8}$ \\
&  & ASR Top-5 (\%) & 66.4$_{\pm 6.1}$ & 7.9$_{\pm 3.4}$ & 65.6$_{\pm 5.8}$ & 5.0$_{\pm 3.0}$ & 66.0$_{\pm 5.6}$ & 8.5$_{\pm 3.4}$ \\
\midrule
\multirow{6}{*}{\textbf{espresso}}
& Clean IN & ACC Top-1 (\%) & 56.5$_{\pm 6.7}$ & 56.1$_{\pm 6.6}$ & 56.9$_{\pm 6.4}$ & 55.2$_{\pm 6.1}$ & 56.5$_{\pm 5.6}$ & 47.9$_{\pm 6.5}$ \\
&  & ACC Top-5 (\%) & 81.1$_{\pm 5.7}$ & 81.0$_{\pm 4.9}$ & 82.4$_{\pm 4.6}$ & 80.8$_{\pm 4.8}$ & 81.9$_{\pm 4.9}$ & 73.8$_{\pm 5.2}$ \\
\cmidrule(lr){2-9}
& Corr. IN & ACC Top-1 (\%) & 31.6$_{\pm 6.0}$ & \textbf{50.1}$_{\pm 6.2}$ & 31.2$_{\pm 6.3}$ & \textbf{49.0}$_{\pm 5.5}$ & 31.7$_{\pm 5.6}$ & \textbf{45.2}$_{\pm 7.3}$ \\
&  & ACC Top-5 (\%) & 67.8$_{\pm 5.5}$ & \textbf{75.8}$_{\pm 5.2}$ & 68.9$_{\pm 6.7}$ & \textbf{74.8}$_{\pm 5.0}$ & 69.0$_{\pm 5.1}$ & \textbf{71.5}$_{\pm 6.5}$ \\
&  & ASR Top-1 (\%) & 45.9$_{\pm 5.8}$ & 4.5$_{\pm 2.9}$ & 46.4$_{\pm 6.2}$ & 1.8$_{\pm 1.7}$ & 46.2$_{\pm 6.0}$ & 4.0$_{\pm 2.4}$ \\
&  & ASR Top-5 (\%) & 74.4$_{\pm 6.0}$ & 9.9$_{\pm 4.1}$ & 75.5$_{\pm 5.4}$ & 8.3$_{\pm 3.6}$ & 74.8$_{\pm 5.7}$ & 14.6$_{\pm 4.4}$ \\
\midrule
\multirow{6}{*}{\textbf{ipod}}
& Clean IN & ACC Top-1 (\%) & 58.0$_{\pm 5.8}$ & 56.5$_{\pm 6.5}$ & 56.6$_{\pm 5.5}$ & 55.7$_{\pm 6.2}$ & 56.9$_{\pm 5.9}$ & 51.7$_{\pm 5.8}$ \\
&  & ACC Top-5 (\%) & 82.4$_{\pm 3.6}$ & 80.9$_{\pm 5.1}$ & 81.9$_{\pm 4.8}$ & 80.9$_{\pm 5.1}$ & 81.6$_{\pm 4.9}$ & 77.7$_{\pm 4.9}$ \\
\cmidrule(lr){2-9}
& Corr. IN & ACC Top-1 (\%) & 35.7$_{\pm 5.5}$ & \textbf{53.5}$_{\pm 6.4}$ & 35.0$_{\pm 6.8}$ & \textbf{50.8}$_{\pm 5.2}$ & 35.2$_{\pm 6.2}$ & \textbf{47.2}$_{\pm 6.7}$ \\
&  & ACC Top-5 (\%) & 71.0$_{\pm 5.2}$ & \textbf{77.9}$_{\pm 5.5}$ & 70.5$_{\pm 6.0}$ & \textbf{77.2}$_{\pm 4.6}$ & 71.0$_{\pm 5.9}$ & \textbf{73.7}$_{\pm 5.8}$ \\
&  & ASR Top-1 (\%) & 38.9$_{\pm 5.5}$ & 3.3$_{\pm 2.3}$ & 39.7$_{\pm 5.7}$ & 2.4$_{\pm 1.8}$ & 39.5$_{\pm 5.7}$ & 3.9$_{\pm 2.5}$ \\
&  & ASR Top-5 (\%) & 71.1$_{\pm 5.2}$ & 9.6$_{\pm 3.3}$ & 69.7$_{\pm 5.3}$ & 8.3$_{\pm 3.5}$ & 69.7$_{\pm 5.6}$ & 12.9$_{\pm 4.5}$ \\
\midrule
\multirow{6}{*}{\textbf{mask}}
& Clean IN & ACC Top-1 (\%) & 57.7$_{\pm 6.0}$ & 52.0$_{\pm 5.5}$ & 56.6$_{\pm 6.5}$ & 55.7$_{\pm 6.2}$ & 57.0$_{\pm 6.2}$ & 46.2$_{\pm 6.5}$ \\
&  & ACC Top-5 (\%) & 82.5$_{\pm 4.3}$ & 78.1$_{\pm 4.9}$ & 81.6$_{\pm 5.4}$ & 80.8$_{\pm 5.1}$ & 81.8$_{\pm 4.6}$ & 72.2$_{\pm 5.9}$ \\
\cmidrule(lr){2-9}
& Corr. IN & ACC Top-1 (\%) & 40.4$_{\pm 6.1}$ & \textbf{50.4}$_{\pm 5.6}$ & 38.9$_{\pm 6.1}$ & \textbf{51.0}$_{\pm 6.3}$ & 38.9$_{\pm 6.3}$ & \textbf{43.1}$_{\pm 6.6}$ \\
&  & ACC Top-5 (\%) & 71.4$_{\pm 6.0}$ & \textbf{76.0}$_{\pm 5.5}$ & 71.0$_{\pm 6.0}$ & \textbf{77.3}$_{\pm 5.0}$ & 69.8$_{\pm 5.6}$ & 68.9$_{\pm 6.7}$ \\
&  & ASR Top-1 (\%) & 31.0$_{\pm 5.3}$ & 1.6$_{\pm 1.4}$ & 30.3$_{\pm 5.6}$ & 1.4$_{\pm 1.4}$ & 30.3$_{\pm 5.5}$ & 1.6$_{\pm 1.7}$ \\
&  & ASR Top-5 (\%) & 65.0$_{\pm 4.8}$ & 5.9$_{\pm 3.0}$ & 64.3$_{\pm 6.2}$ & 7.1$_{\pm 3.3}$ & 64.7$_{\pm 6.3}$ & 11.3$_{\pm 4.3}$ \\
\midrule
\multirow{6}{*}{\textbf{orange}}
& Clean IN & ACC Top-1 (\%) & 56.1$_{\pm 6.9}$ & \textbf{58.0}$_{\pm 6.1}$ & 57.3$_{\pm 6.5}$ & 56.3$_{\pm 6.3}$ & 55.7$_{\pm 6.2}$ & 53.9$_{\pm 6.0}$ \\
&  & ACC Top-5 (\%) & 81.4$_{\pm 5.3}$ & \textbf{82.6}$_{\pm 4.5}$ & 81.9$_{\pm 5.1}$ & 81.0$_{\pm 5.1}$ & 81.1$_{\pm 5.2}$ & 79.5$_{\pm 4.8}$ \\
\cmidrule(lr){2-9}
& Corr. IN & ACC Top-1 (\%) & 25.8$_{\pm 6.1}$ & \textbf{50.6}$_{\pm 6.5}$ & 27.2$_{\pm 5.3}$ & \textbf{49.2}$_{\pm 6.7}$ & 26.5$_{\pm 5.5}$ & \textbf{47.3}$_{\pm 6.2}$ \\
&  & ACC Top-5 (\%) & 72.4$_{\pm 6.5}$ & \textbf{77.6}$_{\pm 5.8}$ & 72.2$_{\pm 6.3}$ & \textbf{75.7}$_{\pm 5.4}$ & 71.5$_{\pm 5.5}$ & \textbf{74.7}$_{\pm 5.3}$ \\
&  & ASR Top-1 (\%) & 59.1$_{\pm 6.8}$ & 4.2$_{\pm 2.2}$ & 58.1$_{\pm 6.3}$ & 1.9$_{\pm 1.8}$ & 57.2$_{\pm 6.4}$ & 4.6$_{\pm 2.7}$ \\
&  & ASR Top-5 (\%) & 88.2$_{\pm 4.6}$ & 12.4$_{\pm 3.8}$ & 88.2$_{\pm 3.9}$ & 11.8$_{\pm 4.5}$ & 88.0$_{\pm 3.9}$ & 25.3$_{\pm 5.1}$ \\
\midrule
\multirow{6}{*}{\textbf{toyshop}}
& Clean IN & ACC Top-1 (\%) & 56.5$_{\pm 6.6}$ & 54.2$_{\pm 6.7}$ & 57.6$_{\pm 6.0}$ & 55.7$_{\pm 6.8}$ & 57.1$_{\pm 6.3}$ & 48.7$_{\pm 6.7}$ \\
&  & ACC Top-5 (\%) & 81.5$_{\pm 3.7}$ & 80.1$_{\pm 5.7}$ & 82.1$_{\pm 4.3}$ & 80.8$_{\pm 5.0}$ & 81.5$_{\pm 5.0}$ & 74.6$_{\pm 5.3}$ \\
\cmidrule(lr){2-9}
& Corr. IN & ACC Top-1 (\%) & 30.9$_{\pm 5.0}$ & \textbf{46.8}$_{\pm 6.6}$ & 31.8$_{\pm 5.2}$ & \textbf{48.5}$_{\pm 6.0}$ & 30.7$_{\pm 6.1}$ & \textbf{43.7}$_{\pm 6.7}$ \\
&  & ACC Top-5 (\%) & 68.1$_{\pm 6.1}$ & \textbf{73.0}$_{\pm 5.8}$ & 68.1$_{\pm 5.8}$ & \textbf{74.7}$_{\pm 6.0}$ & 67.9$_{\pm 5.5}$ & \textbf{70.7}$_{\pm 5.9}$ \\
&  & ASR Top-1 (\%) & 46.8$_{\pm 6.6}$ & 2.4$_{\pm 1.9}$ & 46.6$_{\pm 6.1}$ & 1.6$_{\pm 1.8}$ & 46.9$_{\pm 6.4}$ & 4.3$_{\pm 2.8}$ \\
&  & ASR Top-5 (\%) & 73.6$_{\pm 5.3}$ & 6.7$_{\pm 3.3}$ & 73.3$_{\pm 5.2}$ & 6.3$_{\pm 3.0}$ & 73.7$_{\pm 5.6}$ & 12.9$_{\pm 4.6}$ \\
\midrule
\multirow{6}{*}{\textbf{whistle}}
& Clean IN & ACC Top-1 (\%) & 56.7$_{\pm 5.8}$ & 55.8$_{\pm 5.9}$ & 56.0$_{\pm 6.5}$ & 55.8$_{\pm 6.0}$ & 56.8$_{\pm 6.4}$ & 48.8$_{\pm 6.5}$ \\
&  & ACC Top-5 (\%) & 81.9$_{\pm 5.2}$ & 80.8$_{\pm 4.9}$ & 80.6$_{\pm 4.8}$ & \textbf{80.7}$_{\pm 4.5}$ & 81.7$_{\pm 4.8}$ & 74.8$_{\pm 5.8}$ \\
\cmidrule(lr){2-9}
& Corr. IN & ACC Top-1 (\%) & 32.8$_{\pm 6.1}$ & \textbf{49.7}$_{\pm 5.6}$ & 33.7$_{\pm 5.0}$ & \textbf{48.0}$_{\pm 5.6}$ & 32.9$_{\pm 6.3}$ & \textbf{45.2}$_{\pm 6.8}$ \\
&  & ACC Top-5 (\%) & 68.7$_{\pm 5.3}$ & \textbf{76.7}$_{\pm 4.9}$ & 70.0$_{\pm 5.7}$ & \textbf{73.9}$_{\pm 5.8}$ & 69.0$_{\pm 5.5}$ & \textbf{70.9}$_{\pm 5.7}$ \\
&  & ASR Top-1 (\%) & 42.3$_{\pm 7.3}$ & 3.9$_{\pm 2.6}$ & 43.4$_{\pm 6.3}$ & 2.0$_{\pm 1.9}$ & 43.1$_{\pm 6.0}$ & 3.9$_{\pm 2.5}$ \\
&  & ASR Top-5 (\%) & 72.6$_{\pm 5.7}$ & 12.6$_{\pm 4.1}$ & 73.7$_{\pm 5.5}$ & 10.0$_{\pm 4.0}$ & 74.0$_{\pm 6.0}$ & 14.3$_{\pm 4.8}$ \\
\midrule
\multirow{6}{*}{\textbf{zebra}}
& Clean IN & ACC Top-1 (\%) & 57.1$_{\pm 5.4}$ & 55.0$_{\pm 6.2}$ & 56.8$_{\pm 6.3}$ & 56.3$_{\pm 6.2}$ & 57.4$_{\pm 6.6}$ & 52.7$_{\pm 6.3}$ \\
&  & ACC Top-5 (\%) & 82.1$_{\pm 4.9}$ & 80.1$_{\pm 5.1}$ & 82.0$_{\pm 4.7}$ & 81.3$_{\pm 4.7}$ & 82.6$_{\pm 4.8}$ & 79.0$_{\pm 4.9}$ \\
\cmidrule(lr){2-9}
& Corr. IN & ACC Top-1 (\%) & 43.3$_{\pm 6.4}$ & \textbf{51.2}$_{\pm 6.5}$ & 42.8$_{\pm 6.3}$ & \textbf{52.6}$_{\pm 6.4}$ & 43.1$_{\pm 6.4}$ & \textbf{48.7}$_{\pm 6.3}$ \\
&  & ACC Top-5 (\%) & 73.6$_{\pm 6.2}$ & \textbf{76.6}$_{\pm 5.3}$ & 73.3$_{\pm 5.6}$ & \textbf{78.2}$_{\pm 5.1}$ & 74.1$_{\pm 5.4}$ & \textbf{74.5}$_{\pm 5.5}$ \\
&  & ASR Top-1 (\%) & 24.4$_{\pm 5.2}$ & 1.2$_{\pm 1.2}$ & 25.5$_{\pm 5.7}$ & 1.3$_{\pm 1.4}$ & 25.3$_{\pm 5.4}$ & 1.5$_{\pm 1.5}$ \\
&  & ASR Top-5 (\%) & 53.7$_{\pm 6.0}$ & 4.0$_{\pm 2.3}$ & 54.9$_{\pm 6.8}$ & 5.9$_{\pm 3.2}$ & 54.7$_{\pm 5.9}$ & 7.6$_{\pm 3.6}$ \\
\midrule
\bottomrule \\
\caption{Defended ImageNet (IN) performance under text corruptions (Corr.). 
Results are averaged over 20 circuits, over random batches of size 128. 
\textbf{Bold} indicates improvement over the unsteered base model. 
Steered results are reported at ASR$\downarrow$90\%, i.e.\ the steering 
strength $\alpha$ at which the attack success rate is reduced by 90\% 
relative to the base model.}
\end{longtable}
\end{center}
\label{tab:big_circuit_stering_typographic}

%% file: tables/facets_multiple_plots_normed_mean_steering.tex
\begin{longtable}{>{\centering\arraybackslash}m{1.5cm} >{\centering\arraybackslash}m{0.7\linewidth}}
\caption{%
\textbf{Faceted panel of Text Overlays under a fixed steering type (Pre-Normed Mean).} 
Column shows multiple small text overlays corruption regime. 
} \\
\toprule
\textbf{Text Overlay} & \textbf{Multiple small text} \\
\midrule
\endfirsthead

\toprule
\textbf{Text Overlay} & \textbf{Multiple small text} \\
\midrule
\endhead

\bottomrule
\multicolumn{2}{r}{\textit{Continued on next page}}
\endfoot

\bottomrule
\endlastfoot

\rotatebox{90}{\parbox{2.5cm}{\centering\textbf{Ambulance}}} &
\includegraphics[width=\linewidth,height=\textheight,keepaspectratio]{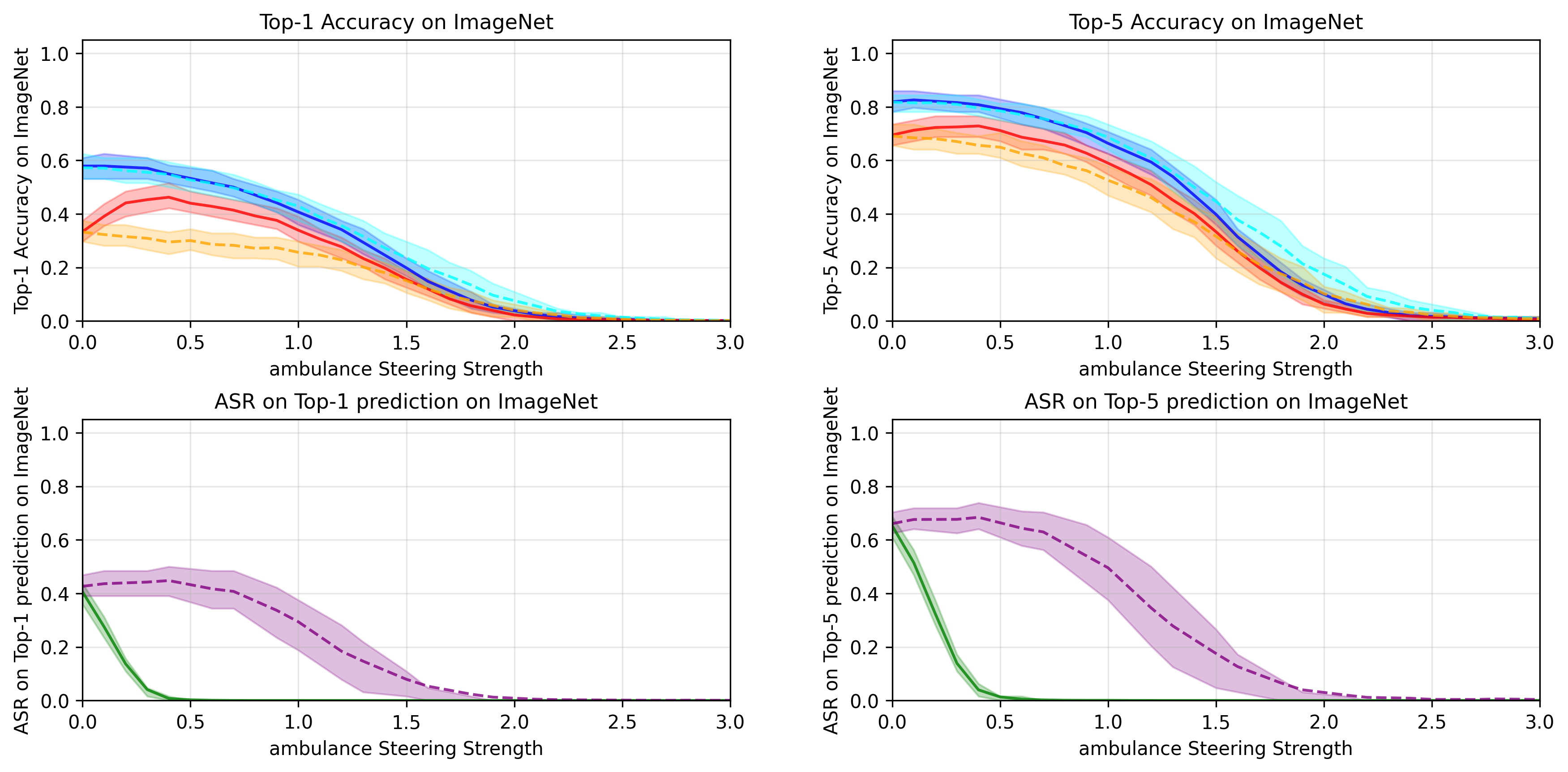} \\

\rotatebox{90}{\parbox{2.5cm}{\centering\textbf{Church}}} &
\includegraphics[width=\linewidth,height=\textheight,keepaspectratio]{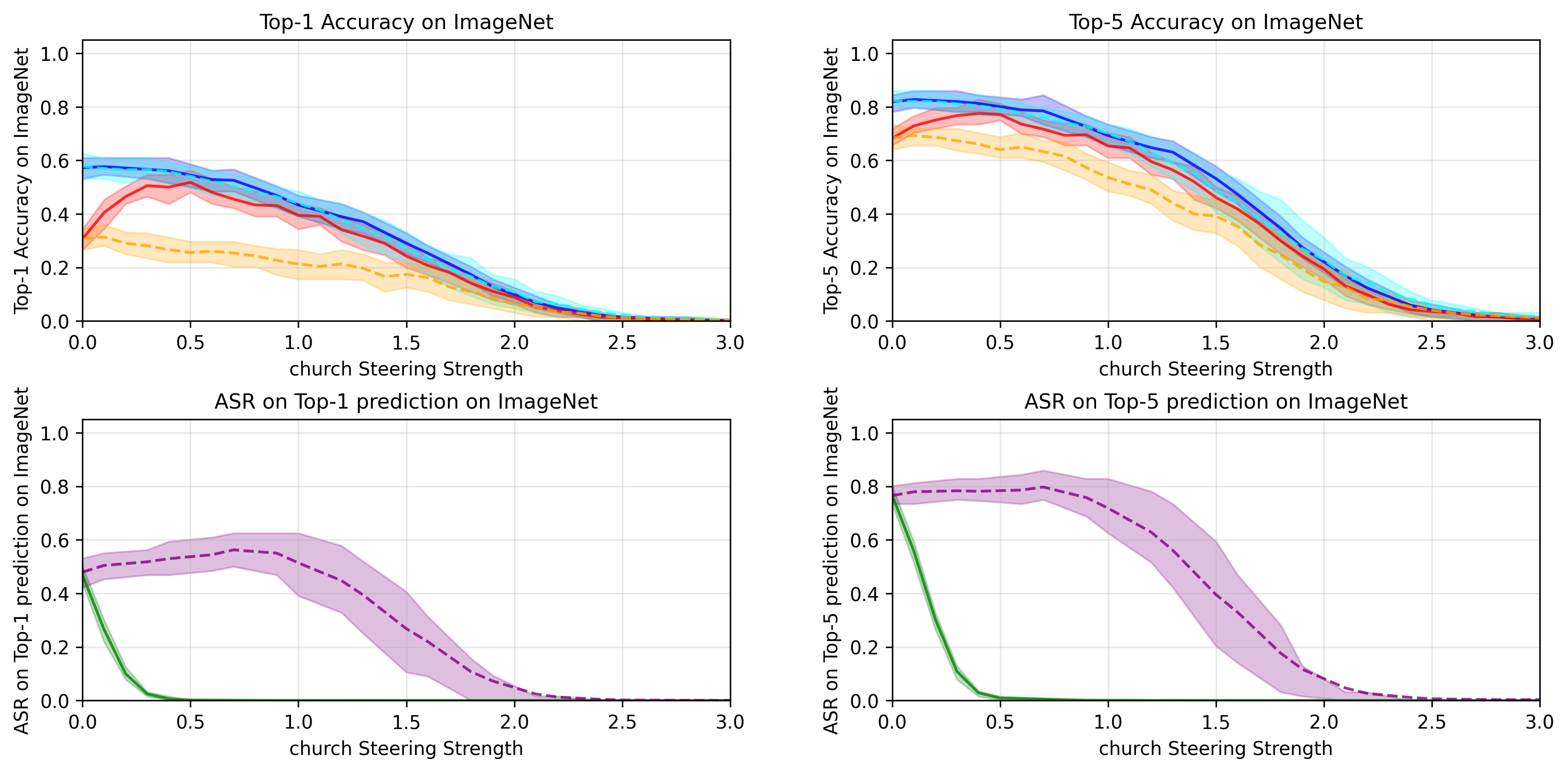} \\

\rotatebox{90}{\parbox{2.5cm}{\centering\textbf{Cloak}}} &
\includegraphics[width=\linewidth,height=\textheight,keepaspectratio]{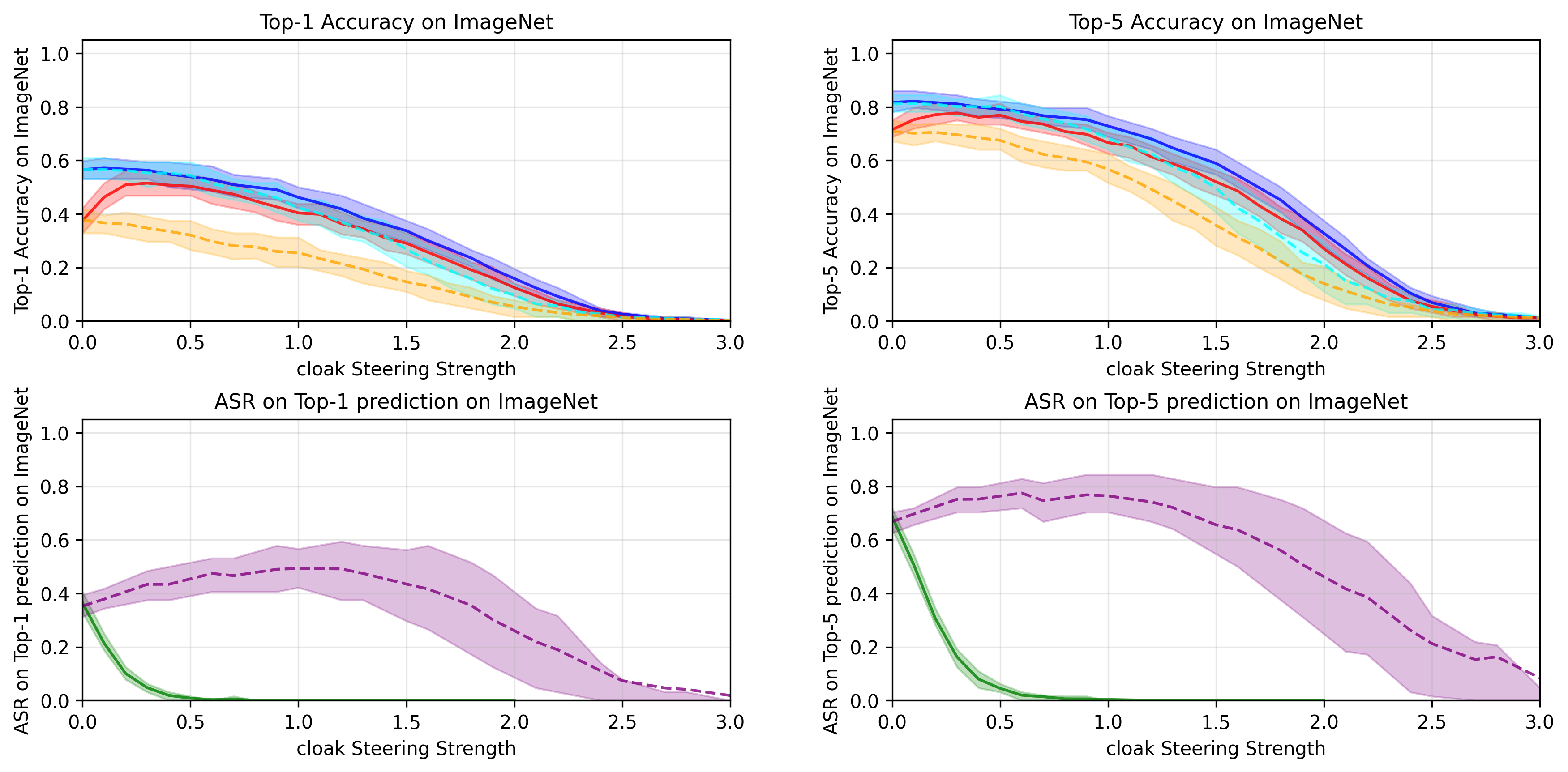} \\

\rotatebox{90}{\parbox{2.5cm}{\centering\textbf{Consomme}}} &
\includegraphics[width=\linewidth,height=\textheight,keepaspectratio]{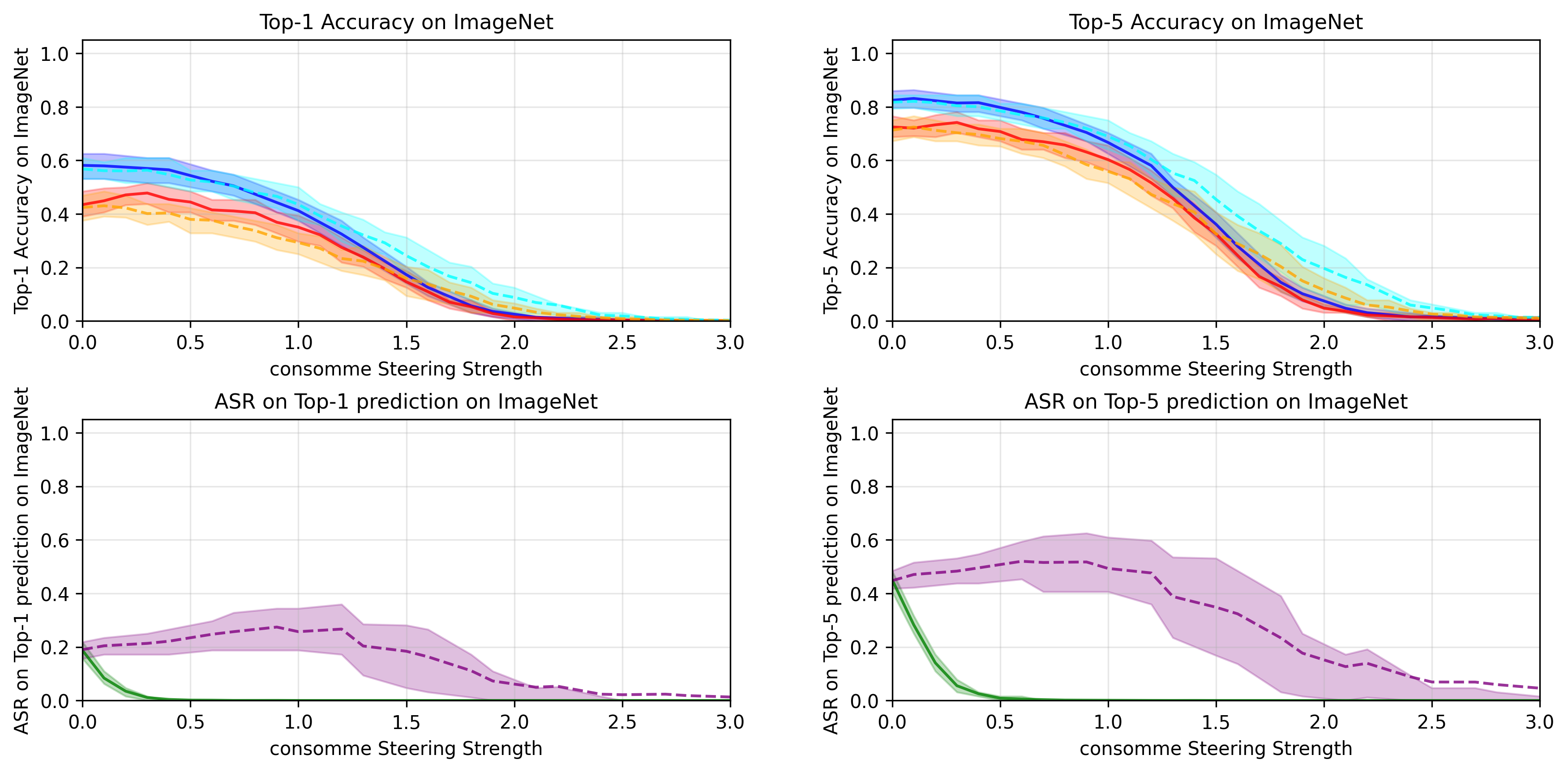} \\

\rotatebox{90}{\parbox{2.5cm}{\centering\textbf{Dishwasher}}} &
\includegraphics[width=\linewidth,height=\textheight,keepaspectratio]{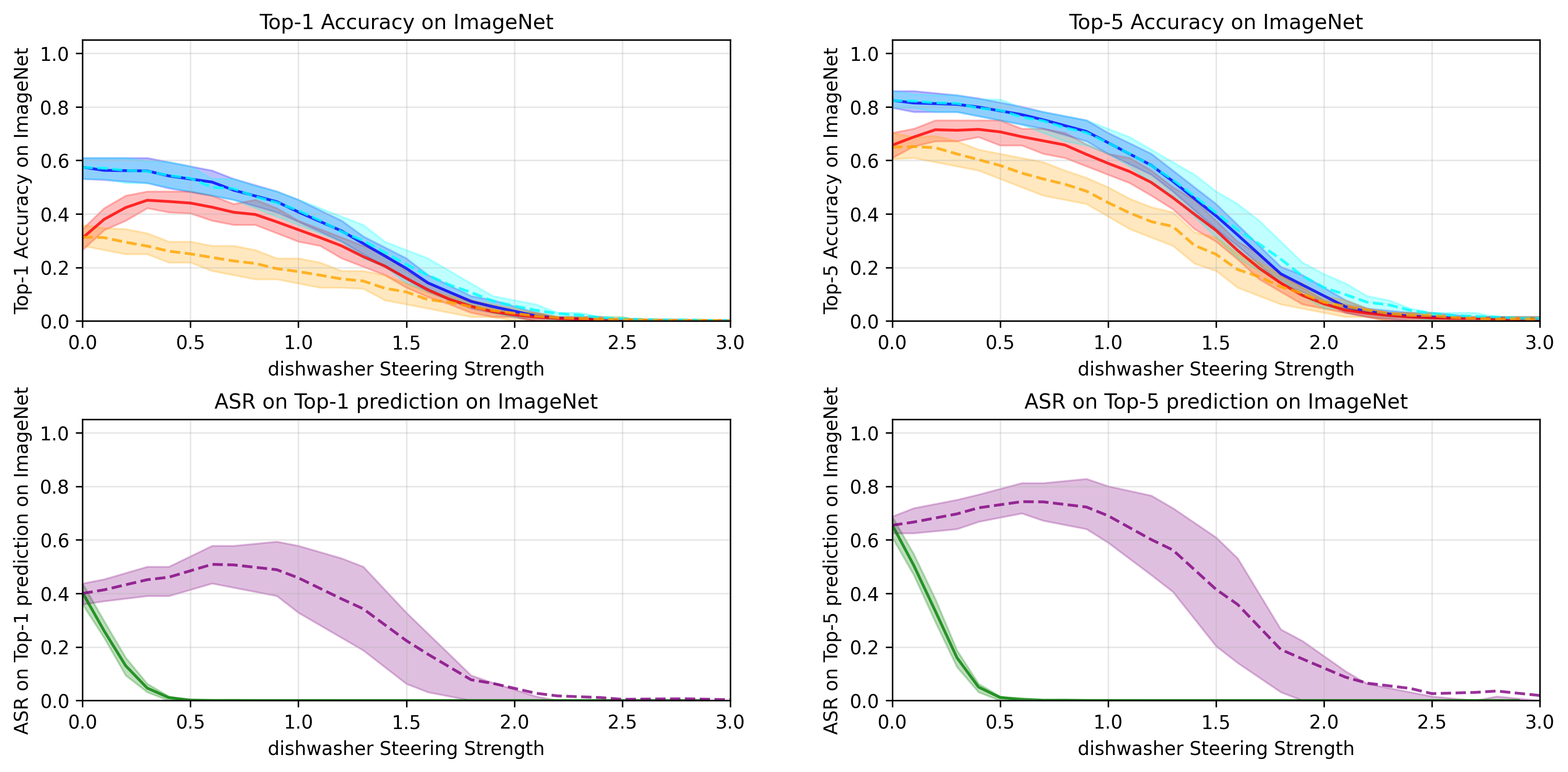} \\

\rotatebox{90}{\parbox{2.5cm}{\centering\textbf{Espresso}}} &
\includegraphics[width=\linewidth,height=\textheight,keepaspectratio]{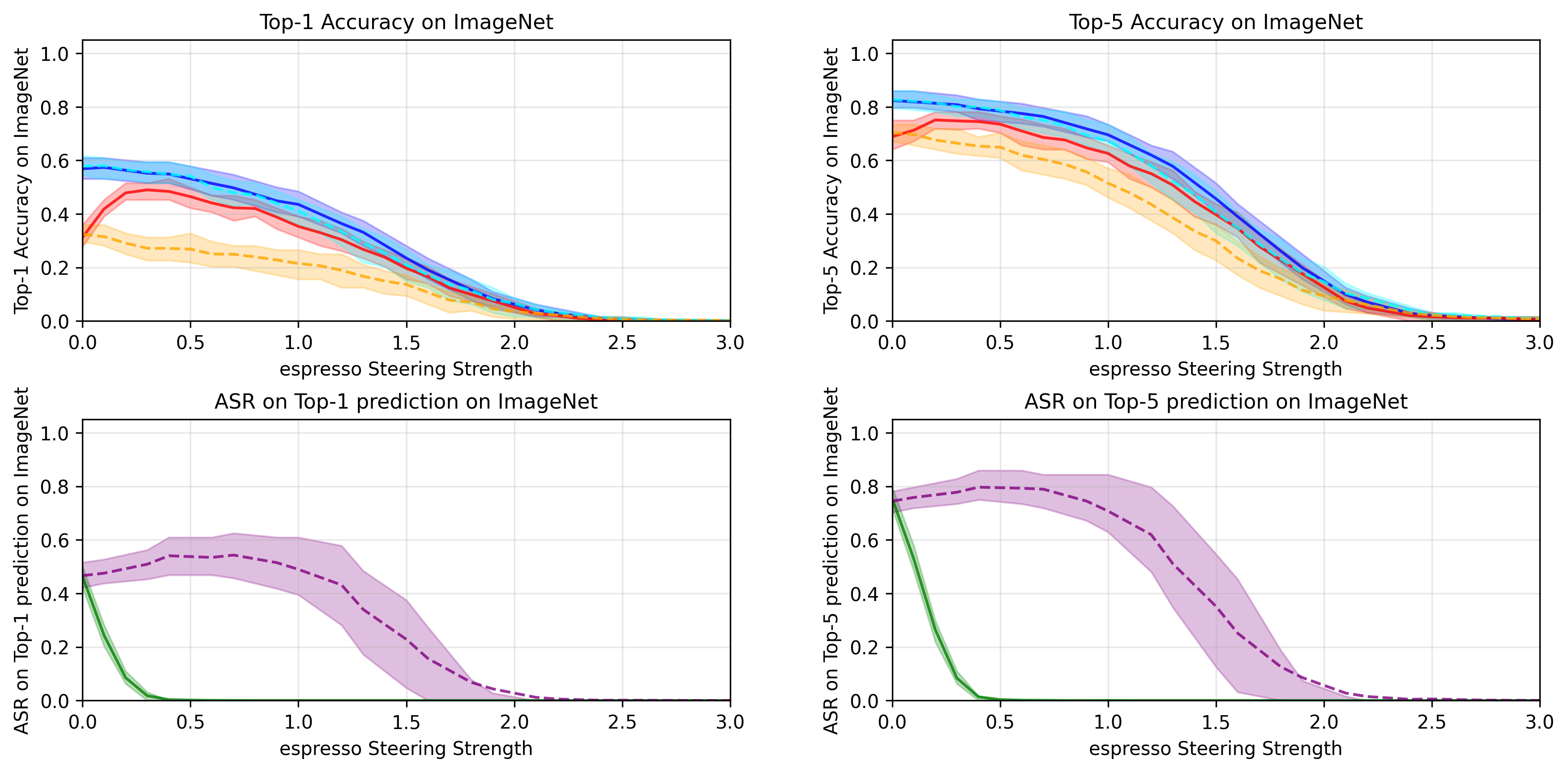} \\

\rotatebox{90}{\parbox{2.5cm}{\centering\textbf{iPod}}} &
\includegraphics[width=\linewidth,height=\textheight,keepaspectratio]{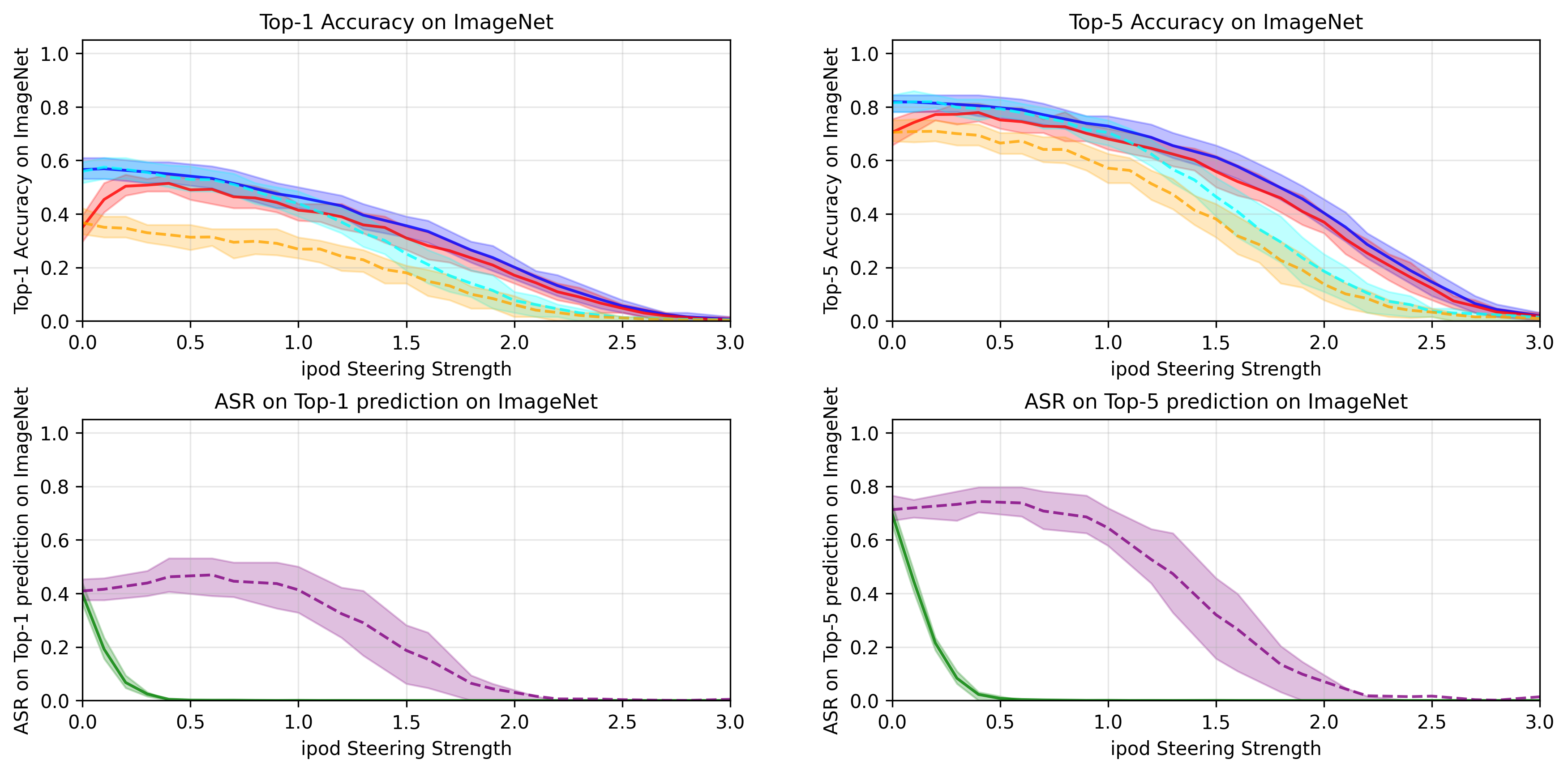} \\

\rotatebox{90}{\parbox{2.5cm}{\centering\textbf{Mask}}} &
\includegraphics[width=\linewidth,height=\textheight,keepaspectratio]{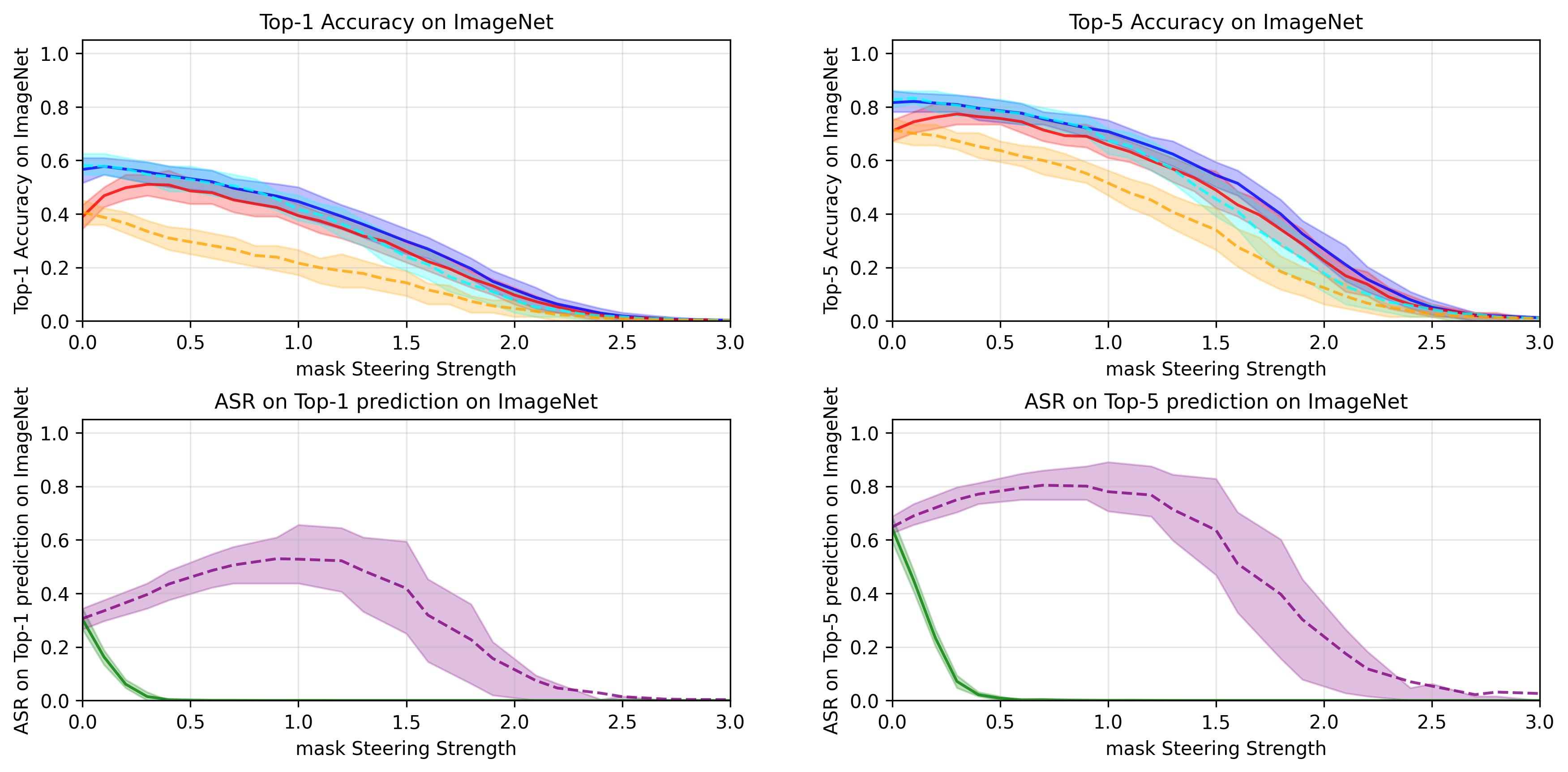} \\

\rotatebox{90}{\parbox{2.5cm}{\centering\textbf{Orange}}} &
\includegraphics[width=\linewidth,height=\textheight,keepaspectratio]{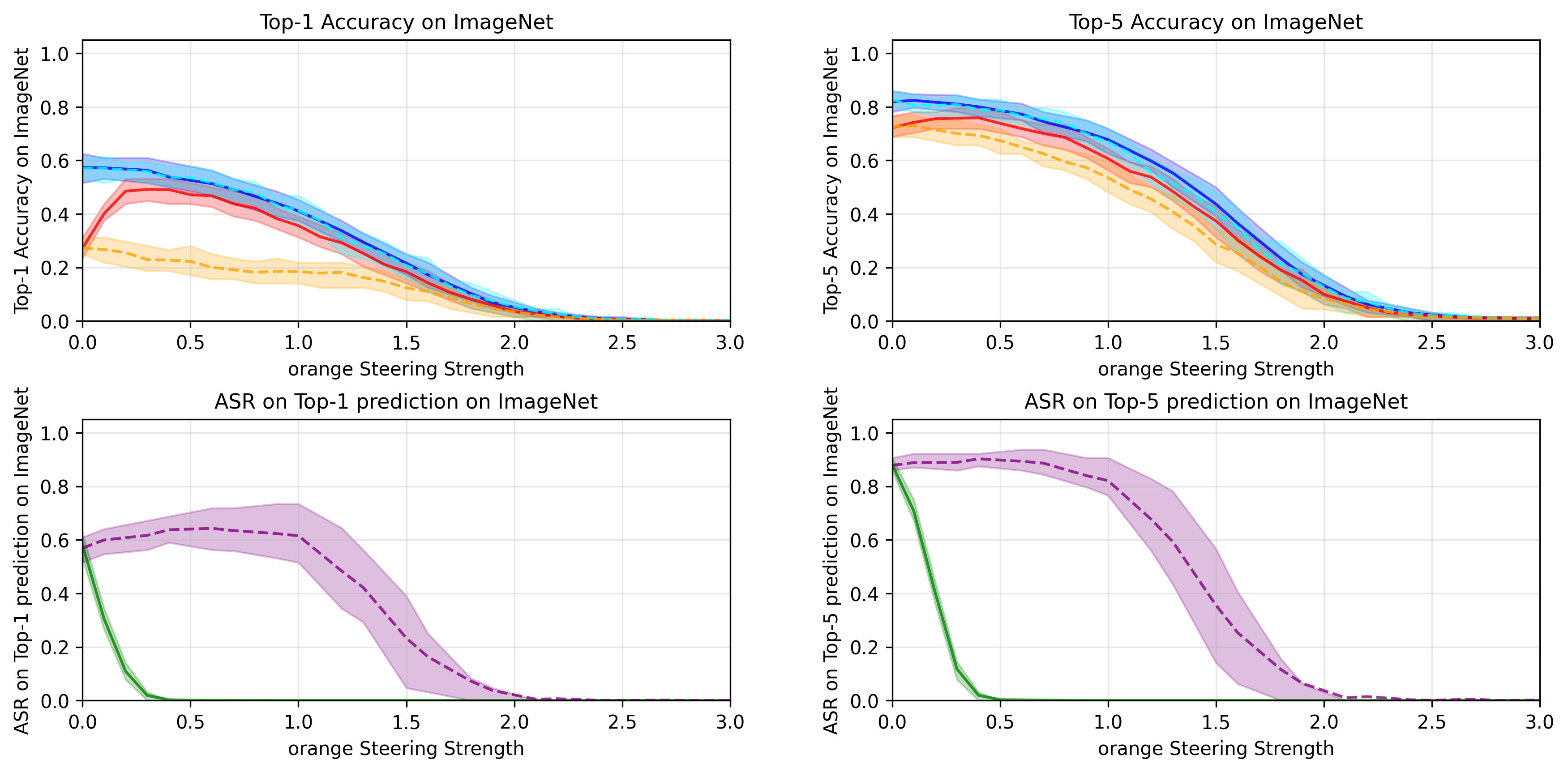} \\

\rotatebox{90}{\parbox{2.5cm}{\centering\textbf{Toyshop}}} &
\includegraphics[width=\linewidth,height=\textheight,keepaspectratio]{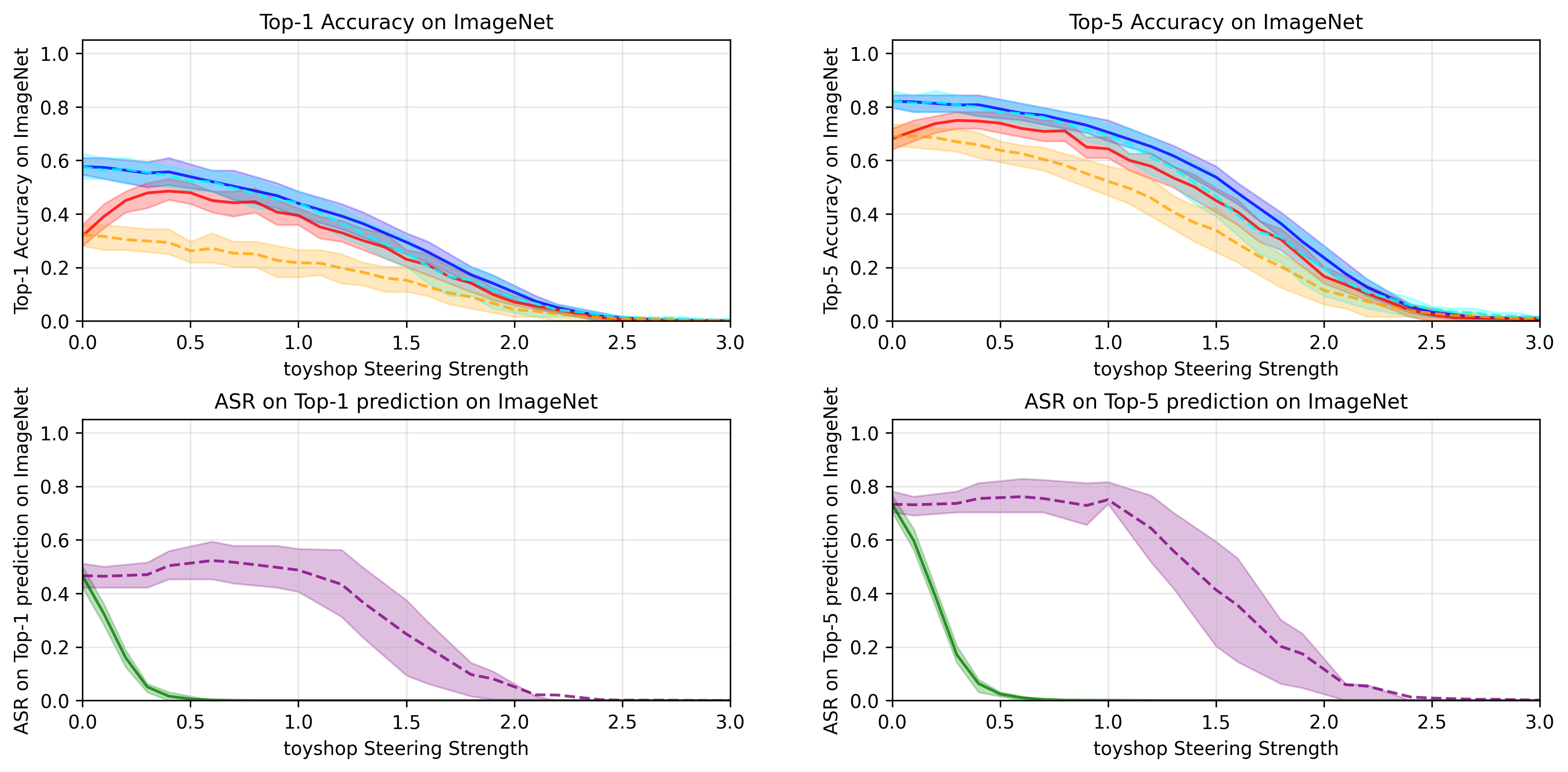} \\

\rotatebox{90}{\parbox{2.5cm}{\centering\textbf{Whistle}}} &
\includegraphics[width=\linewidth,height=\textheight,keepaspectratio]{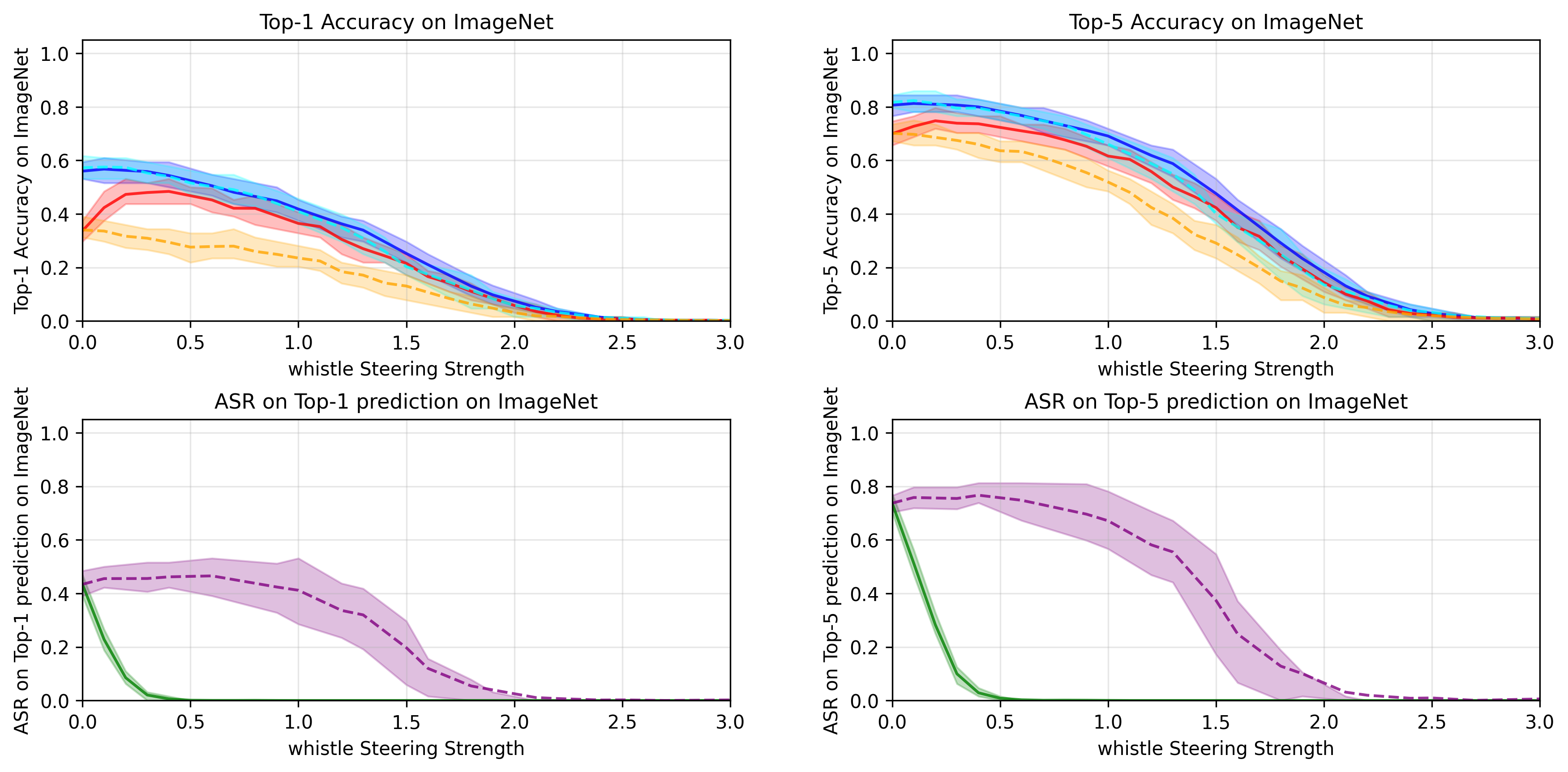} \\

\rotatebox{90}{\parbox{2.5cm}{\centering\textbf{Zebra}}} &
\includegraphics[width=\linewidth,height=\textheight,keepaspectratio]{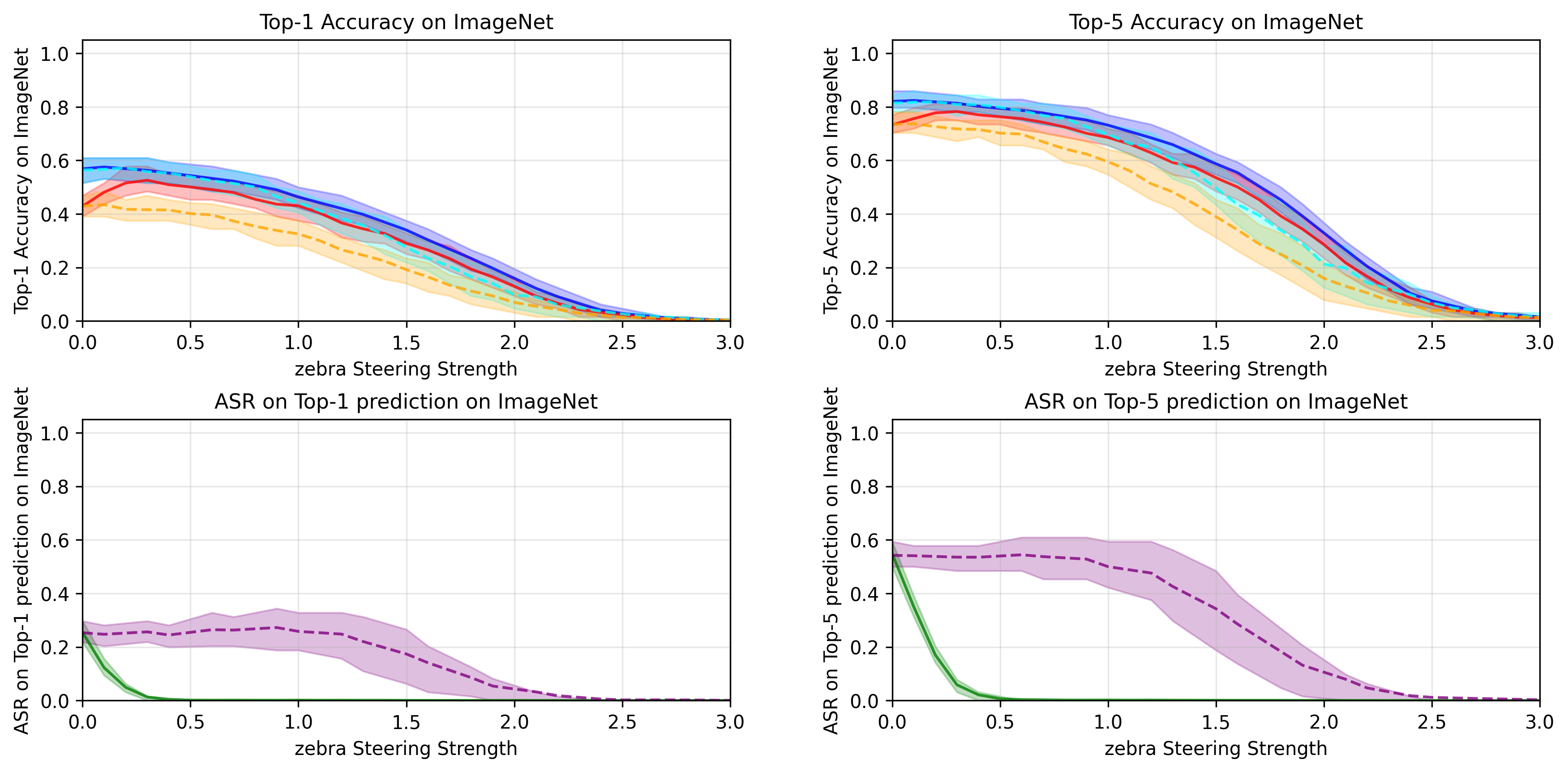} \\

\end{longtable}

\vspace{0.5em}
\begin{minipage}{0.9\linewidth}
\small
\textbf{Legend:} 
The \textit{x}-axis denotes the steering strength $\alpha$, and the \textit{y}-axis reports either ImageNet accuracy or attack success rate (ASR). \textbf{Accuracy:} \textcolor{blue}{Blue} --- clean input, steering on circuit edges; \textcolor{red}{Red} --- text-corrupted input, steering on circuit edges; \textcolor{cyan}{Cyan} (dashed) --- clean input, steering on a same-size set of non-circuit edges; \textcolor{Dandelion}{Orange} (dashed) --- text-corrupted input, steering on a same-size set of non-circuit edges. \textbf{Attack success rate:} \textcolor{ForestGreen}{Green} --- steering on circuit edges; \textcolor{Orchid}{Purple} (dashed) --- steering on a same-size set of non-circuit edges. Results are averaged across 20 circuits, each evaluated on a randomly sampled batch of examples from the relevant category; shaded regions 
denote the interquartile range (Q25--Q75). When steering on edges not in circuit, we do not see a reduction in ASR unless model performance starts heavily deteriorating.
\end{minipage}
\label{fig:facets_multiple_plots_normed_mean_steering}

%% file: tables/rococo_full_table.tex
\begin{table}[t]
\centering
\scriptsize
\setlength{\tabcolsep}{3pt}
\renewcommand{\arraystretch}{1.05}
\begin{tabular}{l|cccc|c}
\toprule
\textbf{Method} & \textbf{R@1} & \textbf{R@5} & \textbf{R@10} & \textbf{R$_\text{mean}$} & \textbf{RSMS} \\
\midrule
Base                  & 40.42 & 69.48 & 78.31 & 62.74 & 11.68 \\
\midrule
\multicolumn{6}{l}{\textit{$\alpha = 0.3$}} \\
\midrule
Assault Rifle Circuit & \textbf{44.19 $\pm$ 0.46} & \textbf{70.42 $\pm$ 0.32} & \textbf{79.71 $\pm$ 0.26} & \textbf{64.77 $\pm$ 0.28} & \textbf{5.66 $\pm$ 0.27} \\
Assault Rifle Random  & 42.28 $\pm$ 1.57          & 69.85 $\pm$ 0.96          & 78.57 $\pm$ 0.97          & 63.57 $\pm$ 1.13          & 9.15 $\pm$ 1.06          \\
\midrule
Revolver Circuit      & \textbf{44.04 $\pm$ 0.46} & \textbf{70.72 $\pm$ 0.45} & \textbf{79.98 $\pm$ 0.27} & \textbf{64.91 $\pm$ 0.36} & \textbf{6.15 $\pm$ 0.29} \\
Revolver Random       & 41.77 $\pm$ 1.18          & 69.81 $\pm$ 0.79          & 78.68 $\pm$ 0.72          & 63.42 $\pm$ 0.87          & 9.76 $\pm$ 0.96          \\
\midrule
Rifle Circuit         & \textbf{44.33 $\pm$ 0.29} & \textbf{71.20 $\pm$ 0.26} & \textbf{80.17 $\pm$ 0.23} & \textbf{65.23 $\pm$ 0.20} & \textbf{5.59 $\pm$ 0.14} \\
Rifle Random          & 42.51 $\pm$ 1.16          & 70.20 $\pm$ 0.65          & 78.97 $\pm$ 0.68          & 63.89 $\pm$ 0.80          & 8.99 $\pm$ 0.99          \\
\midrule
\multicolumn{6}{l}{\textit{$\alpha = 0.4$}} \\
\midrule
Assault Rifle Circuit & \textbf{42.36 $\pm$ 0.42} & 68.59 $\pm$ 0.46          & 78.04 $\pm$ 0.38          & \textbf{63.00 $\pm$ 0.37} & \textbf{5.15 $\pm$ 0.24} \\
Assault Rifle Random  & 41.69 $\pm$ 1.33          & \textbf{69.15 $\pm$ 0.96} & \textbf{78.05 $\pm$ 0.84} & 62.96 $\pm$ 1.01          & 9.09 $\pm$ 1.15          \\
\midrule
Revolver Circuit      & \textbf{42.01 $\pm$ 0.36} & 68.39 $\pm$ 0.24          & 77.65 $\pm$ 0.21          & 62.68 $\pm$ 0.22          & \textbf{5.69 $\pm$ 0.41} \\
Revolver Random       & 41.49 $\pm$ 1.41          & \textbf{69.24 $\pm$ 0.92} & \textbf{78.21 $\pm$ 0.86} & \textbf{62.98 $\pm$ 1.02} & 9.23 $\pm$ 1.09          \\
\midrule
Rifle Circuit         & \textbf{43.47 $\pm$ 0.48} & \textbf{70.22 $\pm$ 0.61} & \textbf{79.41 $\pm$ 0.43} & \textbf{64.37 $\pm$ 0.47} & \textbf{4.86 $\pm$ 0.21} \\
Rifle Random          & 41.95 $\pm$ 2.15          & 69.55 $\pm$ 1.44          & 78.41 $\pm$ 1.29          & 63.30 $\pm$ 1.60          & 8.79 $\pm$ 1.74          \\
\bottomrule
\end{tabular}

\caption{
\textbf{RoCOCO retrieval under steering} for $\alpha \in \{0.3, 0.4\}$.
\textbf{R@$k$}: fraction of queries where the correct item ranks in the
top~$k$ ($\uparrow$). \textbf{R$_\text{mean}$}: mean of R@1/5/10 ($\uparrow$).
\textbf{RSMS}~\cite{park2024rococo}: retrieval of danger-class images for
benign queries ($\downarrow$). Mean $\pm$ std over three seeds.
Bold indicates the better result within each Circuit/Random pair.
}
\label{tab:rococo_steering}
\end{table}

%% file: sections/appendix_stability.tex
\section{Stability of Circuits}
\label{app:stability}
\paragraph{Setup.}
For each class, we mine 20 circuits using a different set of 128 sampled examples from that class, i.e. the \textbf{class circuit }setup. We repeat the circuit mining
procedure multiple times with different random seeds while keeping the data
distribution fixed.

\paragraph{Edge Inclusion Frequency.} To quantify circuit stability, we compute for each class the empirical frequency with which each edge appears across repeated circuit discovery runs, then pool these frequencies across all Imagenette classes. The distribution reveals a clear, stable core: $34.5\%$ of all edges appear in $90$--$100\%$ of runs for the same class circuit, indicating that a substantial subset of edges is consistently recovered regardless of distributional shifts.

\begin{figure}[H]
    \centering
    \includegraphics[width=0.7\linewidth]{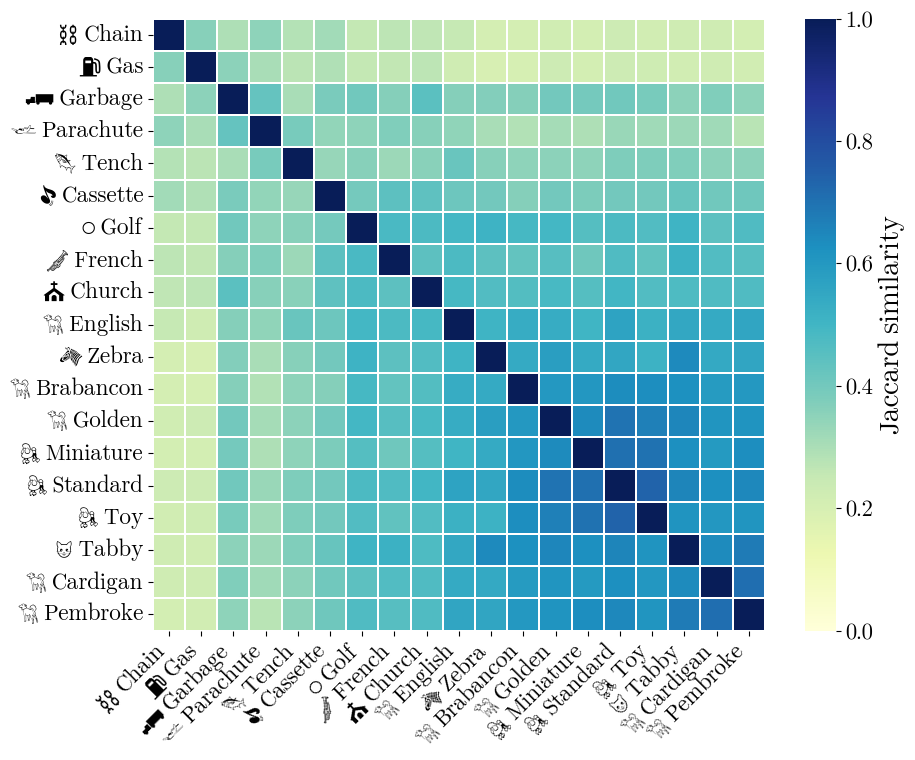}
    \caption{\textbf{Overlap of class circuits in CLIP.} 
    Each cell shows the Jaccard similarity between the sets of edges present 
    in \emph{all} runs (frequency~$= 1.0$) for a pair of classes. Classes are 
    ordered by hierarchical clustering. Dog breeds (bottom-right block) share 
    substantially more core edges with each other than with semantically 
    unrelated classes.}
    \label{fig:core_edges_jaccard}
\end{figure}

The mined circuits can vary across runs due to stochasticity in the mining 
procedure: small differences in which edges are selected for functionally 
similar components can propagate and compound, leading to larger circuit-level 
divergence.
We also show that the ``core edges'' of class circuits, those edges 
present across \emph{all} runs, vary between classes, and similarly 
to \autoref{fig:ridge_plots}, semantically similar classes exhibit 
strong overlap in their core edges, see \autoref{fig:core_edges_jaccard}.

\begin{figure}[H]
    \centering
    \includegraphics[width=0.9\linewidth]{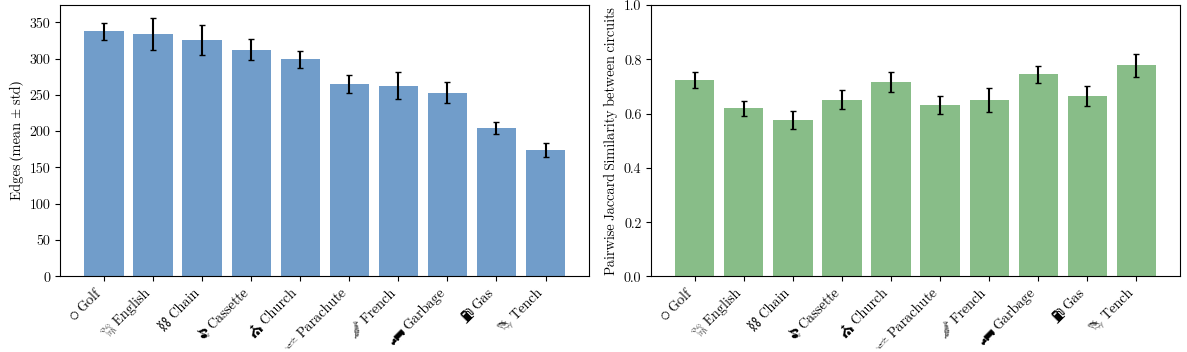}
    \caption{\textbf{Circuit size and stability per class.} Circuit sizes in 
    CLIP and stability across Imagenette\cite{Howard_Imagenette_2019} classes. We report
    the average circuit size and the mean pairwise Jaccard similarity between
    circuits mined from repeated runs.}
    \label{fig:circuit_sizes_and_stability}
\end{figure}

\paragraph{Circuit Size and Stability.}
Finally, we analyze circuit size and stability across Imagenette classes. For
each class, we compute the size of the mined circuits as well as the average
pairwise Jaccard similarity between circuits obtained from different runs,
where the Jaccard similarity between two circuits $A$ and $B$ is defined as
\[
J(A, B) = \frac{|A \cap B|}{|A \cup B|}.
\]
\autoref{fig:circuit_sizes_and_stability} summarizes these results, showing
both circuit size and stability per class.

We also examine which edge types contribute to lower circuit stability by 
analysing their frequency of inclusion across repeated runs for specific class 
circuits. In both ViT-B and CLIP, edges originating from attention heads 
(\texttt{attn→attn\_input}, \texttt{attn→mlp}) exhibit notably lower stability, 
with the majority classified as unstable. See \autoref{fig:stability_edge_type}.

\input{figures/stability_of_circuits/stability_figure_same_and_diff_input}

%% file: figures/stability_of_circuits/stability_figure_same_and_diff_input.tex
\begin{figure}[h]
\centering
\begin{subfigure}[t]{0.48\textwidth}
    \centering
    \includegraphics[width=\linewidth]{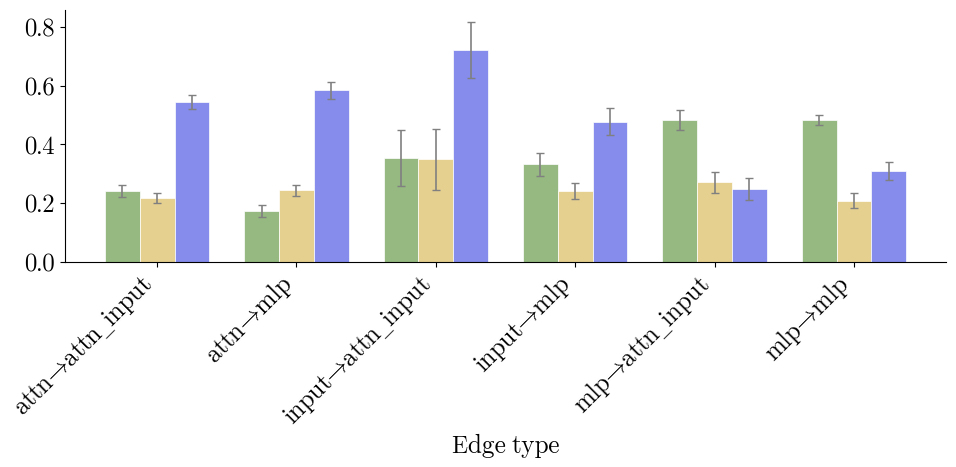}
    \caption{Stability of edges per edge type for ViT-B}
\end{subfigure}\hfill
\begin{subfigure}[t]{0.48\textwidth}
    \centering
    \includegraphics[width=0.9\linewidth]{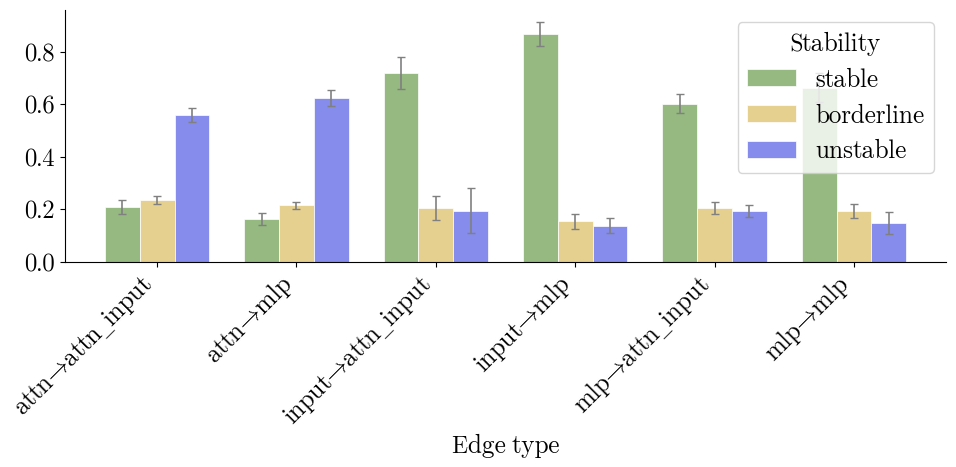}
    \caption{Stability of edges per edge type for CLIP}
\end{subfigure}
\caption{\textbf{Stability of different network components.} Stability of edges per edge type, broken down into stable ($>$90\% 
inclusion frequency), borderline (50--90\%), and unstable ($<$50\%) categories. 
The $y$-axis reports the fraction of edges of each type falling into each 
stability category, averaged over all class circuits. Edges originating from 
attention heads are disproportionately unstable in both models. 
}

\label{fig:stability_edge_type}
\end{figure}
\paragraph{Effect of Dataset Size on Circuit Stability.}
\begin{wrapfigure}[15]{r}{0.5\textwidth}
    \centering
    \vspace{-2em}
    \includegraphics[width=0.4\textwidth]{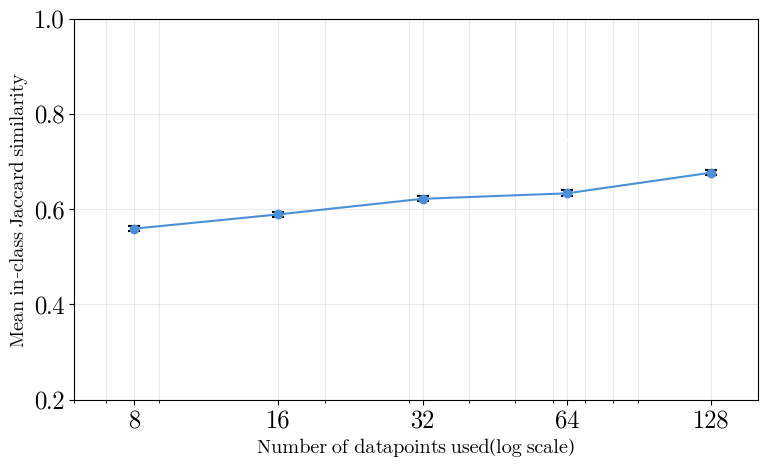}
    \caption{\textbf{Circuit stability with \#samples.} Mean within-class pairwise Jaccard similarity between circuits mined from repeated runs, as a function of the number of datapoints used for circuit mining (log scale). Circuit stability increases consistently with dataset size.}
    \label{fig:batchsize_stability}
\end{wrapfigure}
We investigate how the number of datapoints used for circuit mining affects 
the consistency of the recovered circuits. For each batch size in 
$\{8, 16, 32, 64, 128\}$, we mine 20 circuits per class and compute the 
mean pairwise Jaccard similarity across runs. As shown in 
\autoref{fig:batchsize_stability}, circuit stability increases monotonically 
with dataset size. This suggests 
that larger datasets, which by construction capture greater input diversity, 
lead to more consistent circuit recovery. The model's computational pathways 
for a given class are more reliably identified when the mining procedure is 
exposed to a broader range of within-class variation.

%% file: sections/appendix_binary_classification.tex
\section{Analysing Circuit Unions}
\label{app:binary_classification}

An expected property of circuits is \emph{compositionality}: combining circuits corresponding to different classes should yield a circuit that reflects the combined functionality. In particular, the union of two class-specific circuits should ideally form a discriminative binary-classifier circuit capable of distinguishing between the two classes.

We extract 20 circuits for each Imagenette\cite{Howard_Imagenette_2019} class and retain only those with sufficient faithfulness (maintained accuracy > 80\% on the class-circuit task). For a detailed specification, see \ref{app:class_circuits}. We then evaluate zero-shot performance on merged circuits formed from pairs of classes. The dataset for the \textbf{two-class task} is composed of images form these two classes, with corruptions as described in \autoref{sec:method}. In addition, we compare our results to circuits of similar size and composed of random edges.

\paragraph{Setup.}
We consider the 10 classes of the Imagenette
dataset~\cite{Howard_Imagenette_2019}. For each class, we extract 20 faithful
circuits using our proposed method in CLIP, with the treshold as stated in \autoref{app:class_circuits}, achieving 80\% accuracy or higher. We then for each pair of classes form 8 unions of class-specific circuits and evaluate their discriminative power.

\begin{figure}[H]
    \centering
    \includegraphics[width=0.9\linewidth]{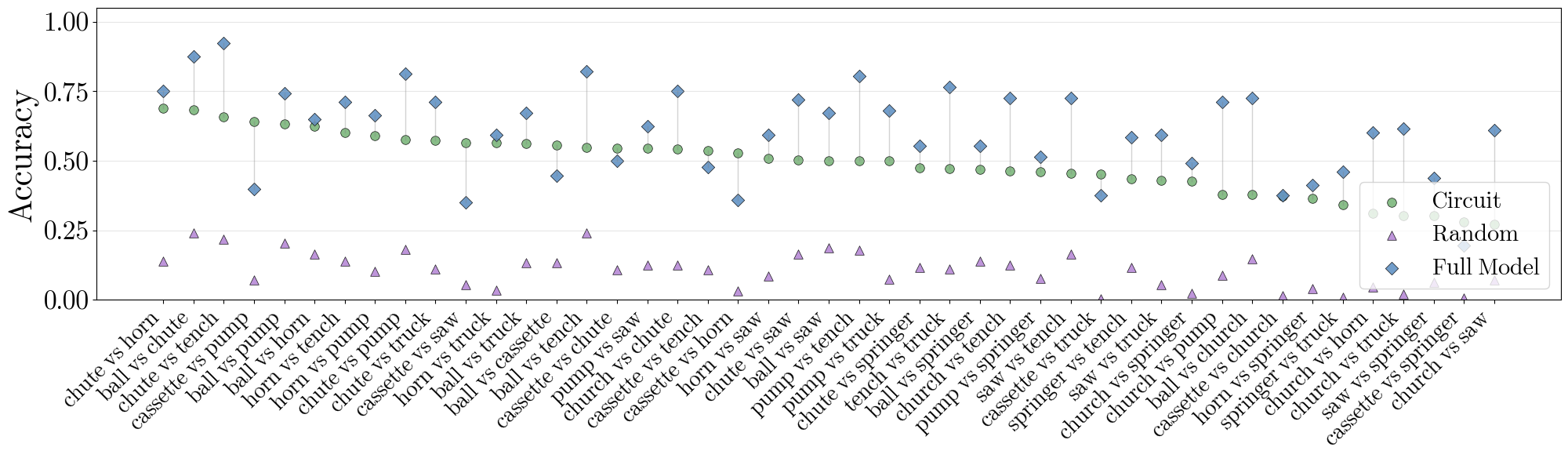}
    \caption{Zero-shot classification performance using unions of class-specific
    circuits.}
    \label{fig:zeroshot_union}
\end{figure}

\paragraph{Zero-Shot Classification.}
We first evaluate the ability of circuit unions to perform zero-shot
classification. \autoref{fig:zeroshot_union} shows the classification
performance obtained by applying the union of two class circuits without
additional training.
\paragraph{Pairwise Classification Circuits.}
\textbf{Circuit compositionality for classification.} We evaluate circuit-based pairwise classification with CLIP, where logits are 
computed via dot product against the full ImageNet-1k text embedding matrix, 
as described in \autoref{app:setup}. We explicitly mine binary circuits for each class pair 
using the target logit difference ($4\times10^{-4}$ threshold) over 8 seeds, with 
datasets composed of 50\% correctly classified examples from each class. We find 
strong overlaps between the mined binary circuits and the individual per-class 
circuits, suggesting that pairwise discrimination reuses the same underlying 
computational pathways.

\begin{figure}[t]
    \centering
    \includegraphics[width=0.9\linewidth]{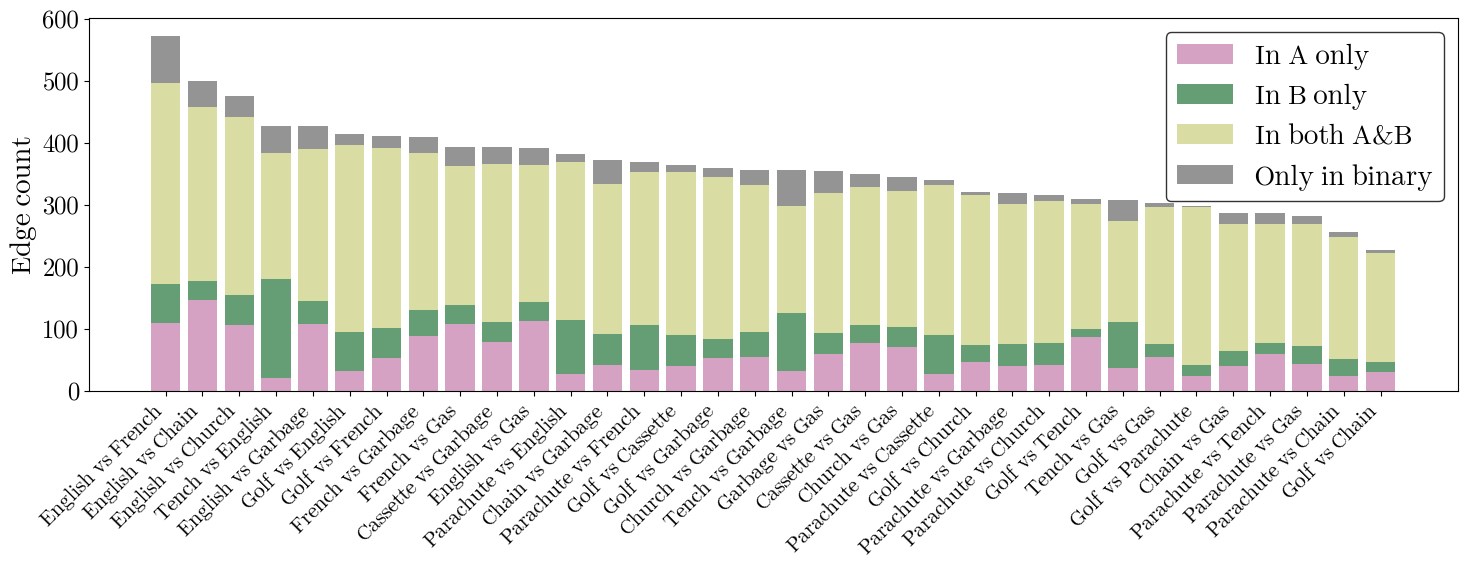}
    \caption{\textbf{Circuit class specificity.} For each class pair (A, B), edges in each binary circuit run are 
    classified as: appearing in the union of all per-class circuits across runs 
    for class A but not B (\textit{A only}); appearing in the union for 
    class B but not A (\textit{B only}); appearing in the union for both 
    classes (\textit{both A\&B}); or appearing in neither class-specific 
    union (\textit{only in binary}). The $y$-axis reports the mean edge 
    count per binary circuit run, averaged over 8 seeds. The dominant contribution 
    of shared edges across all class pairs indicates strong overlap between 
    the individual class circuits and the explicitly mined binary circuit, suggesting 
    that binary discrimination reuses the same computational pathways as the 
    underlying class-specific circuits.}
    \label{fig:binary_circuit_task}
\end{figure}


%% file: sections/appendix_ablations.tex
\section{Choice of Edge Selection Criterion}
\label{app:kl_logit_diff_ablations}

We consider two edge selection criteria: Kullback–Leibler (KL) divergence, following prior work on large language models \cite{conmy2023towards}, and the target logit difference. 

We sweep over 15 different thresholds for each setting. For CLIP, the thresholds are $[5\times10^{-8},\, 2\times10^{-5}]$ for KL divergence and $[1\times10^{-4},\, 1\times10^{-3}]$ for the target logit difference. For ViT-B, the thresholds are $[5\times10^{-8},\, 2\times10^{-1}]$ for KL divergence and $[5\times10^{-3},\, 1\times10^{-1}]$ for the target logit difference.

Circuits are mined on a batch of 128 samples, on a class circuit task \autoref{app:class_circuits}, and the maximum number of visited nodes is set to 900.

\input{tables/ablations_kld_logit_diff}

%% file: tables/ablations_kld_logit_diff.tex
\begin{figure}[t]
\centering

\begin{subfigure}[t]{0.48\textwidth}
	\centering
	\includegraphics[width=\linewidth]{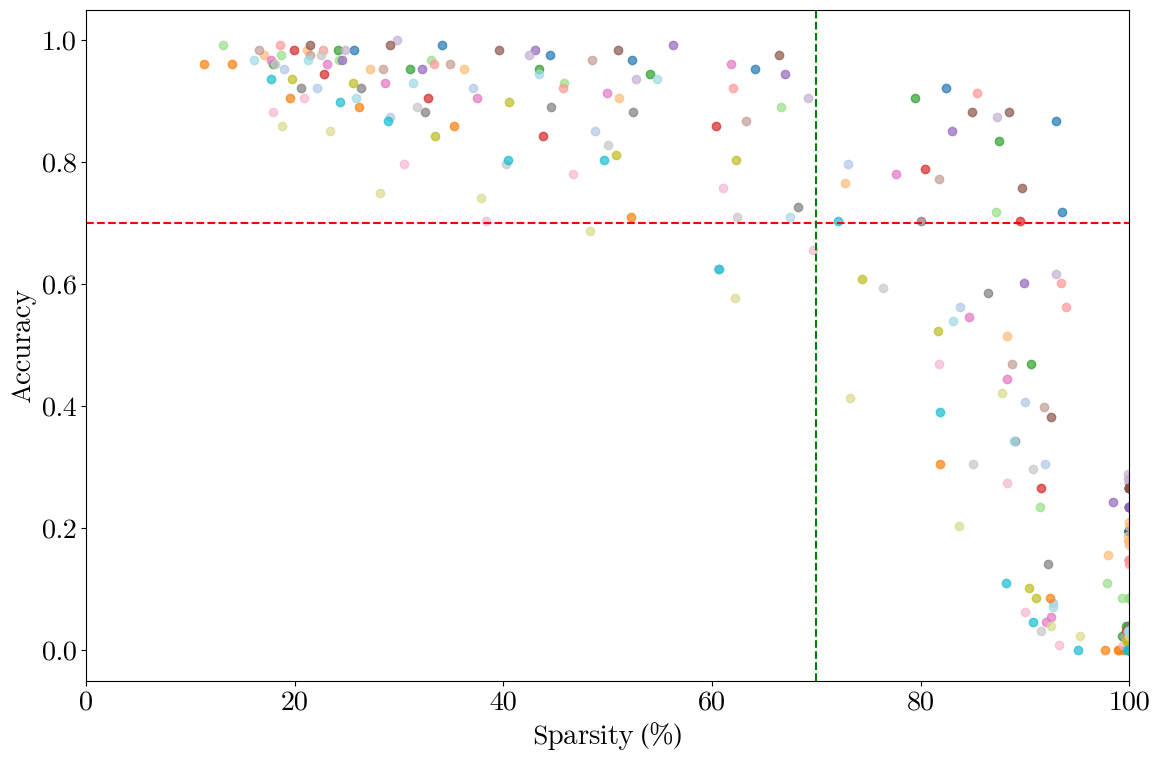}
	\caption{Class Circuits CLIP with KLD}
\end{subfigure}\hfill
\begin{subfigure}[t]{0.48\textwidth}
	\centering
	\includegraphics[width=\linewidth]{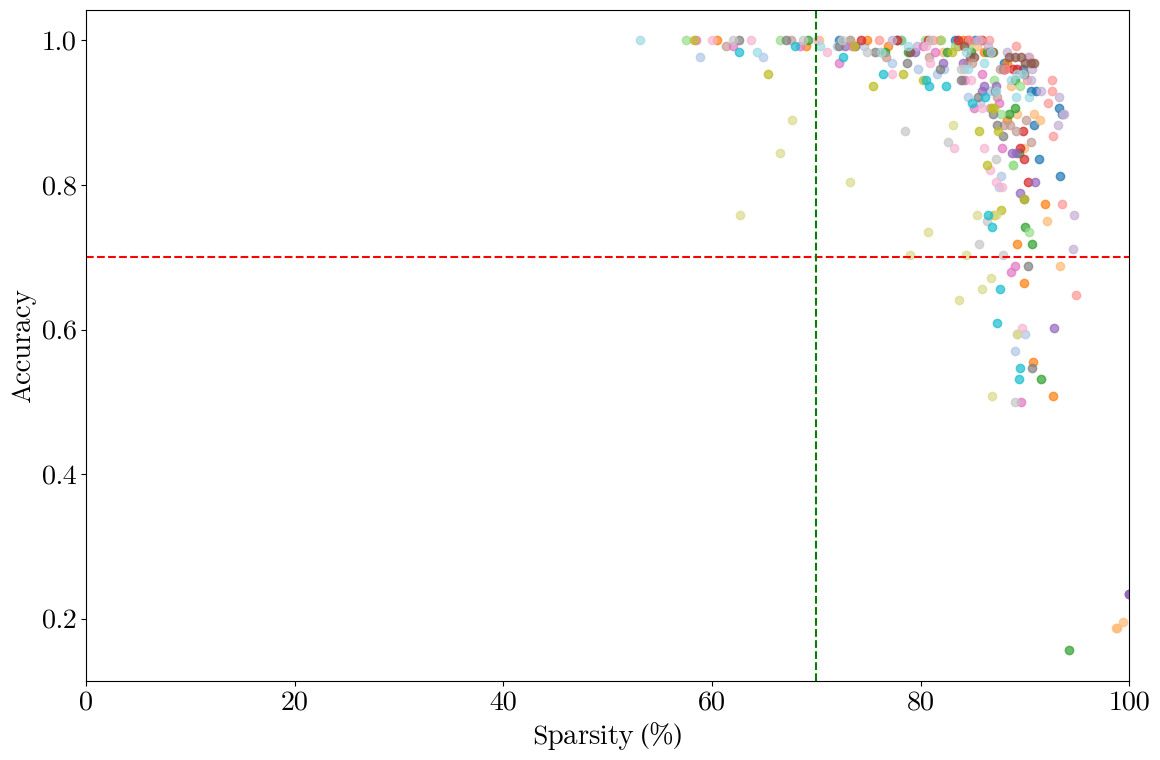}
	\caption{Class Circuits CLIP with target logit difference}
\end{subfigure}

\vspace{0.8em}
\begin{subfigure}[t]{0.48\textwidth}
	\centering
	\includegraphics[width=\linewidth]{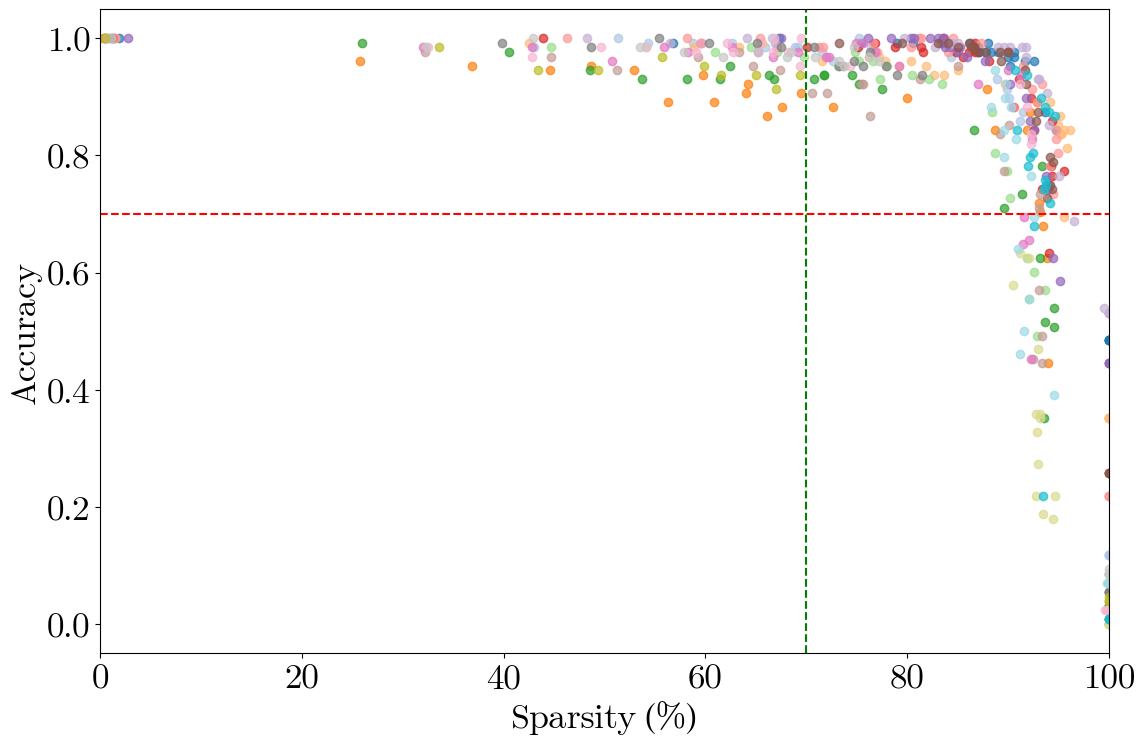}
	\caption{Class Circuits ViT-B with KLD}
\end{subfigure}\hfill
\begin{subfigure}[t]{0.48\textwidth}
	\centering
	\includegraphics[width=\linewidth]{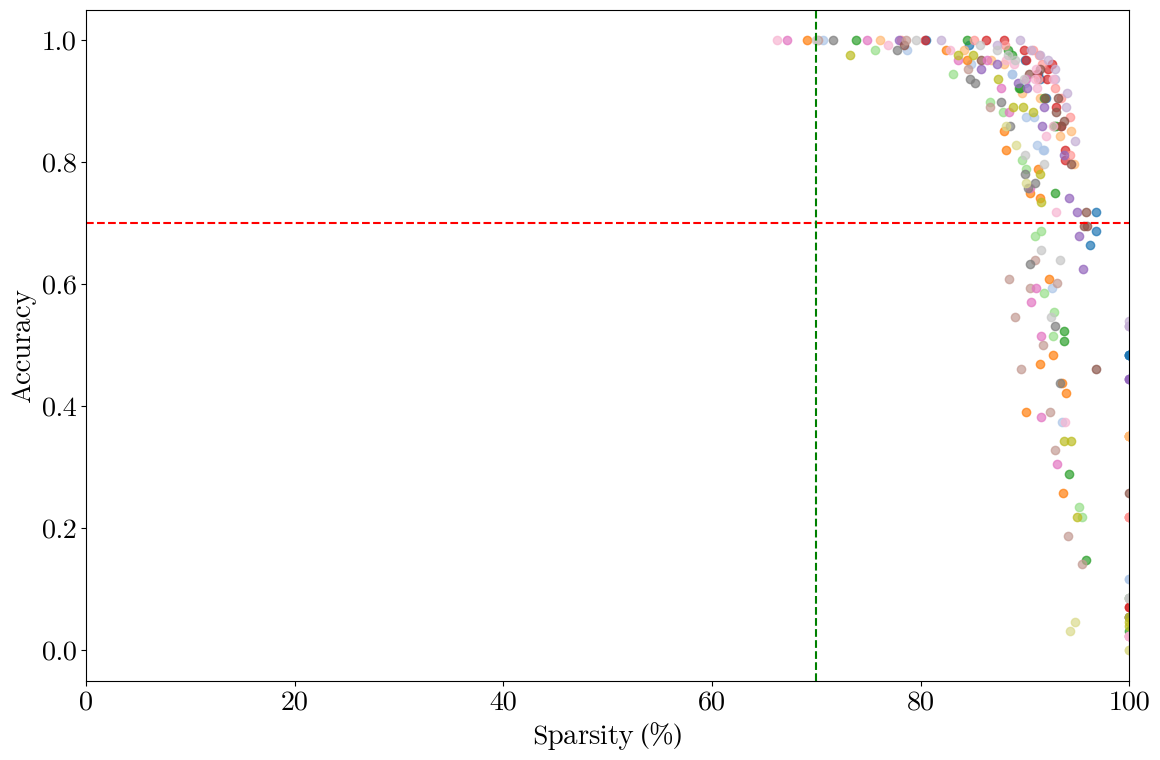}
	\caption{Class Circuits ViT-B with target logit difference}
\end{subfigure}
\vspace{0.8em}

\begin{subfigure}[t]{\textwidth}
	\centering
	\includegraphics[width=\linewidth]{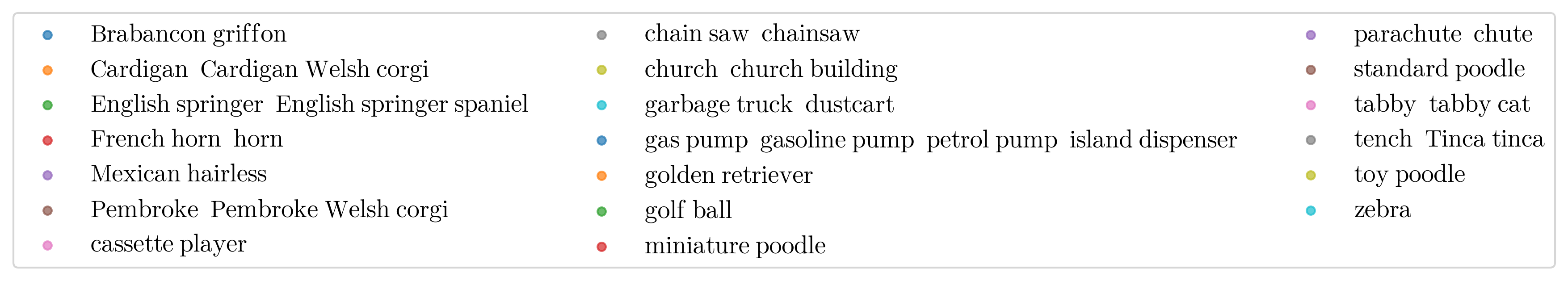}
\end{subfigure}\hfill

\caption{\textbf{Ablations of selection criterion.} For edge typographic circuits we report accuracy and achieved sparsity for each target class for different selection criteria and backbones. Green dotted line marks 70\% sparsity, red dotted line marks 70\% accuracy.}
\label{fig:heatmap-grid-kld_logit}
\end{figure}

%% file: sections/appendix_max_receiver.tex
\section{Subcircuits Ablations}
\label{app:max_receiver_full}

To isolate the effect of circuit-based steering, we conduct ablations across three edge-selection conditions: (i) circuit edges, (ii) random edges of matched size, and (iii) random non-circuit edges of matched size. For each condition, we report performance retention and attack success rate (ASR) as a function of both steering strength $\alpha$ and the maximal receiver layer included in the circuit. Results are shown in Fig.~\ref{fig:subcircuit_ablations}.

\input{tables/max_receiver_plots}

\paragraph{Circuit edges.}
Steering along discovered circuit edges (Fig.~\ref{fig:subcircuit_ablations}a--b) yields a favorable trade-off between safety and utility. At low-to-moderate steering strengths, performance retention remains high (green region in Fig.~\ref{fig:subcircuit_ablations}a), while ASR is simultaneously suppressed across a wide range of receiver layer depths (dark blue region in Fig.~\ref{fig:subcircuit_ablations}b). Notably, even restricting steering to early receiver layers is sufficient to substantially reduce ASR, consistent with our findings in Sec.~4.4 that typographic information is routed through identifiable pathways already in the earlier layers of the model.

\paragraph{Random edges.}
When steering is applied along random edges of matched size (~\autoref{fig:subcircuit_ablations}c--d), performance degrades comparably to the circuit condition as steering strength increases (~\autoref{fig:subcircuit_ablations}c). However, ASR reduction is markedly weaker and less consistent (~\autoref{fig:subcircuit_ablations}d): the suppression visible in ~\autoref{fig:subcircuit_ablations}d is substantially lighter than in ~\autoref{fig:subcircuit_ablations}b, indicating that random edge ablation achieves only limited suppression of attack-relevant information across most receiver layers and steering strengths. This confirms that the ASR reduction observed under circuit steering is not simply a consequence of ablating any arbitrary set of edges.
Steering along random edges explicitly drawn from outside the discovered circuit (~\autoref{fig:subcircuit_ablations}e--f) reveals a further contrast. While performance retention degrades similarly under sufficient steering strength (~\autoref{fig:subcircuit_ablations}e), the ASR response is counterproductive when steering is applied up to middle layers (~\autoref{fig:subcircuit_ablations}f): across much of the steering strength and receiver layer grid, ASR is not meaningfully reduced and for strong steering on edges up to the mid-layers of the model, ASR is elevated relative to non-steered baseline (orange tones in ~\autoref{fig:subcircuit_ablations}f), suggesting that intervening on non-circuit edges can interfere with unrelated computations without suppressing the attack pathway, leading to even higher ASR. This instability further validates the specificity of the discovered circuits as the mechanistically relevant subgraph for the typographic attack behavior.

%% file: tables/max_receiver_plots.tex
\begin{figure}[h]
\centering

\begin{subfigure}[t]{0.48\textwidth}
	\centering
	\includegraphics[width=\linewidth]{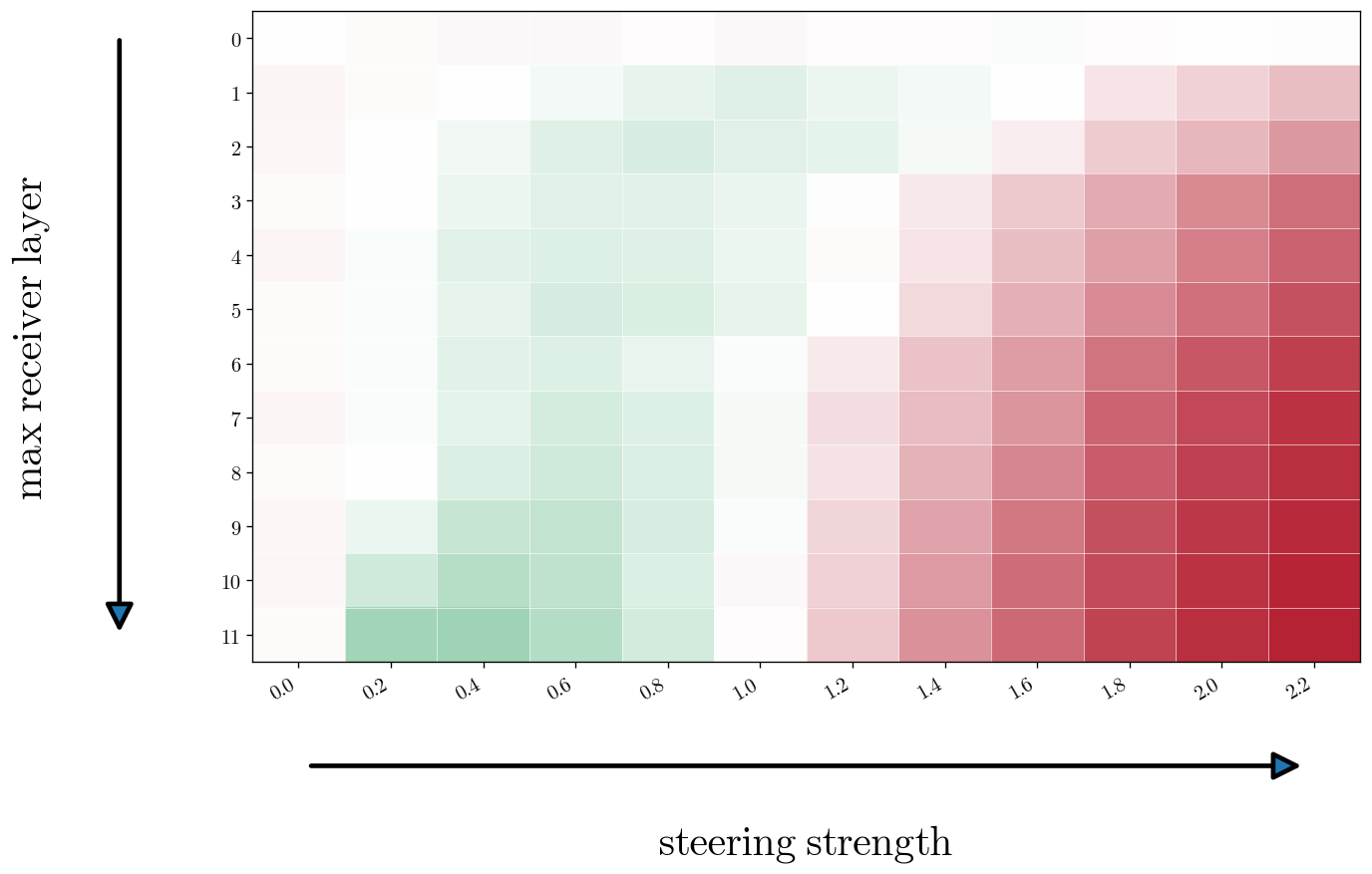}
	\caption{Retention, circuit edges}
\end{subfigure}\hfill
\begin{subfigure}[t]{0.48\textwidth}
	\centering
	\includegraphics[width=\linewidth]{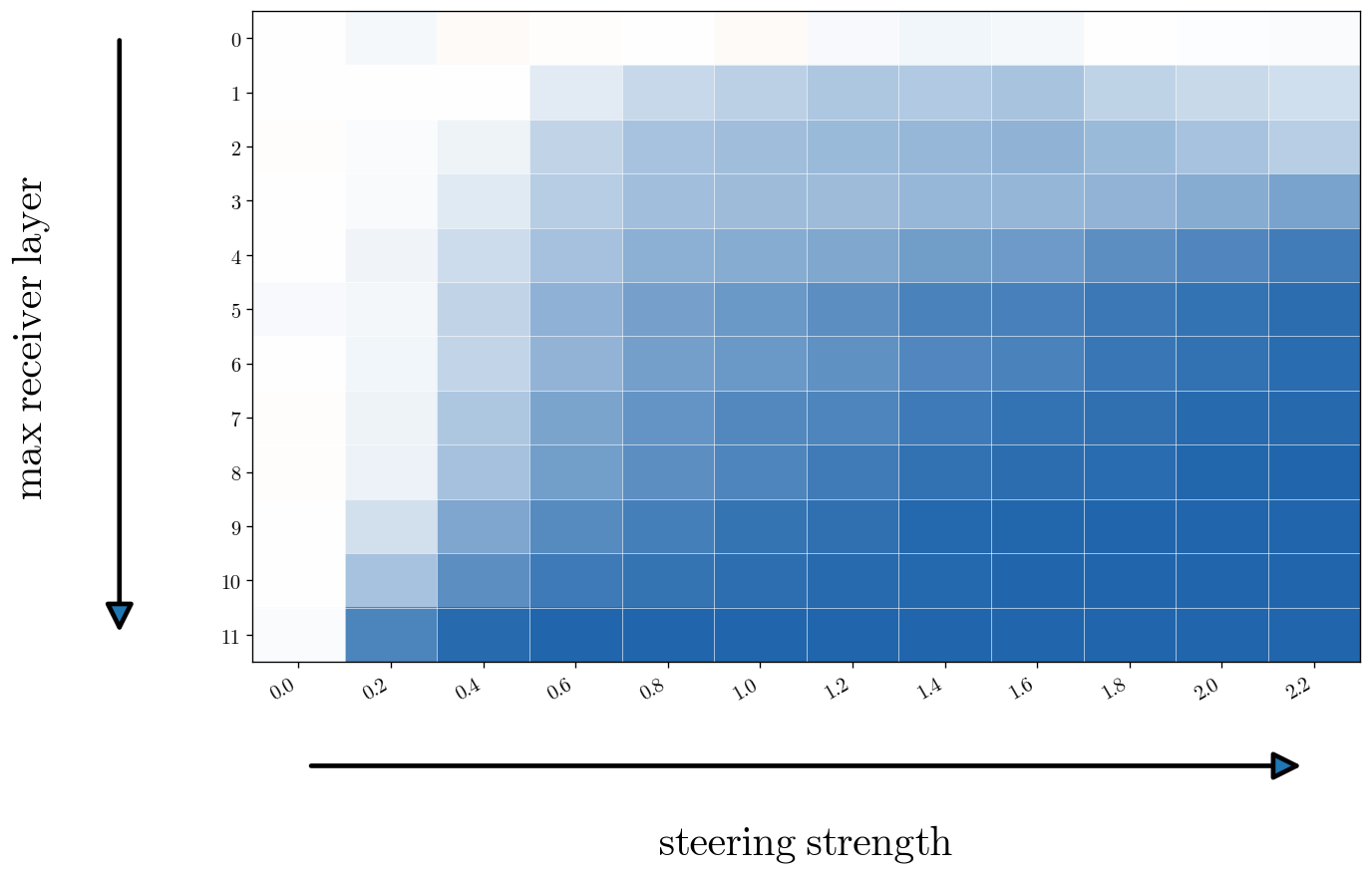}
	\caption{ASR, circuit edges}
\end{subfigure}

\vspace{0.8em}
\begin{subfigure}[t]{0.48\textwidth}
	\centering
	\includegraphics[width=\linewidth]{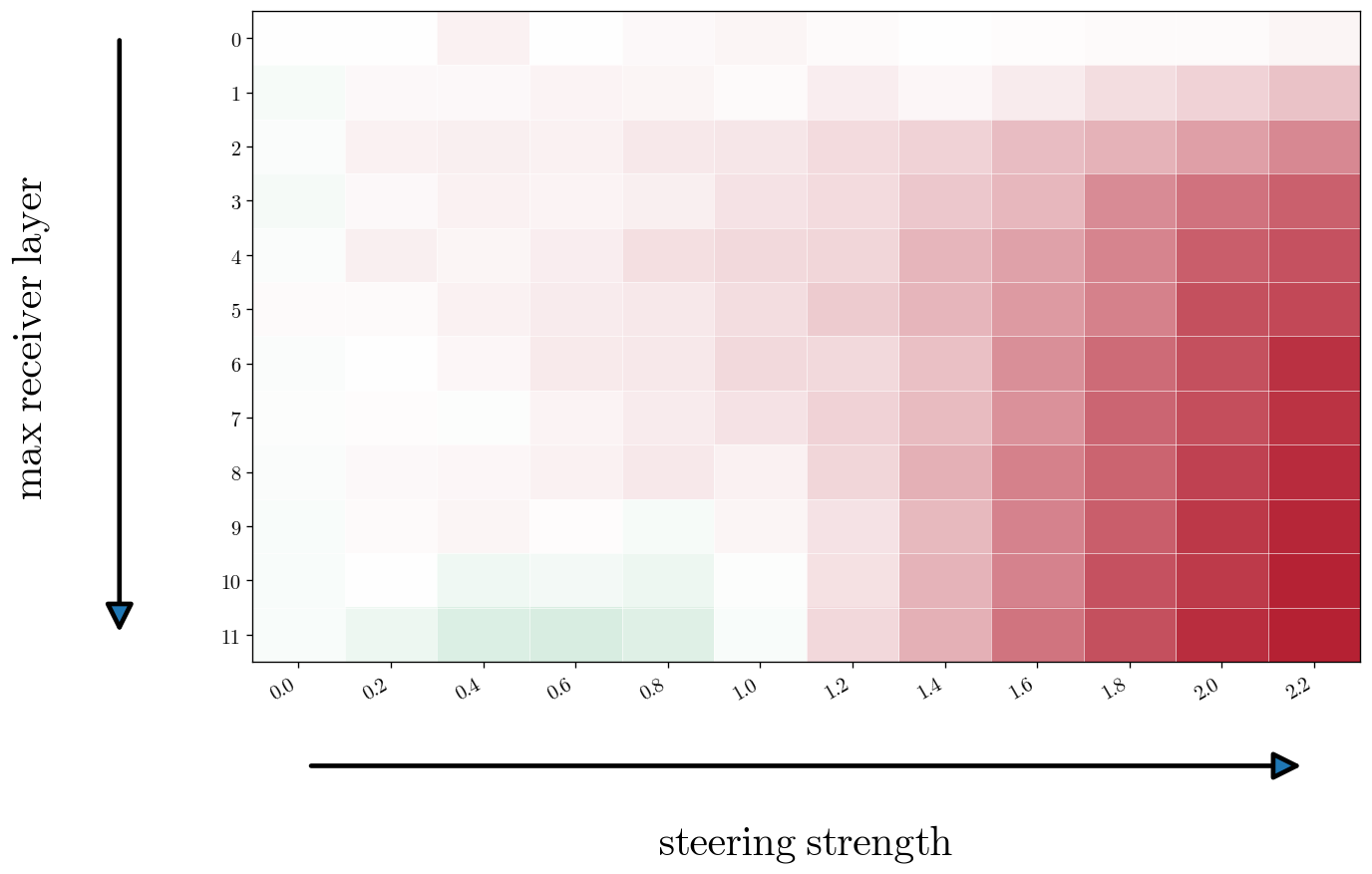}
	\caption{Retention, random edges}
\end{subfigure}\hfill
\begin{subfigure}[t]{0.48\textwidth}
	\centering
	\includegraphics[width=\linewidth]{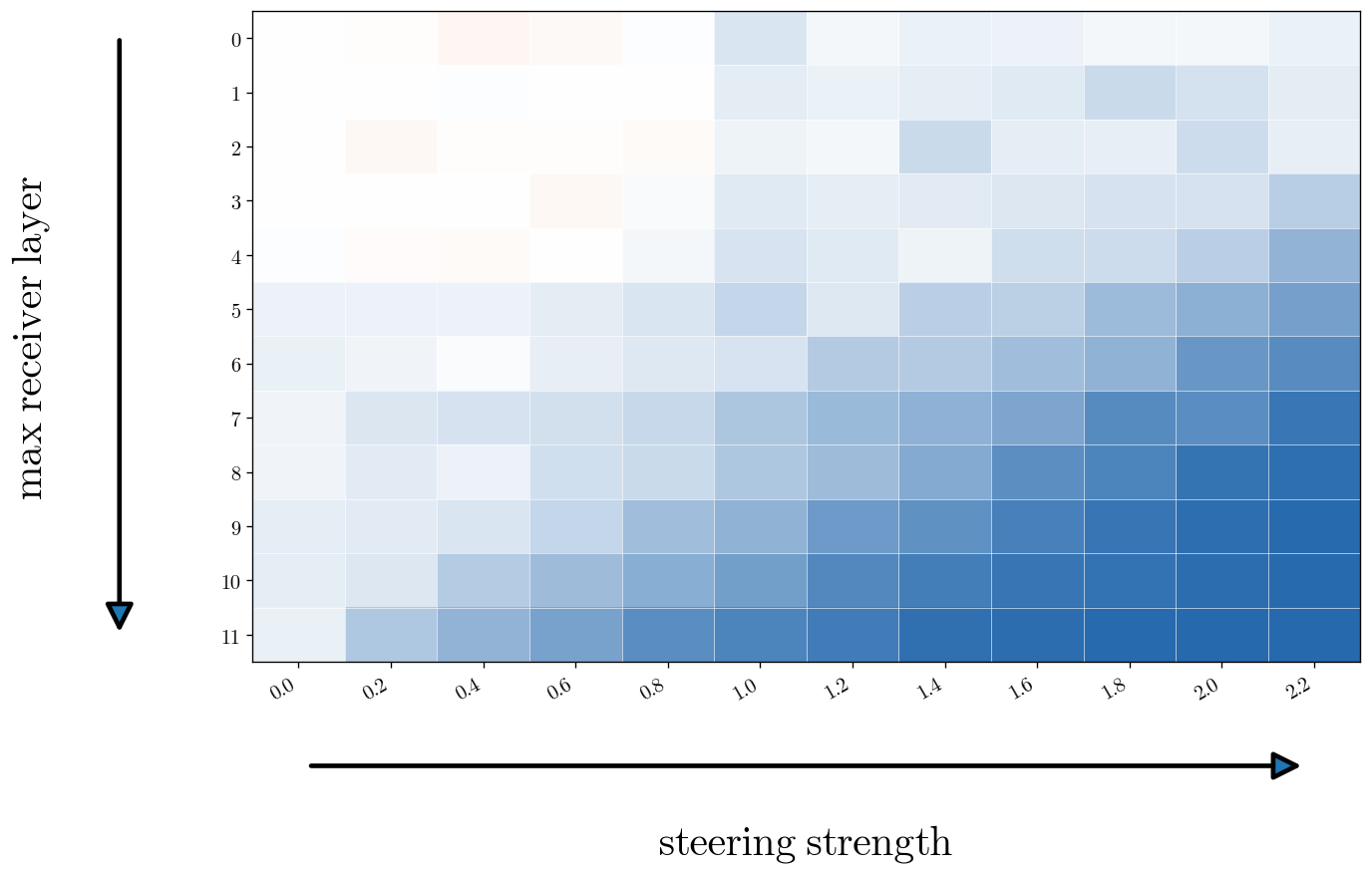}
	\caption{ASR, random edges}
\end{subfigure}
\vspace{0.8em}

\begin{subfigure}[t]{0.48\textwidth}
	\centering
	\includegraphics[width=\linewidth]{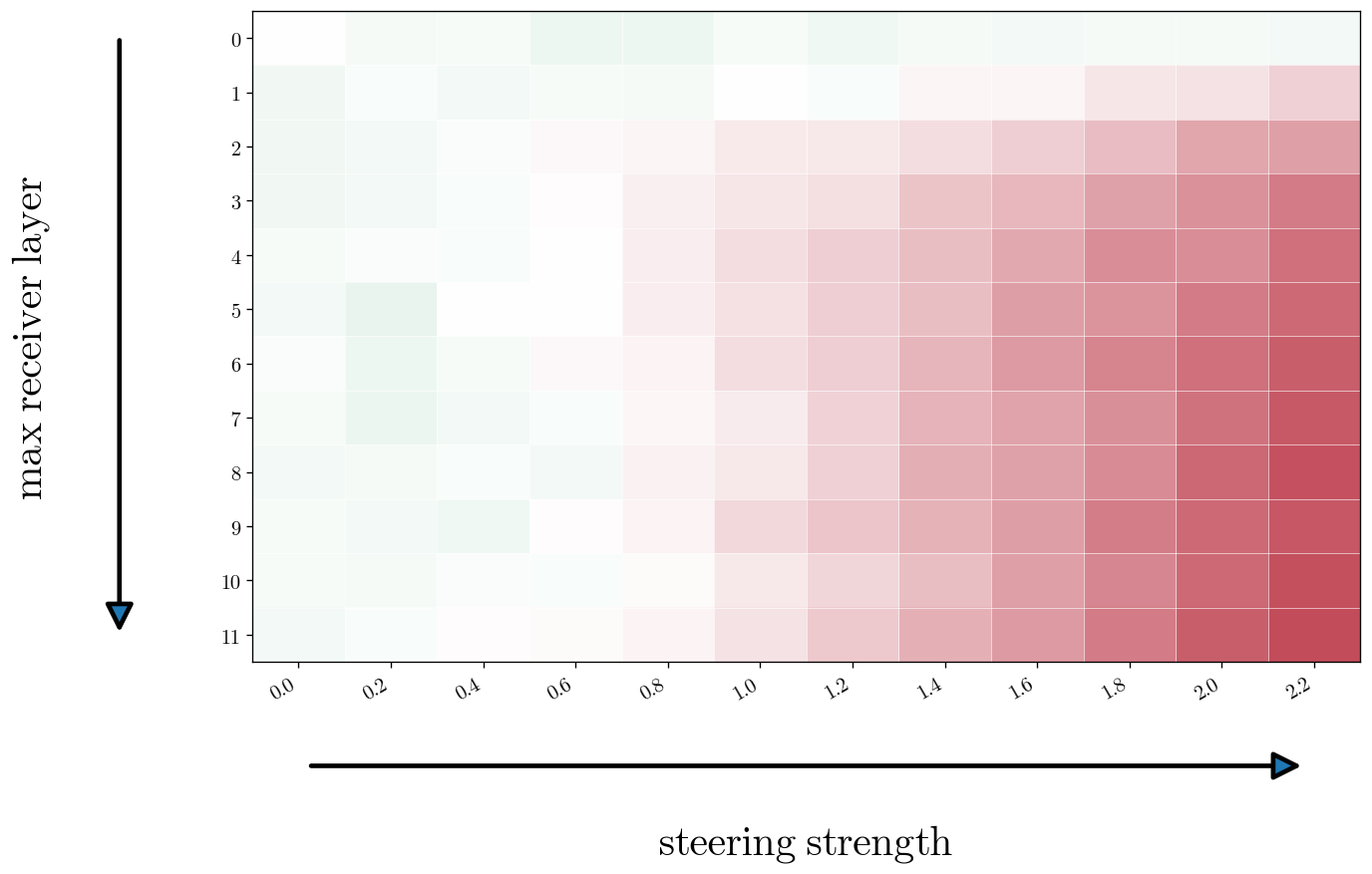}
	\caption{Retention, random non-circuit edges}
\end{subfigure}\hfill
\begin{subfigure}[t]{0.48\textwidth}
	\centering
	\includegraphics[width=\linewidth]{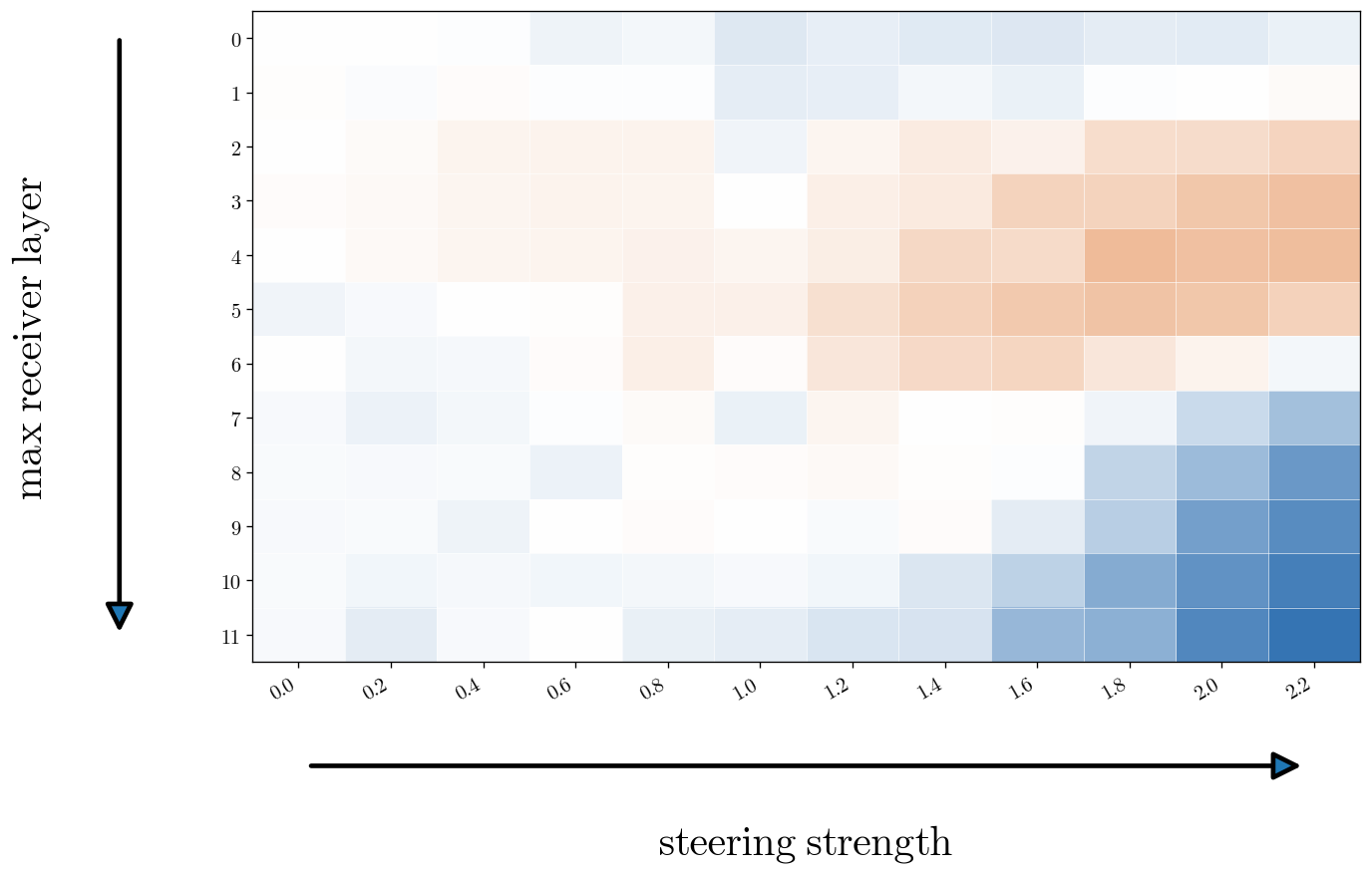}
	\caption{ASR, random non-circuit edges}
\end{subfigure}

\hspace*{0.08\textwidth}
\begin{subfigure}[t]{0.4\textwidth}
	\centering
	\includegraphics[width=\linewidth]{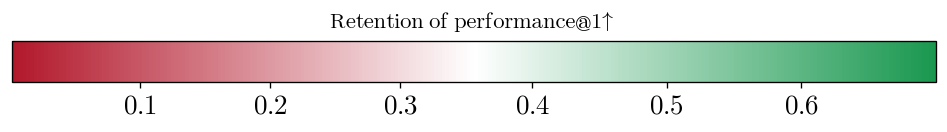}
\end{subfigure}\hfill
\begin{subfigure}[t]{0.4\textwidth}
	\centering
    \includegraphics[width=\linewidth]{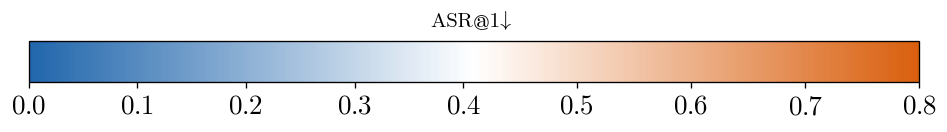}

\end{subfigure}

\caption{\textbf{Ablations of typographic circuits up to a layer.} 
    The $x$-axis denotes steering strength $\alpha$ and the $y$-axis the maximal receiver layer included in the circuit (we are ignoring later layers of the circuit). 
    Left column reports performance retention ($\uparrow$ higher is better); right column reports attack success rate (ASR, $\downarrow$ lower is better). 
    Rows correspond to: circuit edges (top), random edges of matched size (middle), and random non-circuit edges of matched size (bottom). 
    Circuit-edge steering uniquely achieves simultaneous ASR suppression and performance retention on corrupted samples across steering strengths and receiver layers.}
    \label{fig:subcircuit_ablations}
\end{figure}